\documentclass{article}

\usepackage[main, final]{neurips_2026}

\usepackage[utf8]{inputenc}
\usepackage[T1]{fontenc}
\usepackage{hyperref}
\usepackage{url}
\usepackage{booktabs}
\usepackage{amsfonts}
\usepackage{amsmath}
\usepackage{nicefrac}
\usepackage{microtype}
\usepackage{xcolor}
\usepackage{graphicx}
\usepackage{multirow}
\usepackage{algorithm}
\usepackage{algpseudocode}
\usepackage{amssymb}
\usepackage{multicol}
\usepackage{tabularx}
\usepackage{listings}

\definecolor{listingbg}{HTML}{FAFAFA}      %
\definecolor{baselinebg}{HTML}{8A8A8A}     %
\definecolor{gepabg}{HTML}{0BAEF4}         %
\definecolor{gepamergebg}{HTML}{1F4E79}    %
\definecolor{miprobg}{HTML}{4DAF50}        %
\definecolor{reapobg}{HTML}{F68607}        %

\lstdefinestyle{prompt-base}{
  basicstyle=\ttfamily\footnotesize,
  breaklines=true,
  breakatwhitespace=false,
  columns=fullflexible,
  keepspaces=true,
  showstringspaces=false,
  backgroundcolor=\color{listingbg},
  frame=l,
  framerule=2.5pt,
  framesep=10pt,
  xleftmargin=14pt,
  xrightmargin=10pt,
  aboveskip=2pt,
  belowskip=6pt,
  literate={---}{{-{}-{}-}}3 {--}{{-{}-}}2,
}
\lstdefinestyle{prompt-baseline} {style=prompt-base, rulecolor=\color{baselinebg}}
\lstdefinestyle{prompt-gepa}     {style=prompt-base, rulecolor=\color{gepabg}}
\lstdefinestyle{prompt-gepamerge}{style=prompt-base, rulecolor=\color{gepamergebg}}
\lstdefinestyle{prompt-mipro}    {style=prompt-base, rulecolor=\color{miprobg}}
\lstdefinestyle{prompt-reapo}    {style=prompt-base, rulecolor=\color{reapobg}}
\lstdefinestyle{prompt}          {style=prompt-baseline}

\newcommand{\promptbox}[2]{%
  \par\addvspace{6pt}%
  \noindent\hspace*{0pt}%
  \textcolor{#2}{\rule[0.1ex]{0.55em}{0.55em}}\hspace{0.45em}%
  {\color{#2}\textbf{\footnotesize #1}}%
  \par\nobreak\vspace{1pt}%
}

\title{Simplicity is Enough:\\ ReAct Agents for Prompt Optimization}

\author{%
  Andrei Dumitrescu$^{\dagger}$ \quad
  Andrei C. Rusu$^{\dagger*}$ \\
  \textbf{Adrian Catalin Badea} \quad
  \textbf{Bai Li} \quad
  \textbf{Cosmin Maria} \\
  \textbf{Ion Anghelina} \quad
  \textbf{Tomasz Lukasz Religa} \quad
  \textbf{Dragos H Bobolea} \\
  UiPath
}

\begin{document}

\maketitle
{\renewcommand{\thefootnote}{}\footnotetext{$^{\dagger}$Equal contribution.}\footnotetext{$^{*}$Corresponding author: \texttt{rusu.c.andrei@gmail.com}}}

\begin{abstract}
Prompt optimization has recently been approached with increasingly specialized search procedures and evolutionary frameworks. In this work, we ask whether such complexity is necessary to achieve competitive performance. We introduce ReAPO (ReAct Agents for Prompt Optimization), a lightweight framework that casts prompt optimization as an agentic process built around a ReAct loop augmented with evaluation and reflection tools. Rather than relying on bespoke optimization machinery, ReAPO iteratively refines prompts through error analysis, reflective reasoning, and validation against the optimization objective. We evaluate ReAPO on several established benchmarks spanning multi-hop question answering, mathematical reasoning, fact verification, instruction following, and privacy-conscious delegation, as well as on real enterprise agentic tasks. Across these settings, this agentic optimizer can match or outperform specialized prompt optimization methods, with 2–7$\times$ fewer evaluation rollouts and a lower API cost on the public benchmarks. We also report confidence intervals over multiple independent runs and show that single-run evaluations on small benchmarks can give a misleading picture of relative performance, since LLM variance may obscure true gains. Taken together, our results suggest that effective prompt optimization need not depend on increasingly elaborate search procedures and that agentic methods can provide a more efficient alternative.
\end{abstract}

\begin{figure}[!h]
\centering
\includegraphics[width=0.97\linewidth, trim=7 9 7 7, clip]{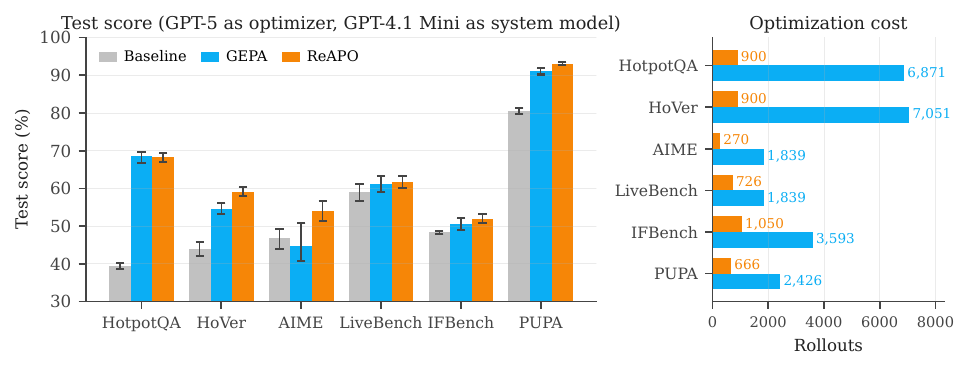}
\caption{ReAPO matches or exceeds GEPA on six public benchmarks. \textbf{Left:} Test scores (GPT-5 optimizer, GPT-4.1~Mini eval) with 95\% CIs over five runs. \textbf{Right:} Optimization cost in terms of evaluation-rollout budgets across training and/or validation splits.}
\label{fig:progression}
\end{figure}

\section{Introduction}

Prompting has emerged as a primary mechanism for adapting Large Language Models (LLMs) to
downstream tasks without parameter updates. This has made prompt
optimization, the automatic improvement of natural language instructions with respect to a task objective, an attractive alternative to fine-tuning for practical deployment~\citep{zhou2023large, pryzant2023automatic, khattab2024dspy}. Recent methods commonly formulate this task as a search problem over a discrete space of candidate instructions: prompts are proposed, evaluated, and iteratively refined in order to improve performance. Within this paradigm, methods such as APO~\citep{pryzant2023automatic}, MIPROv2~\citep{opsahl-ongOptimizingInstructionsDemonstrations2024}, and GEPA~\citep{agrawalGEPAReflectivePrompt2025}
 have demonstrated substantial gains, but they do so through increasingly specialized optimization workflows, including evolutionary operators, Pareto-front maintenance, and mutation heuristics. This trend 
reflects a recurring pattern in prompt optimization: empirical 
progress has come at the cost of escalating outer-loop complexity, 
large rollout budgets, and, at times, 
hand-crafted feedback signals that require
task-specific engineering effort.

To address these limitations, we introduce \textbf{ReAPO} (\textbf{Re}Act \textbf{A}gents for \textbf{P}rompt \textbf{O}ptimization), which formulates prompt optimization as a ReAct-style loop~\citep{yao2023react} equipped with evaluation and introspection tools. Starting from an initial LLM system and its component prompts, the agent repeatedly evaluates the system on a training set, analyzes failures and divergences from the ground truth, identifies error patterns, and proposes revisions. ReAPO follows a single greedy lineage, similar to TextGrad~\citep{yuksekgonul2025textgrad}: it records the prompts of every round and returns the best on the training set. Its central claim is one of sufficiency: competitive performance in prompt optimization does not necessitate elaborate evolutionary mechanisms.

We evaluate on six established benchmarks, spanning tasks from multi-hop question answering and fact verification to mathematical reasoning, instruction following, and privacy-conscious delegation. Across these benchmarks, we investigate the behavior of several optimizer models and compare our approach against state-of-the-art baselines, MIPROv2 and GEPA, in terms of both task performance and optimization cost. To assess how well our approach generalizes to smaller-scale models, we first apply the optimized prompts zero-shot to Qwen3.5-9B~\citep{qwen2025qwen3}. We then repeat the full optimization procedure with this model serving as both the optimizer and the system model. Complementing these public-benchmark evaluations, we study five enterprise agentic tasks to test whether conclusions drawn from curated academic suites persist in practical, real-world settings.

Additionally, our ablations expose benchmark saturation: with a GPT-5 optimizer~\citep{openai2025gpt5}, ReAPO stays competitive on the public benchmarks without its introspection tools but trails substantially on specialized enterprise workflows. This asymmetry suggests that frontier models have internalized public-benchmark patterns through repeated exposure, while enterprise workloads, unlikely to appear in training data, require genuine task-specific refinement. Evaluating prompt optimization on less-saturated agentic benchmarks is therefore essential. To summarize, our main contributions are:
\begin{itemize}
    \item We introduce \textbf{ReAPO}\footnote{Code and prompts will be made available at: \url{https://github.com/UiPath/agents-gym-core}.}, a prompt optimization framework built around a ReAct agent loop with evaluation and reflection tools, which refines a single prompt lineage rather than a population of candidates. Across six public benchmarks, it matches or outperforms GEPA and MIPROv2 on most tasks while using 2--7\(\times\) fewer evaluation rollouts and incurring a lower API cost. These findings hold at smaller scale: the resulting prompts transfer zero-shot to Qwen3.5-9B, and ReAPO remains effective even when this model serves as both the optimizer and the system model.

    \item To account for the documented variability in LLM evaluation~\citep{biderman2024lessons}, we report bootstrap confidence intervals across multiple independent runs. Our results show that this variability can materially alter the apparent ranking of prompt optimization methods, complicating the interpretation of performance gains reported from a single run.

    \item We extend the evaluation to real enterprise agentic tasks, where ReAPO yields strong gains in most cases. Tool ablations expose a key asymmetry in its components: without its introspection tools, ReAPO with a GPT-5 optimizer remains competitive on public benchmarks but underperforms on specialized enterprise workflows. This suggests that the optimizer’s latent familiarity with a task can compensate for the absence of introspection.
\end{itemize}

\section{Related Work}

Adapting LLMs to downstream tasks can proceed either by modifying their weights, for example via reinforcement learning methods such as GRPO~\citep{shao2024deepseekmath}, or by modifying the input context. Weight-based methods require gradient access and numerous rollouts, motivating a growing line of work on prompt optimization, which operates solely in text space and treats the LLM as a black box.

\paragraph{Single-prompt optimization.} A first line of work targets the single-prompt setting, with the LLM responsible for both proposing and evaluating candidate instructions against a task objective. APE~\citep{zhou2023large} introduces a generate-then-select pipeline, while OPRO~\citep{yang2024large} extends this loop by conditioning each new proposal on the trajectory of prior candidates and their scores, framing the LLM itself as the optimizer. APO~\citep{pryzant2023automatic} adds error-driven refinement, at the cost of per-task tuning of bandit and beam hyperparameters, while PromptAgent~\citep{wang2023promptagent} performs Monte Carlo tree search over LLM-generated edit operators, requiring evaluation at every expanded node. More recent agentic approaches retain this single-prompt scope but layer additional optimization machinery on top. ACE~\citep{zhang2026ace} decomposes optimization across three LLM roles (Generator, Reflector, Curator), with per-benchmark prompts and per-datapoint iterative refinement across multiple adaptation epochs that push its LLM-call budget on a different scale. Its optimized state is a playbook consumed by the Generator at inference rather than a standalone prompt, coupling deployment to ACE's pipeline. AutoPDL~\citep{spiess2025autopdl} frames the problem as configuration-space search over a predefined set of prompting patterns, requiring extensive per-task configuration over a combinatorial space that grows with the number of patterns. Despite their algorithmic diversity, none of these methods address credit assignment across multi-module pipelines, the regime in which ReAPO operates, and most require substantial rollout budgets.

\paragraph{Multi-stage and compound system optimization.} As LLM applications grew into multi-stage pipelines, prompt optimization became considerably harder, since changes to one module's instructions can harm downstream modules and credit assignment across the pipeline is non-trivial. TextGrad~\citep{yuksekgonul2025textgrad} addresses this through gradient-style backpropagation, sending LLM critiques backward through a user-defined computation graph over the target system, with an LLM call at every node during the backward pass. DSPy~\citep{khattab2024dspy} provides a complementary foundation by decoupling program structure from optimizable signature prompts, defining a well-posed multi-stage credit-assignment problem. Building on this abstraction, MIPROv2~\citep{opsahl-ongOptimizingInstructionsDemonstrations2024} factorizes the joint optimization of instructions and few-shot demonstrations across modules via a Bayesian surrogate, while GEPA~\citep{agrawalGEPAReflectivePrompt2025} applies reflective evolution over a Pareto archive of candidates, optionally guided by hand-crafted per-module feedback functions. GEPA is shown to outperform both TextGrad and MIPROv2, but requires substantial rollout budgets, justifying its use as the principal baseline and focusing the comparison on sample efficiency.

\paragraph{Agentic optimization.} Our approach builds on the ReAct paradigm~\citep{yao2023react}, which has been applied across software engineering~\citep{jimenez2024swebench}, machine learning design~\citep{huang2024benchmarking}, and scientific discovery~\citep{boiko2023autonomous}. In an adjacent domain, AVO~\citep{chen2026avo} demonstrates that agentic procedures can match elaborate search in GPU kernel discovery, providing evidence that agentic loops can substitute for specialized machinery. We extend this thesis to multi-stage prompt optimization by casting the optimizer as a ReAct agent whose environment exposes an evaluation infrastructure, supplanting Pareto fronts, surrogate models, and hand-crafted feedback functions with a unified reasoning-and-action loop guided by a natural-language protocol.

\section{Method}
\label{sec:method}

\subsection{Problem Formulation}

Let $\mathcal{A}$ denote a compound LLM-based system parameterized 
by the model $\mathcal{M}_{\text{sys}}$ and a set of natural-language instructions 
$\boldsymbol{\pi} = (\pi_1, \ldots, \pi_K)$, where $K$ is the number of independently modifiable signature prompts. Without loss of generality, we will henceforth refer to such systems as \textit{agents} for simplicity. Given a set of $M$ metric functions
$\{m_j : \hat{\mathcal{Y}} \times \mathcal{C} \to [0,1]\}_{j=1}^{M}$,
where $\hat{\mathcal{Y}}$ is the space of agent outputs and $\mathcal{C}$
is the space of evaluation criteria, and a training set
$\mathcal{D}_{\text{train}} = \{(x_i, \boldsymbol{c}_i)\}_{i=1}^{N}$,
where $\boldsymbol{c}_i \in \mathcal{C}^M$ is a tuple of $M$ per-evaluator
criteria (e.g.\ reference outputs, expected tool call sequences, or
behavioral constraints), we define the per-datapoint and overall system
scores, respectively, as:
\begin{equation}
\label{eq:per_datapoint}
s_i(\boldsymbol{\pi};\mathcal{M}_{\text{sys}}) \;=\; 
\frac{1}{M}\sum_{j=1}^{M} m_j\bigl(\mathcal{A}(x_i;\boldsymbol{\pi},\mathcal{M}_{\text{sys}}),\; \boldsymbol{c}_i^j\bigr),
\end{equation}
\begin{equation}
\label{eq:objective}
S(\boldsymbol{\pi}; \mathcal{M}_{\text{sys}},\mathcal{D}) \;=\; 
\frac{1}{|\mathcal{D}|} \sum_{i=1}^{|\mathcal{D}|} s_i(\boldsymbol{\pi};\mathcal{M}_{\text{sys}}).
\end{equation}
The goal of prompt optimization is to find 
$\boldsymbol{\pi}^* = \arg\max_{\boldsymbol{\pi}}\, 
S(\boldsymbol{\pi}; \mathcal{M}_{\text{sys}}, \mathcal{D}_{\text{train}})$ 
that generalizes to unseen data $\mathcal{D}_{\text{test}}$. Most existing approaches solve this by actively maintaining a \emph{population} of candidate prompt sets $\{\boldsymbol{\pi}^{(0)}, \ldots, \boldsymbol{\pi}^{(P)}\}$, and employing evolutionary search operators (e.g.\ mutation, Pareto selection, etc.) to explore the prompt space over multiple generations. 

Our paper proposes a more straightforward strategy: delegating the entire optimization loop, including what to analyze, which prompts to modify, and when to revert potentially harmful changes, to a single ReAct agent equipped with evaluation and introspection tools, operating across $R$ sequential rounds. After all rounds are complete, we return the prompt set that achieves the highest training score:
\begin{equation}
\label{eq:selection}
\boldsymbol{\pi}^* \;=\; \boldsymbol{\pi}^{(r^*)}, \qquad 
r^* = \arg\max_{r \in \{0,\ldots,R\}} S(\boldsymbol{\pi}^{(r)}; \mathcal{M}_{\text{sys}},\mathcal{D}_{\text{train}}).
\end{equation}
ReAPO thus optimizes and selects prompts exclusively on the training set and evaluates the selected prompt unchanged on the disjoint test set. Whether a validation stage is beneficial depends on the data regime: in enterprise agentic applications with few labeled examples, a validation split may be too small for reliable selection while leaving fewer examples for optimization. On the public benchmarks, where labeled data are plentiful, a validation stage would be warranted if training-set selection led to overfitting, yet the held-out results in Section~\ref{sec:experiments} show no systematic evidence of this. On the contrary, the selected prompts encode reusable procedures rather than example-specific answers (Appendix~\ref{app:protocol}).

\subsection{ReAPO: Agentic Prompt Optimization}
\label{sec:reapo}

More formally, we frame prompt optimization as a \emph{multi-round agentic task}. An optimizer LLM $\mathcal{M}_{\text{opt}}$ operates in a ReAct loop, interleaving reasoning steps (\textsc{Think}) with tool invocations (\textsc{Act}) and environment observations (\textsc{Obs}) (see Algorithm~\ref{alg:reapo}). ReAPO’s environment consists entirely of workflow discovery and evaluation infrastructure: it has no access to code execution but read access to candidate prompts, LLM system structure, and cached outputs/scores, orchestrating its own reasoning runs, directing target workflow evaluations, and proposing prompt updates from feedback.

\paragraph{Tool interface.} The optimizer interacts with the target LLM system and its outputs through a Model Context Protocol (MCP) server that exposes five categories of tools: \emph{Agent Inspection} for examining agent structure, prompts, and available tools; \emph{Prompt Editing} for applying candidate revisions; \emph{Results Analysis} for viewing scores, per-datapoint breakdowns, and execution traces; \emph{Agent Execution} for launching and monitoring evaluation runs; and \emph{Logging} for persisting snapshots, score summaries, and tracking budget usage. For a more detailed breakdown of the tools refer to Table~\ref{table:tools_detail}.

\paragraph{Optimization protocol.} The optimizer agent operates under a 
\emph{skill prompt} $\sigma$ (Appendix~\ref{app:skill_prompts}) that encodes the protocol as 
structured instructions rather than algorithmic code. Instead of prescribing 
a fixed execution trace, $\sigma$ scaffolds a round-based plan that 
ReAPO interprets as a behavioral guide. Like the tool interface, $\sigma$ is task-independent and reused unchanged across tasks, making the design of both a one-time cost rather than benchmark-specific search engineering. The protocol begins with a baseline round (round $r = 0$), in which the agent is incentivized to 
inspect the target system to discover its signature prompts
$\{\pi_1, \ldots, \pi_K\}$, saves them as the current best 
$\boldsymbol{\pi}^\dagger$, and runs an initial evaluation to obtain baseline scores $\{s_i^{(0)}\}$. The agent 
then analyzes up to $N_w = 15$ worst-scoring datapoints by 
examining their execution traces alongside the evaluation criteria 
$\boldsymbol{c}_i$, establishing an informed starting point for optimization. Subsequent rounds $r \leq R$ should involve: 

\begin{itemize}
    
    \item \textbf{Evaluate.} Run $\mathcal{A}$ using $\boldsymbol{\pi}^{(r)}$ on $\mathcal{D}_{\text{train}}$ and compute the new score $S(\boldsymbol{\pi}^{(r)}; \mathcal{M}_{\text{sys}}, \mathcal{D}_{\text{train}})$.
    
    \item \textbf{Accept or revert.} If $S(\boldsymbol{\pi}^{(r)}) \geq S(\boldsymbol{\pi}^\dagger)$, the revision is accepted and $\boldsymbol{\pi}^\dagger \leftarrow \boldsymbol{\pi}^{(r)}$. Otherwise, the previous best is restored from saved checkpoints, unless the drop is small and the datapoints switching between pass and fail indicate evaluation noise.

    \item \textbf{Generate prompt.} The agent distills its diagnostics into a prompt set $\boldsymbol{\pi}^{(r+1)}$ that encodes recovered strategies as explicit guidance. An anti-memorization constraint discourages verbatim reuse of training inputs, gold outputs, or reasoning traces, promoting transferability.

    \item \textbf{Enhanced error attribution.} The agent diagnoses the worst-scoring datapoints of round $r{-}1$, scrutinizing gold reasoning trajectories when available or else the expected outputs. It clusters failures into dynamically chosen categories of \emph{reasoning} errors. In multi-stage systems, pipeline-aware attribution traces inter-module data flow to localize the prompt $\pi_k$ most responsible for each downstream failure, unlike GEPA's round-robin selection.
\end{itemize}

\begin{figure}[t]
\begin{minipage}[t]{0.48\textwidth}
\begin{algorithm}[H]
\caption{\textsc{ReAPO}}
\label{alg:reapo}
\begin{algorithmic}[1]

\Require Agent $\mathcal{A}$ parametrized by $\boldsymbol{\pi}_0$ and
$\mathcal{M}_{\text{sys}}$, training set $\mathcal{D}_{\text{train}}$,
optimizer $\mathcal{M}_{\text{opt}}$, environment $\varepsilon$,
round budget $R$
\Ensure Optimized prompt set $\boldsymbol{\pi}^{*}$

\State $\mathcal{H} \leftarrow \emptyset$ \hspace{3.1em}\textcolor{gray}{\texttt{$\triangleright$ agent trajectory}}
\State $\boldsymbol{\pi}^{(0)} \leftarrow \boldsymbol{\pi}_0$ \hspace{1.5em}\textcolor{gray}{\texttt{$\triangleright$ round-0 prompts}}
\State $\mathcal{S} \leftarrow [\,]$ \hspace{3.05em}\textcolor{gray}{\texttt{$\triangleright$ round train scores}}
\For{$r = 0 \ldots R$}
    \Repeat
        \State $\tau \leftarrow \mathcal{M}_{\text{opt}}.\textsc{Think}(\mathcal{H})$
        \State $(a, \theta) \leftarrow \mathcal{M}_{\text{opt}}.\textsc{Act}(\tau, \mathcal{H})$
        \State $o \leftarrow \varepsilon.\textsc{Step}(a, \theta, r, \mathcal{A}, \mathcal{D}_{\text{train}}, \boldsymbol{\pi}, \mathcal{S})$
        \State $\mathcal{H} \leftarrow \varepsilon.\textsc{Observe}(\mathcal{H}, o)$
    \Until{$a = \textsc{AgentExecution}$}
\EndFor
\State $r^* \leftarrow \arg\max_{r \in \{0,\ldots,R\}} \mathcal{S}[r]$
\State $\boldsymbol{\pi}^* \leftarrow \boldsymbol{\pi}^{(r^*)}$
\State \Return $\boldsymbol{\pi}^*$

\end{algorithmic}
\end{algorithm}
\end{minipage}
\hfill
\begin{minipage}[t]{0.48\textwidth}
\begin{algorithm}[H]
\caption{\textsc{$\varepsilon$.Step}}
\label{alg:env-dispatch}
\begin{algorithmic}[1]

\Require Action $a$, args $\theta$, round $r$,
agent $\mathcal{A}$, data $\mathcal{D}_{\text{train}}$,
prompts $\boldsymbol{\pi}$, scores $\mathcal{S}$
\Ensure Observation $o$

\State \textbf{switch} $a$\textbf{:}
\State \hspace{1em}\textbf{case} \textsc{AgentInspection}\textbf{:}
\State \hspace{2em}$o \leftarrow \varepsilon.\textsc{InspectAgent}(\mathcal{A}, \theta)$
\State \hspace{1em}\textbf{case} \textsc{AgentExecution}\textbf{:}
\State \hspace{2em}$o \leftarrow \varepsilon.\textsc{RunAgent}(\mathcal{A}, \mathcal{D}_{\text{train}}, \theta)$
\State \hspace{2em}$\mathcal{S}[r] \leftarrow o.S$
\State \hspace{1em}\textbf{case} \textsc{AnalyzeFullEvalRun}\textbf{:}
\State \hspace{2em}$o \leftarrow \varepsilon.\textsc{GetFullResults}(\theta)$
\State \hspace{1em}\textbf{case} \textsc{AnalyzeSpecificDatapoint}\textbf{:}
\State \hspace{2em}$o \leftarrow \varepsilon.\textsc{GetDatapointResults}(\theta)$
\State \hspace{1em}\textbf{case} \textsc{PromptEditing}\textbf{:}
\State \hspace{2em}$o \leftarrow \varepsilon.\textsc{UpdatePrompts}(\mathcal{A}, \theta)$
\State \hspace{2em}$\boldsymbol{\pi}^{(r+1)} \leftarrow o.\boldsymbol{\pi}$
\State \hspace{1em}\textbf{case} \textsc{Logging}\textbf{:}
\State \hspace{2em}$o \leftarrow \varepsilon.\textsc{Log}(\theta)$
\State \Return $o$

\end{algorithmic}
\end{algorithm}
\end{minipage}
\caption{Overview of ReAPO. \textbf{Left:} Each round $r = 0, \ldots, R$, $\mathcal{M}_{\text{opt}}$ alternates \textsc{Think}/\textsc{Act} until \textsc{AgentExecution} evaluates $\mathcal{A}$ on $\mathcal{D}_{\text{train}}$. \textbf{Right:} Environment $\varepsilon$ dispatches each tool call.}
\end{figure}

\section{Experiments}
\label{sec:experiments}

\paragraph{Benchmarks.} We evaluate on six public benchmarks spanning multi-hop reasoning, fact verification, mathematics, instruction following, and privacy-conscious delegation: \textit{HotpotQA} \citep{yang2018hotpotqa}, \textit{HoVer} \citep{jiang2020hover}, \textit{AIME~2025} \citep{balunovic2025aime}, \textit{LiveBench-Math} \citep{white2024livebench}, \textit{IFBench} \citep{pyatkin2025ifbench}, and \textit{PUPA} \citep{li2024papillon}. For each benchmark we adopt the same train / validation / test splits and DSPy program definitions as \citet{agrawalGEPAReflectivePrompt2025}, so that all methods optimize the same programs on the same data. Appendix~\ref{app:protocol} defines each benchmark's scoring function and lists the information each optimizer receives. Each method runs through its own released interface, hence the comparisons are between complete optimization procedures rather than information-matched variants of their search protocols. To assess whether conclusions drawn from curated academic suites persist in more realistic settings, we introduce five enterprise agentic benchmarks (Appendix~\ref{app:agents}). Unlike the considered public datasets, these tasks involve ReAct agents equipped with task-specific tools, operating over problems 
that vary considerably in their likely exposure within $\mathcal{M}_{\text{opt}}$'s 
pretraining corpus. These range from widely common tasks, such as \textit{SQL} (natural language to SQL translation and execution) and \textit{Product Questions} (product Q\&A), through more operationally nuanced settings like \textit{Meeting Management} (calendar scheduling with conflict detection), to proprietary workflows including \textit{Customer Support Triage} (routing inquiries) and \textit{Claim History} (insurance claims lookup and appeal recommendation). The spread from common to proprietary matters: as the tool ablation below shows, more capable optimizer models may distill task-specific knowledge into the optimized prompts, effectively transferring prior knowledge of a benchmark into the target system. That suggests an implicit familiarity with benchmark data which public benchmark scores alone may not reveal, and which the enterprise tasks help expose.

\begin{table}[t]
\caption{Test-set scores (\%) on six public benchmarks with GPT-4.1~Mini as the optimizer. Mean of 5~runs evaluated on GPT-4.1~Mini, with 95\% CI in brackets. MIPROv2-ND is MIPROv2 without demonstrations. Best result per benchmark in \textbf{bold}.}
\label{tab:main_mini}
\centering
\small
\setlength{\tabcolsep}{3.5pt}
\begin{tabular}{llcccccc}
\toprule
 & \textbf{Method} & HotpotQA & HoVer & AIME & LiveBench & IFBench & PUPA \\
\midrule
\multirow{3}{*}{\rotatebox[origin=c]{90}{\scriptsize DSPy}}
 & Baseline    & 39.3 {\scriptsize[38.5, 40.1]} & 43.9 {\scriptsize[42.1, 45.8]} & 46.7 {\scriptsize[44.0, 49.3]} & 59.0 {\scriptsize[56.7, 61.2]} & 48.4 {\scriptsize[47.8, 48.8]} & 80.5 {\scriptsize[79.8, 81.4]} \\
 & MIPROv2-ND     & 34.3 {\scriptsize[33.5, 34.9]} & 47.1 {\scriptsize[46.3, 47.6]} & 47.3 {\scriptsize[42.0, 52.7]} & 59.4 {\scriptsize[58.7, 60.2]} & 50.2 {\scriptsize[49.1, 51.0]} & 87.9 {\scriptsize[87.1, 88.7]} \\
 & MIPROv2 & 62.7 {\scriptsize[61.9, 63.3]} & 48.9 {\scriptsize[47.2, 50.4]} & 44.7 {\scriptsize[40.0, 51.3]} & 57.5 {\scriptsize[55.5, 60.0]} & 51.6 {\scriptsize[50.7, 52.5]} & 84.3 {\scriptsize[83.5, 85.2]} \\
\midrule
\multirow{2}{*}{\rotatebox[origin=c]{90}{\scriptsize GEPA}}
 & GEPA        & \textbf{68.5} {\scriptsize[67.5, 69.5]} & 52.3 {\scriptsize[50.6, 53.9]} & 48.0 {\scriptsize[45.3, 51.3]} & 61.2 {\scriptsize[60.0, 62.3]} & 50.8 {\scriptsize[49.7, 52.2]} & 93.0 {\scriptsize[92.1, 93.5]} \\
 & GEPA-Merge  & 67.1 {\scriptsize[66.7, 67.6]} & 51.3 {\scriptsize[49.9, 52.7]} & 48.0 {\scriptsize[45.3, 51.3]} & 60.2 {\scriptsize[59.2, 61.5]} & \textbf{51.7} {\scriptsize[50.9, 52.6]} & \textbf{95.3} {\scriptsize[94.5, 96.0]} \\
\midrule
 & \textbf{ReAPO} & \textbf{68.5} {\scriptsize[67.9, 69.3]} & \textbf{54.3} {\scriptsize[52.3, 55.6]} & \textbf{50.7} {\scriptsize[48.7, 52.7]} & \textbf{61.6} {\scriptsize[59.8, 63.5]} & 50.8 {\scriptsize[50.1, 51.4]} & 91.4 {\scriptsize[90.3, 92.4]} \\
\bottomrule
\end{tabular}
\end{table}

\begin{table}[t]
\caption{Test-set scores (\%) with GPT-5 as optimizer. Mean of 5~runs on GPT-4.1~Mini, with 95\% CI in brackets. Best per benchmark in \textbf{bold}.}
\label{tab:main_gpt5}
\centering
\small
\setlength{\tabcolsep}{4pt}
\begin{tabular}{lcccccc}
\toprule
\textbf{Method} & HotpotQA & HoVer & AIME & LiveBench & IFBench & PUPA \\
\midrule
GEPA                       & \textbf{68.5} {\scriptsize[66.8, 69.6]} & 54.6 {\scriptsize[53.1, 56.1]} & 44.7 {\scriptsize[40.7, 50.7]} & 61.2 {\scriptsize[59.0, 63.3]} & 50.5 {\scriptsize[48.9, 52.1]} & 91.1 {\scriptsize[90.2, 91.9]} \\
\textbf{ReAPO} & 68.2 {\scriptsize[66.9, 69.3]} & \textbf{59.1} {\scriptsize[57.9, 60.3]} & \textbf{54.0} {\scriptsize[51.3, 56.7]} & \textbf{61.8} {\scriptsize[60.1, 63.4]} & \textbf{52.0} {\scriptsize[50.8, 53.3]} & \textbf{93.1} {\scriptsize[92.7, 93.5]} \\
\bottomrule
\end{tabular}
\end{table}

\paragraph{Test performance on public benchmarks.}
\label{sec:test}
Table~\ref{tab:main_mini} reports held-out test scores across the public benchmarks. Following the experimental configuration of \citet{agrawalGEPAReflectivePrompt2025}, we compare three baselines (the unoptimized prompt, MIPROv2 and MIPROv2-No-Demos) against two GEPA configurations (GEPA and GEPA-Merge) and our proposed ReAPO, with both $\mathcal{M}_{\text{opt}}$ and $\mathcal{M}_{\text{sys}}$ instantiated as GPT-4.1~Mini. All competing methods use the optimized prompts provided by the original authors, without re-optimization. For LiveBench-Math, a single-module task on which GEPA and GEPA-Merge are equivalent as methods, we follow the authors' release and use each variant's reported prompt. The table presents averages over five independent runs, with bootstrap confidence intervals reported throughout. ReAPO outperforms all competing methods on HoVer, AIME, and LiveBench-Math, while matching GEPA on HotpotQA and IFBench. Across these five benchmarks, the two methods exhibit largely overlapping confidence intervals, consistent with stochastic noise rather than systematic differences. The only exception is PUPA, on which both GEPA variants outperform ReAPO ($95.3\%$ for GEPA-Merge and $93.0\%$ for GEPA, against $91.4\%$). Decomposing this score into its two components (Appendix~\ref{app:pupa_decomp}) shows that the ReAPO prompt achieves higher answer quality on GPT-4.1~Mini but leaks more private information in the request sent to the external model. Although this prompt contains explicit privacy instructions, GPT-4.1~Mini adheres to them less closely than to the corresponding instructions produced by GEPA. Transferred zero-shot to Qwen3.5-9B, the same prompt leaks considerably less and surpasses GEPA ($95.9\%$ vs.\ $90.5\%$, Table~\ref{tab:transfer}). As shown below, a stronger optimizer removes this weakness even for GPT-4.1~Mini. Tables~\ref{tab:main_gpt5} and~\ref{tab:optimizer_model} pair the same system model with substantially stronger optimizers, GPT-5 and Claude Opus~4.6~\citep{anthropic2026opus46}. With GPT-5, ReAPO achieves the best result with non-overlapping confidence intervals on HoVer, AIME, and PUPA. On LiveBench-Math and IFBench, ReAPO also outperforms GEPA nominally, although the overlapping intervals preclude any definitive conclusion.

To examine whether ReAPO's gains extend beyond proprietary optimizers, we evaluate a fully open-source setting with the same model used for optimization and task execution, $\mathcal{M}_{\text{opt}} = \mathcal{M}_{\text{sys}}$ (Table~\ref{tab:opensource}). We use two Qwen3.5 models spanning different scales and architectures: the dense Qwen3.5-9B and the mixture-of-experts (MoE) Qwen3.5-122B-A10B~\citep{qwen2026qwen35122b}, with 122B total and 10B active parameters. With Qwen3.5-9B, ReAPO consistently improves over the unoptimized baseline and is comparable with GEPA, trailing visibly only on HotpotQA. The resulting prompts perform comparably to those optimized with GPT-4.1~Mini and transferred to Qwen3.5-9B (Table~\ref{tab:transfer}), achieving higher scores on HoVer and AIME. Qwen3.5-9B is additionally effective as an optimizer on two agentic tasks (Appendix~\ref{app:qwen_agents}). With Qwen3.5-122B-A10B, ReAPO outperforms GEPA across the evaluated tasks, with non-overlapping confidence intervals on HotpotQA and AIME. AIME highlights the advantage at both scales: while GEPA yields no improvement over the unoptimized prompt with either model, ReAPO achieves gains of $9.0$ and $4.0$ points with Qwen3.5-9B and Qwen3.5-122B-A10B, respectively. These findings indicate that ReAPO's gains are not contingent on access to large proprietary optimizers and persist across open-source models that differ substantially in scale.

\begin{table}[t]
\caption{Open-source self-optimization. The same Qwen3.5 model acts as the optimizer and system model, for a dense 9B (top) and a MoE 122B-total/10B-active model (bottom). Mean (\%) of 5~runs, with 95\% CI in brackets. Best per row in \textbf{bold}.}
\label{tab:opensource}
\centering
\small
\setlength{\tabcolsep}{5pt}
\begin{tabular}{lccc}
\toprule
\textbf{Benchmark} & Baseline & GEPA & \textbf{ReAPO} \\
\midrule
\multicolumn{4}{l}{\textit{Qwen3.5-9B as optimizer and system model}} \\
HotpotQA      & 60.2 {\scriptsize[59.1, 61.0]} & \textbf{69.7} {\scriptsize[69.3, 70.0]} & 67.1 {\scriptsize[66.7, 67.4]} \\
HoVer         & 36.9 {\scriptsize[35.3, 38.2]} & \textbf{57.3} {\scriptsize[56.4, 58.5]} & 57.1 {\scriptsize[56.1, 58.0]} \\
AIME          & 63.0 {\scriptsize[59.7, 66.3]} & 60.7 {\scriptsize[56.7, 63.3]} & \textbf{72.0} {\scriptsize[66.0, 76.0]} \\
LiveBench     & 63.2 {\scriptsize[62.1, 64.0]} & 70.6 {\scriptsize[69.2, 71.6]} & \textbf{71.6} {\scriptsize[69.3, 73.0]} \\
\midrule
\multicolumn{4}{l}{\textit{Qwen3.5-122B-A10B as optimizer and system model}} \\
HotpotQA      & 60.2 {\scriptsize[59.4, 61.2]} & 69.1 {\scriptsize[68.4, 70.1]} & \textbf{72.3} {\scriptsize[71.1, 73.4]} \\
HoVer         & 45.4 {\scriptsize[44.5, 46.3]} & 58.5 {\scriptsize[57.1, 59.7]} & \textbf{60.4} {\scriptsize[59.2, 61.7]} \\
AIME          & 78.7 {\scriptsize[77.3, 80.0]} & 76.0 {\scriptsize[74.0, 78.0]} & \textbf{82.7} {\scriptsize[81.3, 83.3]} \\
\bottomrule
\end{tabular}
\end{table}

\begin{figure}[t]
\centering
\includegraphics[width=0.9\linewidth]{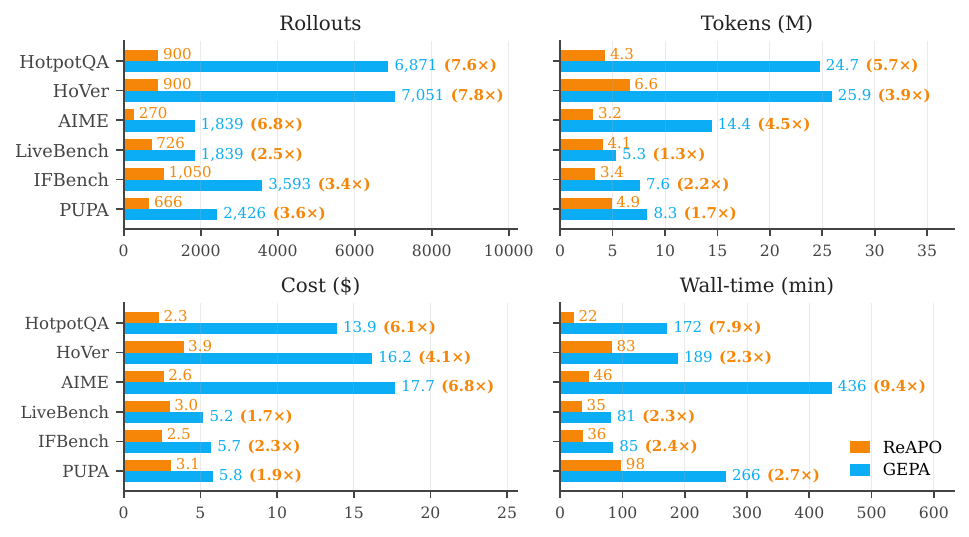}
\caption{Optimization cost across six public benchmarks. Rollouts, tokens, and API cost with GPT-4.1~Mini, and wall-time to the selected prompt with GPT-5. Bar labels report values, with bold GEPA/ReAPO ratios. API costs are recomputed from logged input/output tokens at October 2026 GPT-4.1~Mini list prices, excluding cached-input discounts, superseding GEPA's released dollar costs.}
\label{fig:cost}
\end{figure}

\begin{table}[t]
\caption{Cross-model prompt transfer. Prompts optimized on GPT-4.1~Mini are applied zero-shot to Qwen3.5-9B. Mean (\%) over 5~runs (10 for AIME, given its wide CIs and small dataset), with 95\% CI in brackets. Best per benchmark in \textbf{bold}.}
\label{tab:transfer}
\centering
\small
\setlength{\tabcolsep}{4pt}
\begin{tabular}{lcccccc}
\toprule
\textbf{Method} & HotpotQA & HoVer & AIME & LiveBench & IFBench & PUPA \\
\midrule
Baseline       & 60.2 {\scriptsize[59.1, 61.0]} & 36.9 {\scriptsize[35.3, 38.2]} & 63.0 {\scriptsize[59.7, 66.3]} & 63.2 {\scriptsize[62.1, 64.0]} & 61.1 {\scriptsize[60.0, 62.1]} & 83.7 {\scriptsize[82.9, 84.7]} \\
GEPA           & 67.9 {\scriptsize[66.8, 69.3]} & 49.9 {\scriptsize[47.5, 51.7]} & 63.0 {\scriptsize[58.3, 68.0]} & 70.5 {\scriptsize[69.5, 71.1]} & 63.6 {\scriptsize[61.9, 65.6]} & 90.5 {\scriptsize[90.1, 90.8]} \\
\textbf{ReAPO} & \textbf{68.9} {\scriptsize[68.4, 69.6]} & \textbf{53.5} {\scriptsize[52.5, 54.3]} & \textbf{65.0} {\scriptsize[61.7, 68.0]} & \textbf{71.4} {\scriptsize[70.7, 72.3]} & \textbf{63.7} {\scriptsize[61.2, 66.1]} & \textbf{95.9} {\scriptsize[95.3, 96.7]} \\
\bottomrule
\end{tabular}
\end{table}

\paragraph{Rollout and cost efficiency.}
\label{sec:rollouts}
A central advantage of ReAPO is its sample efficiency, as evidenced by Figures~\ref{fig:progression} and \ref{fig:cost}. By design,
its rollout cost is deterministic and tied directly to the round
budget: ReAPO consumes at most $(R+1) \times |\mathcal{D}_{\text{train}}|$
agent-on-datapoint evaluations for $R$ optimization rounds. We use
$R = 5$ as the default across benchmarks, but for IFBench where we set $R = 6$. In either case, the cost is set
\emph{a priori} and is independent of the optimizer model. GEPA, by contrast, consumes a fixed number of rollouts over the training and validation sets, identical to that of MIPROv2 \citep{agrawalGEPAReflectivePrompt2025}. ReAPO requires $2.5\times$ (LiveBench) to $7.8\times$ (HoVer) fewer rollouts (Figure~\ref{fig:cost}, top left), with the largest absolute savings on the two multi-hop reasoning benchmarks.

Since rollout counts do not reflect the calls issued by the optimizer itself, Figure~\ref{fig:cost} also reports the token usage and API cost of the optimization phase, with GPT-4.1~Mini serving as both optimizer and system model, together with the wall-clock time required to obtain the final prompt when GPT-5 is the optimizer. Depending on the benchmark, GEPA consumes $1.3$--$5.7\times$ more tokens and incurs $1.7$--$6.8\times$ higher API cost than ReAPO, for a total of \$64.5 versus \$17.4. Its optimization also takes $2.3$--$9.4\times$ longer, or $20.5$ versus $5.3$ hours in total (Appendix~\ref{app:cost_details}). These additional resources rarely translate into higher test scores, as the two methods perform comparably on the six public benchmarks, with GEPA meaningfully ahead only on PUPA under GPT-4.1~Mini (Tables~\ref{tab:main_mini} and~\ref{tab:main_gpt5}). In our experiments, ReAPO thus reaches this level of performance at roughly a quarter of GEPA's optimization cost and time.

\paragraph{Cross-model prompt transfer.}
\label{sec:cross-model}
A central practical question for any prompt optimization method is
whether the resulting prompts retain their value when deployed on other models, where
re-running the full optimization for each model is costly. To study this, we apply the prompts that ReAPO and GEPA optimize on GPT-4.1~Mini (cf.\ Table~\ref{tab:main_mini}) zero-shot to Qwen3.5-9B. Table~\ref{tab:transfer} compares the resulting test scores with those of the unoptimized prompt on the same model.

Across the transfer experiment, both methods retain meaningful gains on Qwen3.5-9B. For both optimizers, the largest absolute gain
is on HoVer, the multi-hop reasoning task on which the unoptimized baseline is weakest. Head-to-head, ReAPO and GEPA exhibit overlapping confidence intervals on four out of six benchmarks. Differences emerge on HoVer, where ReAPO outperforms GEPA by $+3.6$ points with non-overlapping confidence intervals, and on PUPA, where the
picture reverses relative to the same-model results: ReAPO reaches $95.9$\% on Qwen3.5-9B against GEPA's $90.5$\%, whereas GEPA
led on GPT-4.1~Mini ($93.0$\% vs.\ $91.4$\%). The ReAPO prompt explicitly separates PUPA's two
objectives: it imposes a strict privacy boundary on the request sent
to the external model, while instructing the responder to preserve the
user's intent and answer utility. GEPA's prompt instead entangles
anonymization with task-fidelity instructions, leaving the
privacy-utility trade-off to the system model. GPT-4.1~Mini resolves this trade-off in GEPA's favor (Table~\ref{tab:pupa_decomp}), owing to higher external-request privacy ($93.9$\% vs.\ $88.9$\%) despite lower answer quality ($92.0$\% vs.\ $93.8$\%). Qwen3.5-9B follows
ReAPO's explicit privacy hierarchy more reliably, giving ReAPO both
higher privacy ($96.1$\% vs.\ $86.9$\%) and higher quality ($95.8$\%
vs.\ $94.0$\%). The gain from GEPA's instructions on PUPA is thus specific to GPT-4.1~Mini, whereas the privacy-first structure of the ReAPO prompt carries over to Qwen3.5-9B.
After transfer, ReAPO therefore remains competitive with GEPA on the considered benchmarks, despite having used $2.5$--$7.8\times$ fewer rollouts during optimization.

\paragraph{Agent benchmark results.}
Table~\ref{tab:agents} reports head-to-head test-set scores between
ReAPO and GEPA across the five enterprise agentic tasks. Both
methods use GPT-5 as optimizer and evaluation model under
approximately matched rollout budgets, isolating the effect of the
optimization procedure from its sample efficiency. For GEPA, the
training split also serves as its validation set, so the Pareto
archive is maintained directly on the same datapoints used for
per-sample reflection. On Customer Support Triage, neither approach was able to replace the baseline (${\sim}75.5$\%), as prompt modifications to the multi-agent router consistently destabilized the tool-calling control loop, forcing both optimizers to revert. Both therefore returned the unchanged baseline prompt, and Table~\ref{tab:agents} reports its evaluation for all three methods.  Of the four remaining tasks, ReAPO outperforms GEPA
with non-overlapping confidence intervals on three: Product
Questions ($+19.3$), Meeting Management
($+15.9$), and SQL ($+4.1$). The Product Questions result is particularly noteworthy,
as ReAPO improves the agent by $+8.9$ points over the unoptimized
baseline while GEPA's prompts \emph{regress} below it by $-10.4$
points. GEPA, in turn, outperforms ReAPO on
Claim History ($-10.1$), although ReAPO still improves on the baseline by $+22.9$ points. The gap traces back to an early revision that instructed the agent to return its answer as plain text rather than through the runtime's submission tool, so the affected episodes end without a scored answer. Since ReAPO refines a single prompt lineage, every later revision inherited this instruction, whereas a population-based search like GEPA can recover by continuing from another candidate (Section~\ref{sec:limitations}, Appendix~\ref{app:failures}). Overall, ReAPO outperforms GEPA on three of the four tasks on which an optimizer modified the baseline prompt, suggesting that the trends observed on the public benchmarks carry over to enterprise multi-turn, tool-calling settings.

 \begin{table}[t]                                                                                                                        
  \caption{Agent benchmark results. GPT-5 as both optimizer and evaluation model. Test-set scores (\%) are the mean over 5~independent trials, with 95\% CI in brackets. On Customer Support neither optimizer changed the baseline prompt, so all three rows report the same evaluation. Best in \textbf{bold}.}                                                           
  \label{tab:agents}                                                                                                                      
  \centering                                                                                                                              
  \small                                                                                                                                
  \setlength{\tabcolsep}{3pt}
  \begin{tabular}{lccccc}
  \toprule
  \textbf{Method} & Claim History & Customer Support & Meeting Management & Product Qs & SQL \\
  \midrule
  Baseline       & 57.3 {\scriptsize[55.3, 59.5]} & 75.5 {\scriptsize[74.8, 76.4]} & 44.0 {\scriptsize[39.2, 48.3]} & 83.7 {\scriptsize[80.0, 86.7]} & 69.8 {\scriptsize[67.8, 71.4]} \\
  GEPA           & \textbf{90.3} {\scriptsize[86.1, 94.4]} & 75.5 {\scriptsize[74.8, 76.4]} & 70.0 {\scriptsize[65.6, 74.3]} & 73.3 {\scriptsize[65.9, 78.5]} & 80.8 {\scriptsize[78.0, 83.6]} \\
  \textbf{ReAPO} & 80.2 {\scriptsize[76.5, 82.9]} & 75.5 {\scriptsize[74.8, 76.4]} & \textbf{85.9} {\scriptsize[81.0, 88.5]} & \textbf{92.6} {\scriptsize[90.4, 94.8]} & \textbf{84.9} {\scriptsize[84.1, 85.5]} \\
  \bottomrule
  \end{tabular}
  \end{table}  

\paragraph{Tool ablation.}
\label{para:tools_ablation}
To quantify the contribution of ReAPO's introspection 
capabilities, we ablate the \textit{Results Analysis} tool category 
under three progressively restricted configurations. \textit{Full} 
provides the complete tool suite described in Section~\ref{sec:method}. 
\textit{Score-Only} strips all \textit{Results Analysis} tools, leaving 
the optimizer with only the aggregate score after each evaluation round, 
and no access to per-datapoint breakdowns, execution traces, or failure 
diagnostics. \textit{Macro} occupies an intermediate point: it reports each datapoint's score and the evaluator's feedback on failures, but hides the datapoint inputs and the agent's execution traces. We restrict the ablation to \textit{Results Analysis}: \textit{Agent Execution} and \textit{Prompt Editing} are structurally essential to the loop, and the remaining categories provide auxiliary functionality whose removal would confound the comparison. \textit{Results Analysis} 
tools, by contrast, directly mediate the optimizer's capacity to diagnose 
failures and attribute errors across pipeline modules, constituting the 
core capability that distinguishes ReAPO from a purely 
score-chasing procedure. The ablation is conducted under two optimizer configurations (GPT-4.1~Mini 
and GPT-5 as $\mathcal{M}_{\text{opt}}$, with $\mathcal{M}_{\text{sys}}$ fixed to 
GPT-4.1~Mini) to disentangle tool utility from optimizer 
capability.

\begin{table}[t]
\caption{Ablation of the \textit{Results Analysis} tools. Test-set scores (\%) on GPT-4.1~Mini. \textit{Full} provides all introspection tools, \textit{Score-Only} only aggregate scores, and \textit{Macro} per-datapoint scores with failure feedback. Mean of 5~runs, with 95\% CI in brackets. Best per optimizer in \textbf{bold}. On HotpotQA and LiveBench, GPT-4.1~Mini \textit{Score-Only} retained the unoptimized prompt.}
\label{tab:tool_ablation}
\centering
\small
\setlength{\tabcolsep}{3pt}
\begin{tabular}{lcccccc}
\toprule
 & \multicolumn{3}{c}{\textsc{GPT-5 Optimizer}} & \multicolumn{3}{c}{\textsc{GPT-4.1~Mini Optimizer}} \\
\cmidrule(lr){2-4} \cmidrule(lr){5-7}
\textbf{Bench.} & Full & Score-Only & Macro & Full & Score-Only & Macro \\
\midrule
HotpotQA  & \textbf{68.2} {\scriptsize[66.9, 69.3]} & 66.8 {\scriptsize[65.3, 68.1]} & 66.6 {\scriptsize[65.4, 67.8]} & \textbf{68.5} {\scriptsize[67.9, 69.3]} & 39.3 {\scriptsize[38.5, 40.1]} & 36.1 {\scriptsize[35.0, 37.1]} \\
HoVer     & \textbf{59.1} {\scriptsize[57.9, 60.3]} & 58.5 {\scriptsize[57.3, 59.8]} & 55.7 {\scriptsize[54.8, 56.5]} & \textbf{54.3} {\scriptsize[52.3, 55.6]} & 51.9 {\scriptsize[50.7, 53.1]} & 52.6 {\scriptsize[51.2, 53.7]} \\
IFBench   & 52.0 {\scriptsize[50.8, 53.3]} & 51.0 {\scriptsize[49.6, 52.3]} & \textbf{53.3} {\scriptsize[51.5, 54.5]} & 50.8 {\scriptsize[50.1, 51.4]} & 50.6 {\scriptsize[49.5, 51.8]} & \textbf{51.1} {\scriptsize[49.8, 52.2]} \\
LiveBench & \textbf{61.8} {\scriptsize[60.1, 63.4]} & 59.8 {\scriptsize[58.5, 61.1]} & 59.5 {\scriptsize[58.0, 61.0]} & \textbf{61.6} {\scriptsize[59.8, 63.5]} & 59.0 {\scriptsize[56.7, 61.2]} & 59.1 {\scriptsize[57.7, 60.4]} \\
\bottomrule
\end{tabular}
\end{table}

\begin{table}[t]
\caption{Tool ablation on agents. GPT-5 optimizer, GPT-4.1~Mini system model. 95\% CI in brackets.}
\label{tab:tool_agents}
\centering
\small
\setlength{\tabcolsep}{3pt}
\begin{tabular}{lccccc}
\toprule
\textbf{Method} & Claim History & Customer\ Support & Meeting Management & Product Qs & SQL \\
\midrule
Baseline    & 66.6 {\scriptsize[58.9, 71.8]} & \textbf{79.8} {\scriptsize[78.3, 81.3]} & 51.4 {\scriptsize[49.4, 53.3]} & 73.0 {\scriptsize[66.3, 76.7]} & 70.7 {\scriptsize[68.4, 72.9]} \\
Score-Only  & 90.2 {\scriptsize[87.4, 92.6]} & 72.3 {\scriptsize[72.0, 72.7]} & 36.1 {\scriptsize[34.1, 38.7]} & 76.7 {\scriptsize[76.7, 76.7]} & 76.5 {\scriptsize[75.5, 77.5]} \\
Full        & \textbf{97.8} {\scriptsize[97.6, 97.9]} & 76.3 {\scriptsize[73.4, 79.0]} & \textbf{71.1} {\scriptsize[68.0, 75.3]} & \textbf{77.5} {\scriptsize[76.7, 78.2]} & \textbf{77.1} {\scriptsize[73.8, 80.1]} \\
\bottomrule
\end{tabular}
\end{table}

Under GPT-4.1~Mini optimization, the \textit{Results Analysis} tools prove critical (Table~\ref{tab:tool_ablation}). Once they are withheld, the original skill prompt $\sigma$ still directs the optimizer to call them, and GPT-4.1~Mini repeats these calls indefinitely. When removing the tool references from the skill prompt, the ablation shows degradation: HotpotQA \textit{Full} ($68.5$\%) collapses to \textit{Score-Only} ($39.3$\%), matching the baseline, and \textit{Macro} falls to $36.1$\%. Similarly, LiveBench drops from $61.6$\% to $59.0$\%. These results indicate that introspection tools are essential when the optimizer lacks strong internalized task knowledge. Surprisingly, this degradation largely vanishes under the GPT-5 optimizer setup. Under GPT-5 the largest public-benchmark gap is $3.4$ points (HoVer, \textit{Full} vs.\ \textit{Macro}), and both ablated variants stay above the unoptimized baseline on every benchmark. These results suggest that a frontier-class optimizer can largely 
compensate for the absence of diagnostic feedback on public 
benchmarks, distilling its latent task knowledge into the target model's prompts without explicit error analysis.

We test our distillation hypothesis by extending the ablation to the enterprise agentic tasks (Table~\ref{tab:tool_agents}). On SQL and Product Questions, two common task types that are likely well represented in pretraining data, restricting the GPT-5 optimizer to aggregate scores costs less than one point, comparable to the differences on the public benchmarks. On the more specialized and proprietary workflows, the same ablation leads to larger losses, of up to $35.0$ points on Meeting Management. This pattern is consistent with our hypothesis: when the optimizer can draw on prior knowledge of a problem, aggregate scores largely suffice to guide the optimization. On less familiar tasks, however, the per-datapoint feedback and execution traces provided by the \textit{Results Analysis} tools become considerably more valuable.

\paragraph{Skill prompt ablation.}
We isolate the contribution of the skill prompt by ablating its analytical rules. The default
$\sigma_\emph{Full}$ contains the error-analysis, attribution, and noise-handling guidance of
Section~\ref{sec:method}, applied to up to $N_w = 15$ worst-scoring datapoints. The
$\sigma_\emph{Trimmed}$ variant removes this guidance and narrows the window to $N_w = 5$, which steers
the optimizer toward instance-specific patches rather than patterns that generalize across the data
distribution. Both variants were run with GPT-5 as $\mathcal{M}_{\text{opt}}$ and GPT-4.1~Mini
as $\mathcal{M}_{\text{sys}}$ (refer to Table~\ref{tab:skill}).

Skill prompt $\sigma_\emph{Full}$ outperforms $\sigma_\emph{Trimmed}$ on almost every benchmark, with gains ranging
from $+2.8$ (LiveBench) to $+7.4$ (HoVer). The only exception is
IFBench, where the overlapping confidence intervals leave the two
variants indistinguishable. The general
direction of these gaps indicates that the protocol components
contribute additively across reasoning, mathematical, and
instruction-following tasks. On AIME, the
gold-solution-comparison component directs the optimizer to
extract generalizable problem-solving patterns (recurring
formulas, structural approaches) from provided solutions rather than copying specific answers, a regime particularly suited to
mathematical reasoning where each datapoint admits a transferable
technique. A complementary signal emerges from the tools ablation: on HotpotQA and HoVer, $\sigma_\emph{Trimmed}$ scores below \textit{Score-Only}, suggesting that narrow failure analysis without the pipeline-awareness rule can hurt multi-hop performance.

\section{Limitations}
\label{sec:limitations}

\paragraph{Protocol adherence.} The protocol is a natural-language specification whose execution is delegated to the optimizer model. Adherence is not enforced, and weaker models might deviate from the prescribed steps or stop before the round budget is exhausted. The budget is fixed in advance rather than adapted to the task, so tasks may be under- or over-optimized. A protocol that monitors its own progress and stops once the training score plateaus is a natural direction for future work.

\paragraph{Single lineage.} ReAPO trades exploration breadth for sample efficiency. A revision accepted early propagates to every later round, and checkpointing can roll back but cannot recombine complementary candidates as a population does. Appendix~\ref{app:failures} examines three such instances: an early revision on Claim History that later rounds did not undo, the loss of complementary per-example gains on PUPA, and Customer Support Triage, where neither ReAPO nor GEPA improves on the baseline.

\paragraph{Resource measurements.} Wall-clock times were measured from a single orchestration machine and depend on provider latency and throughput during each run. Costs are computed from logged token counts at list prices. Since the GEPA artifact does not record cached-input tokens, both methods are priced without cache discounts (Appendix~\ref{app:cost_details} presents discounted costs). The rollout counts, which depend on neither provider conditions nor pricing, remain the most durable measure of efficiency.

\begin{table}[t]
\caption{Skill ablation. GPT-5 optimizer, GPT-4.1~Mini system model. 95\% CI in brackets.}
\label{tab:skill}
\centering
\small
\setlength{\tabcolsep}{4pt}
\begin{tabular}{lccccc}
\toprule
\textbf{Method} & HotpotQA & HoVer & AIME & LiveBench & IFBench \\
\midrule
$\sigma_{Trimmed}$        & 63.6 {\scriptsize[62.4, 64.9]} & 51.7 {\scriptsize[50.9, 52.5]} & 50.0 {\scriptsize[50.0, 50.0]} & 59.0 {\scriptsize[56.7, 61.2]} & \textbf{52.2} {\scriptsize[50.8, 53.7]} \\
$\boldsymbol{\sigma_{Full}}$ & \textbf{68.2} {\scriptsize[66.9, 69.3]} & \textbf{59.1} {\scriptsize[57.9, 60.3]} & \textbf{54.0} {\scriptsize[51.3, 56.7]} & \textbf{61.8} {\scriptsize[60.1, 63.4]} & 52.0 {\scriptsize[50.8, 53.3]} \\
\bottomrule
\end{tabular}
\end{table}

\section{Conclusion}
\label{sec:conclusion}

We presented \textbf{ReAPO}, a prompt optimizer that replaces population-based machinery with a single ReAct agent driven by a uniform natural-language protocol and a reusable set of tools. Across six public benchmarks, ReAPO matches or surpasses GEPA and MIPROv2 with $2$--$7\times$ fewer evaluation rollouts and lower optimization cost, and it outperforms GEPA on three of five enterprise tasks. The prompts transfer across model families, and the approach remains effective when driven by a 9B open-weight model. Multi-run analysis shows that gains from a single run can fall within stochastic variance, which calls for confidence intervals when ranking prompt optimizers. The tool ablation shows that introspection is indispensable on enterprise tasks outside the optimizer's pretraining corpus. Together, these findings argue that prompt optimization need not rely on elaborate search protocols: as models grow more capable at analysis, agentic simplicity is enough.

\begin{ack}
We are grateful to Robert Sorete for important contributions to our experiments. This work was funded by UiPath, which employs all authors and develops commercial agent products. No other funding was received. The authors declare no other competing interests.
\end{ack}

\bibliographystyle{plainnat}
\bibliography{refs}

\appendix

\section{Results with a GPT-4.1~Mini optimizer}
\label{app:gpt41mini}

\begin{figure}[h]
\centering
\includegraphics[width=\linewidth]{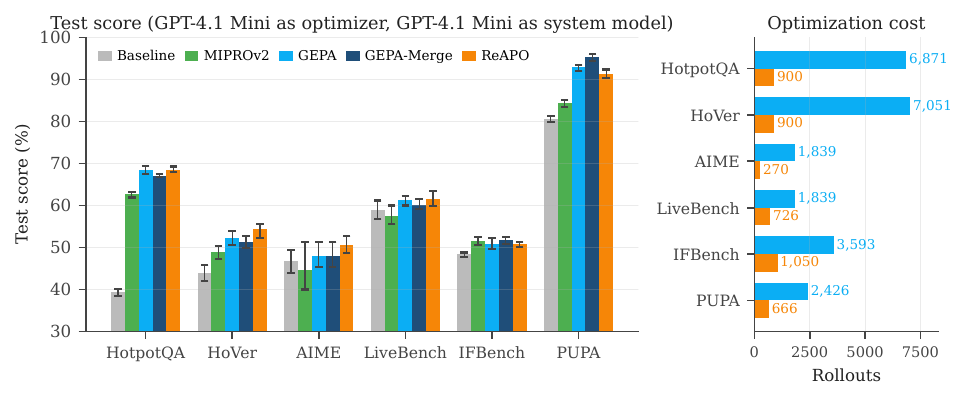}
\caption{Same protocol as Figure~\ref{fig:progression} with GPT-4.1~Mini as both optimizer and system model, extended to MIPROv2 and GEPA-Merge. \textbf{Left:} Test scores with 95\% CIs over five runs. \textbf{Right:} Optimization cost in evaluation rollouts, with MIPROv2, GEPA and GEPA-Merge on the same budget.}
\label{fig:gpt41mini}
\end{figure}

The headline figure in the main text reports test scores under a GPT-5 optimizer. Figure~\ref{fig:gpt41mini} repeats the same protocol with GPT-4.1~Mini as both $\mathcal{M}_{\text{opt}}$ and $\mathcal{M}_{\text{sys}}$, holding everything else fixed.

\section{Implementation Details}
\label{sec:impl}

ReAPO is implemented as a LangGraph ReAct agent with several engineering adaptations for long-horizon optimization:

\begin{itemize}
    \item \textbf{Context management.} When the conversation exceeds a context budget (600K characters, 400K for local models), a \texttt{pre\_model\_hook} prunes older messages while preserving the system prompt, skill instructions, and recent evaluation results. This enables optimization runs spanning $>$30 tool calls without context overflow.

    \item \textbf{Anti-termination guard.} A \texttt{post\_model\_hook} detects when the model attempts to terminate prematurely (producing text without tool calls before completing all requested rounds) and appends a continuation instruction.

    \item \textbf{Version tracking.} Each prompt modification is tagged with a round identifier, enabling deterministic rollback via saved checkpoints. The optimizer writes prompts to a log directory before applying them, creating an immutable audit trail.
\end{itemize}

The optimizer is model-agnostic: every optimizer model uses the same skill prompt $\sigma$.

\paragraph{Target program wrapping.} On the public benchmarks, the systems being optimized are the same DSPy programs released with GEPA's artifact\footnote{\url{https://github.com/gepa-ai/gepa-artifact}} (e.g.\ \texttt{HoverMultiHop}, \texttt{HotpotQAProgram}), run as-is rather than reimplemented. HotpotQA and HoVer are four-predictor retrieval pipelines, IFBench and PUPA have two predictors, and AIME and LiveBench-Math a single one. In every case the optimization target is each predictor's instruction, and a selection of the prompts is listed in Appendix~\ref{app:all_prompts}. Each program is wrapped into a LangGraph node.

\paragraph{Optimizer environment.} The optimizer agent communicates with the gym through an MCP server that exposes the five tool categories of Section~\ref{sec:method} as JSON-schemed function calls. Table~\ref{table:tools_detail} groups the optimizer's MCP tools into five high-level
categories, listing each
individual tool exposed to the ReAPO ReAct loop alongside a short
description of its functionality. Each evaluation run streams OpenTelemetry spans into a per-experiment workspace, from which the Results Analysis tools reconstruct aggregate scores, per-datapoint pass/fail breakdowns, and execution traces.

\begin{table}[t]
\centering
\caption{Per-tool functionality of the MCP tools exposed to the ReAct optimizer agent.}
\label{table:tools_detail}
\small
\setlength{\tabcolsep}{6pt}
\begin{tabularx}{\linewidth}{@{}llX@{}}
\toprule
\textbf{Tool} & \textbf{Category} & \textbf{Functionality} \\
\midrule
\texttt{get\_agent\_files}      & Inspection & List the files that constitute an agent's directory. \\
\texttt{read\_agent\_file}      & Inspection & Read the raw contents of any agent file (e.g.\ \texttt{agent.yaml}). \\
\texttt{summarize\_agent}       & Inspection & Produce a concise summary of input/output schemas, prompts, and tools. \\
\texttt{inspect\_agent}         & Inspection & Return detailed agent metadata (full schemas, tags, evaluators). \\
\texttt{inspect\_tools}         & Inspection & Enumerate the tools attached to an agent with their signatures. \\
\texttt{get\_prompt}            & Inspection & Read a single named prompt (e.g.\ \texttt{system}, \texttt{user}). \\
\addlinespace
\texttt{update\_prompt}         & Editing & Replace an agent prompt with a candidate revision, tagged with a round label. \\
\addlinespace
\texttt{run\_agent\_async}      & Execution & Launch an asynchronous evaluation of the agent on $\mathcal{D}_{\text{train}}$. \\
\texttt{get\_run\_status}       & Execution & Poll the status of an in-progress run. \\
\texttt{wait\_for\_run}         & Execution & Block until an asynchronous run completes. \\
\addlinespace
\texttt{check\_evals}           & Analysis & Inspect the evaluators and evaluation sets attached to an agent. \\
\texttt{get\_datapoint}         & Analysis & Retrieve a single datapoint, including gold expected behavior. \\
\texttt{list\_experiments}      & Analysis & Enumerate completed and in-progress experiments, e.g.\ when resuming a run. \\
\texttt{get\_experiment\_results} & Analysis & Return per-datapoint scores and details for a specific experi\-ment/agent. \\
\texttt{get\_full\_results}     & Analysis & Aggregate score plus a per-datapoint pass/fail breakdown. \\
\texttt{get\_agent\_messages}   & Analysis & Recover the full LLM message history (turns, tool calls, results) of a run. \\
\addlinespace
\texttt{save\_optimizer\_log}   & Logging & Persist a file (prompt snapshot, score summary) in the optimizer workspace. \\
\texttt{read\_optimizer\_log}   & Logging & Read back a previously saved optimizer log entry. \\
\texttt{get\_usage\_stats}      & Logging & Report cumulative rollouts and prompt updates consumed so far. \\
\texttt{reset\_usage\_stats}    & Logging & Reset the usage tracker for an agent. \\
\bottomrule
\end{tabularx}
\end{table}

\section{Qwen3.5-9B on enterprise agentic tasks}
\label{app:qwen_agents}

Table~\ref{tab:qwen_agents} complements the open-source results of Table~\ref{tab:opensource} with two enterprise agentic tasks, using Qwen3.5-9B as the system model. Both optimized prompts improve on the baseline, by 41 to 48 points on Claim History and by approximately 25 points on Product Questions. The prompt optimized by Qwen3.5-9B itself matches the one optimized by GPT-5 within the wide intervals of both tasks. A dense 9B model thus suffices as a tool-capable ReAPO optimizer on these tasks.

\begin{table}[h]
\caption{Enterprise agentic tasks with Qwen3.5-9B as system model (\%). Subscripts denote the optimizer of the ReAPO prompt. Mean of 5~runs, with 95\% CI in brackets. Best per row in \textbf{bold}.}
\label{tab:qwen_agents}
\centering
\small
\setlength{\tabcolsep}{5pt}
\begin{tabular}{lccc}
\toprule
\textbf{Agentic task} & Baseline & ReAPO$_{\text{GPT-5}}$ & \textbf{ReAPO}$_{\text{Qwen3.5-9B}}$ \\
\midrule
Claim History     & 46.0 {\scriptsize[41.1, 50.9]} & 87.4 {\scriptsize[83.8, 91.6]} & \textbf{93.7} {\scriptsize[91.6, 95.8]} \\
Product Questions & 63.0 {\scriptsize[58.5, 67.4]} & \textbf{88.6} {\scriptsize[81.9, 95.2]} & 87.8 {\scriptsize[79.7, 95.2]} \\
\bottomrule
\end{tabular}
\end{table}

\section{Enterprise results with a forced final answer}
\label{app:agents_forced}

Table~\ref{tab:agents_forced} re-evaluates the methods of Table~\ref{tab:agents} under a protocol in which every episode ends with a scored answer. Once an agent reaches a task-specific call limit, the loop requires a final tool call carrying the agent's structured output. The limit is the 95th percentile of model calls in successful episodes run with a 30-call budget. The protocol is identical for all methods. Relative to Table~\ref{tab:agents}, it mostly lifts prompts whose episodes would otherwise not terminate, such as the baseline and ReAPO on Claim History and GEPA on Product Questions. ReAPO leads GEPA on all four tasks where an optimizer changed the baseline prompt, with non-overlapping confidence intervals on Meeting Management, Product Questions and SQL.

\begin{table}[h]
\caption{Agent benchmark results (\%) with a forced final tool call at a task-specific call limit. GPT-5 as both optimizer and evaluation model. Scores are the mean over 5~independent trials, with 95\% CI in brackets. On Customer Support neither optimizer changed the baseline prompt, so all three rows report the same evaluation. Best in \textbf{bold}.}
\label{tab:agents_forced}
\centering
\small
\setlength{\tabcolsep}{3pt}
\begin{tabular}{lccccc}
\toprule
\textbf{Method} & Claim History & Customer\ Support & Meeting Management & Product Qs & SQL \\
\midrule
Baseline       & 62.9 {\scriptsize[60.7, 64.4]} & 74.9 {\scriptsize[72.5, 77.0]} & 51.0 {\scriptsize[42.5, 56.9]} & 84.4 {\scriptsize[80.7, 88.1]} & 73.8 {\scriptsize[72.4, 75.3]} \\
GEPA           & 89.6 {\scriptsize[86.4, 93.2]} & 74.9 {\scriptsize[72.5, 77.0]} & 72.1 {\scriptsize[67.9, 77.4]} & 85.2 {\scriptsize[81.5, 88.1]} & 80.4 {\scriptsize[79.2, 81.7]} \\
\textbf{ReAPO} & \textbf{94.4} {\scriptsize[92.6, 96.0]} & 74.9 {\scriptsize[72.5, 77.0]} & \textbf{92.3} {\scriptsize[91.8, 92.8]} & \textbf{91.8} {\scriptsize[90.4, 92.6]} & \textbf{84.5} {\scriptsize[82.8, 86.3]} \\
\bottomrule
\end{tabular}
\end{table}

\section{Optimization cost: exact values}
\label{app:cost_details}

Tables~\ref{tab:cost_details} and~\ref{tab:walltime} list the exact values behind the four panels of Figure~\ref{fig:cost}. Table~\ref{tab:cost_details} covers the GPT-4.1~Mini runs, with rollout budgets, the input/output token split and cost, and Table~\ref{tab:walltime} the wall-time of the GPT-5 runs. GEPA's token overhead is mostly input tokens on HotpotQA and HoVer, whereas on AIME it is dominated by output tokens ($9.9$M vs.\ $1.1$M for ReAPO). On LiveBench, GEPA's input volume is slightly below ReAPO's and the gap comes from output tokens.

\paragraph{Measurement.} Tokens and cost cover the optimization phase only (optimizer, reflection, and system-model calls), excluding the held-out test evaluation and its judge calls. GEPA's token counts come from the optimization logs released with its artifact. For both methods, cost is recomputed from these input and output tokens at GPT-4.1~Mini list prices as of October 2026 (\$0.40 per million input tokens and \$1.60 per million output tokens), without cached-input discounts, so that both are priced by the same rule. Table~\ref{tab:cost_details} also lists ReAPO's cost with its cached input charged at \$0.10 per million tokens, which lowers it by 16--26\%. Wall-time (Table~\ref{tab:walltime}) is the end-to-end time to the selected prompt, taken from the logs of the GPT-5 optimization runs, which were orchestrated from the same machine (Apple M2 Pro, 32\,GB RAM) with 50 concurrent rollouts for both methods. This measure does not control per-call latency or cloud-service conditions.

\begin{table}[t]
\caption{Optimization-phase rollouts, tokens and cost per benchmark, with GPT-4.1~Mini as optimizer and system model. Rollouts: optimization budgets (GEPA's matched to MIPROv2's, ReAPO's $(R{+}1)\times|\mathcal{D}_{\text{train}}|$). Tokens in millions, cost in USD at list prices without cached-input discounts. Values in parentheses give ReAPO's cost with cached input at the discounted rate.}
\label{tab:cost_details}
\centering
\small
\setlength{\tabcolsep}{4pt}
\begin{tabular}{llrrrrr}
\toprule
 & & & \multicolumn{3}{c}{Tokens (M)} & \\
\cmidrule(lr){4-6}
\textbf{Benchmark} & \textbf{Method} & Rollouts & In & Out & Total & Cost (\$) \\
\midrule
\multirow{2}{*}{HotpotQA} & GEPA & 6,871 & 21.43 & 3.31 & 24.74 & 13.9 \\
 & \textbf{ReAPO} & 900 & 3.82 & 0.47 & 4.30 & 2.3 {\scriptsize(1.9)} \\
\multirow{2}{*}{HoVer} & GEPA & 7,051 & 21.06 & 4.84 & 25.90 & 16.2 \\
 & \textbf{ReAPO} & 900 & 5.56 & 1.07 & 6.63 & 3.9 {\scriptsize(3.2)} \\
\multirow{2}{*}{AIME} & GEPA & 1,839 & 4.52 & 9.92 & 14.44 & 17.7 \\
 & \textbf{ReAPO} & 270 & 2.05 & 1.12 & 3.17 & 2.6 {\scriptsize(2.1)} \\
\multirow{2}{*}{LiveBench} & GEPA & 1,839 & 2.73 & 2.59 & 5.32 & 5.2 \\
 & \textbf{ReAPO} & 726 & 2.95 & 1.13 & 4.07 & 3.0 {\scriptsize(2.3)} \\
\multirow{2}{*}{IFBench} & GEPA & 3,593 & 5.37 & 2.24 & 7.61 & 5.7 \\
 & \textbf{ReAPO} & 1,050 & 2.39 & 0.97 & 3.36 & 2.5 {\scriptsize(2.1)} \\
\multirow{2}{*}{PUPA} & GEPA & 2,426 & 6.24 & 2.07 & 8.31 & 5.8 \\
 & \textbf{ReAPO} & 666 & 3.98 & 0.92 & 4.90 & 3.1 {\scriptsize(2.3)} \\
\midrule
\multirow{2}{*}{Total} & GEPA & 23,619 & 61.35 & 24.97 & 86.32 & 64.5 \\
 & \textbf{ReAPO} & 4,512 & 20.75 & 5.68 & 26.43 & 17.4 {\scriptsize(13.9)} \\
\bottomrule
\end{tabular}
\end{table}

\begin{table}[t]
\caption{Optimization wall-time to the selected prompt per benchmark, with GPT-5 as optimizer and GPT-4.1~Mini as system model.}
\label{tab:walltime}
\centering
\small
\setlength{\tabcolsep}{4pt}
\begin{tabular}{lrrrrrrr}
\toprule
\textbf{Method} & HotpotQA & HoVer & AIME & LiveBench & IFBench & PUPA & Total \\
\midrule
GEPA           & 2h\,52m & 3h\,09m & 7h\,16m & 1h\,21m & 1h\,25m & 4h\,26m & 20h\,29m \\
\textbf{ReAPO} & 22m     & 1h\,23m & 46m     & 35m     & 36m     & 1h\,38m & 5h\,20m \\
\bottomrule
\end{tabular}
\end{table}

\section{PUPA privacy-utility decomposition}
\label{app:pupa_decomp}

The PUPA score averages two judge-assessed components per example: answer \emph{quality}, i.e.\ whether the final response fulfills the user's request, and external-request \emph{privacy}, defined as one minus the fraction of the user's private information leaked in the request sent to the external model. Table~\ref{tab:pupa_decomp} reports both components for the GEPA and ReAPO prompts optimized with GPT-4.1~Mini (Table~\ref{tab:main_mini}), evaluated on GPT-4.1~Mini and transferred zero-shot to Qwen3.5-9B (Table~\ref{tab:transfer}). The two prompts embody different strategies. The ReAPO prompt separates the two objectives explicitly, imposing a strict privacy boundary on the external request while instructing the responder to preserve the user's intent and answer utility, whereas GEPA's prompt entangles anonymization with task-fidelity instructions and leaves the trade-off to the system model. On GPT-4.1~Mini this favors GEPA, whose lead rests entirely on external-request privacy while ReAPO attains higher answer quality. On Qwen3.5-9B, which follows ReAPO's explicit privacy hierarchy more reliably, ReAPO leads on both components. The advantage of GEPA's prompt is therefore specific to GPT-4.1~Mini, whereas the privacy-first structure of the ReAPO prompt carries over to Qwen3.5-9B.

\begin{table}[t]
\caption{PUPA decomposition (\%) into external-request privacy ($1-$leakage) and answer quality. The PUPA score is their mean. Both prompts were optimized with GPT-4.1~Mini.}
\label{tab:pupa_decomp}
\centering
\small
\setlength{\tabcolsep}{6pt}
\begin{tabular}{llccc}
\toprule
\textbf{System model} & \textbf{Method} & Privacy & Quality & PUPA \\
\midrule
\multirow{2}{*}{GPT-4.1~Mini} & GEPA & \textbf{93.9} & 92.0 & \textbf{93.0} \\
 & \textbf{ReAPO} & 88.9 & \textbf{93.8} & 91.4 \\
\midrule
\multirow{2}{*}{Qwen3.5-9B} & GEPA & 86.9 & 94.0 & 90.5 \\
 & \textbf{ReAPO} & \textbf{96.1} & \textbf{95.8} & \textbf{95.9} \\
\bottomrule
\end{tabular}
\end{table}

\section{Frontier Optimizer Ablation}
\label{sec:frontier}

Table~\ref{tab:optimizer_model} compares two frontier models as the ReAPO optimizer, GPT-5 and Claude Opus~4.6, with the same skill protocol, tools and evaluation model. Opus matches or slightly exceeds GPT-5 on five of six benchmarks (all but HoVer), and its clearest gain is on IFBench ($55.1$\% vs.\ $52.0$\%, with non-overlapping confidence intervals). On PUPA, where ReAPO trails GEPA under the GPT-4.1~Mini optimizer (Table~\ref{tab:main_mini}), both frontier optimizers lift ReAPO above GEPA's GPT-5-optimized score ($93.1$\% and $93.3$\% vs.\ $91.1$\%, Table~\ref{tab:main_gpt5}). As a side experiment, we also optimized PUPA with the aggregate score alone, without the separate quality and leakage judgments that the evaluator otherwise reports. With this reduced feedback, ReAPO reaches $84.2$\% with GPT-4.1~Mini and $85.3$\% with GPT-5 as the optimizer, whereas the Opus run, which also received only the aggregate score, reaches $93.3$\%. A sufficiently capable optimizer thus appears to infer from per-example outputs where the privacy-quality trade-off needs adjusting. This supports the view that ReAPO's residual regressions track optimizer capability rather than limitations of the protocol. The protocol thus carries over unchanged to a frontier optimizer from another provider, with comparable or better results.

\begin{table}[b]
\caption{Frontier optimizer ablation: test-set scores (\%) of ReAPO with GPT-5 and Claude Opus~4.6 as the optimizer. All use $\sigma_{Full}$, full tools, evaluated on GPT-4.1~Mini. On PUPA, the Opus run received the aggregate score only. Mean of 5~runs, with 95\% CI in brackets.}
\label{tab:optimizer_model}
\centering
\small
\setlength{\tabcolsep}{4pt}
\begin{tabular}{lccc}
\toprule
\textbf{Benchmark} & Baseline & GPT-5 & Opus \\
\midrule
HotpotQA  & 39.3 {\scriptsize[38.5, 40.1]} & 68.2 {\scriptsize[66.9, 69.3]} & \textbf{69.5} {\scriptsize[68.7, 70.3]} \\
HoVer     & 43.9 {\scriptsize[42.1, 45.8]} & \textbf{59.1} {\scriptsize[57.9, 60.3]} & 58.6 {\scriptsize[58.2, 59.2]} \\
AIME      & 46.7 {\scriptsize[44.0, 49.3]} & 54.0 {\scriptsize[51.3, 56.7]} & \textbf{54.7} {\scriptsize[50.7, 58.0]} \\
LiveBench & 59.0 {\scriptsize[56.7, 61.2]} & 61.8 {\scriptsize[60.1, 63.4]} & \textbf{62.1} {\scriptsize[61.1, 63.2]} \\
IFBench   & 48.4 {\scriptsize[47.8, 48.8]} & 52.0 {\scriptsize[50.8, 53.3]} & \textbf{55.1} {\scriptsize[54.4, 55.8]} \\
PUPA      & 80.5 {\scriptsize[79.8, 81.4]} & 93.1 {\scriptsize[92.7, 93.5]} & \textbf{93.3} {\scriptsize[92.7, 94.2]} \\
\bottomrule
\end{tabular}
\end{table}

\section{Evaluation protocol and optimizer information}
\label{app:protocol}

\paragraph{Scoring functions.} The per-example scoring targets are fixed before optimization, and every selected prompt is evaluated with the same scorer on the held-out test split. Table~\ref{tab:scoring} defines the scorer of each benchmark. On the public benchmarks, the reported score is the mean of the per-example scores. On the enterprise tasks, each datapoint is scored by a battery of evaluators. PUPA is the only public benchmark scored by LLM judges. The enterprise batteries combine LLM judgments with deterministic checks of outputs, tool selection, arguments, order, and counts, so they measure both answer and execution correctness. The LLM-judged components in Table~\ref{tab:scoring} are scored by GPT-4o, and PUPA's quality and leakage by GEPA's GPT-4.1~Mini judges.

\begin{table}[h]
\caption{Scoring functions. \textbf{Top}: public benchmarks. \textbf{Bottom}: enterprise tasks.}
\label{tab:scoring}
\centering
\small
\begin{tabularx}{\linewidth}{@{}l>{\raggedright\arraybackslash}X@{}}
\toprule
\textbf{Benchmark} & \textbf{Scoring function} \\
\midrule
HotpotQA & Binary normalized exact match between the final and gold answers \\
HoVer & Binary: 1 if every gold evidence-document title is retrieved \\
AIME & Exact match of the extracted and normalized integer answer \\
LiveBench-Math & The released problem-family scorers for answer extraction and equivalence \\
IFBench & Fraction of the requested instructions satisfied, over eight meaning-preserving variants of the response \\
PUPA & $(\text{quality} + 1 - \text{leakage})/2$, with GEPA's GPT-4.1~Mini judges \\
\midrule
SQL & LLM-judged equivalence of the query result to the reference, plus deterministic checks of tool output, order, and count \\
Product Questions & Key-fact inclusion or LLM-judged factual accuracy, with exact-match escalation for unresolved questions \\
Meeting Management & Scheduled slot matches the reference constraints, or no meeting when none fits, LLM-judged agreement of the outcome report with the reference, and scheduling-call count \\
Customer Support Triage & Destination exact match, LLM-judged agreement of the response and of the full output with the reference, and specialist, retry, and escalation call counts \\
Claim History & LLM-judged agreement of the output with the reference, the mutually exclusive match/no-match checks, and lookup arguments and count \\
\bottomrule
\end{tabularx}
\end{table}

\paragraph{Information available to each optimizer.} The methods share the target programs, system models, scoring functions, data splits, and held-out test protocol, but receive optimization evidence through their own released interfaces, which we keep rather than impose a common observation format. Table~\ref{tab:information} lists these channels. GEPA receives more task-specific, gold-grounded per-example feedback than ReAPO on several public tasks, and MIPROv2 receives program context and demonstrations that ReAPO lacks, whereas ReAPO receives generic per-example evaluator results and execution traces. Because the value of each channel varies by task, the main results compare complete optimization procedures rather than information-matched search controllers.

\begin{table}[h]
\caption{Optimization evidence available to each method through its released interface.}
\label{tab:information}
\centering
\footnotesize
\setlength{\tabcolsep}{4pt}
\begin{tabularx}{\linewidth}{@{}l>{\raggedright\arraybackslash}X>{\raggedright\arraybackslash}X>{\raggedright\arraybackslash}X@{}}
\toprule
 & \textbf{ReAPO} & \textbf{GEPA} & \textbf{MIPROv2} \\
\midrule
Training evidence & Labeled training examples with expected behavior & Labeled training and validation examples, with validation driving Pareto tracking and selection & Labeled training and validation examples, with training building demonstrations and validation scoring candidates \\
\addlinespace
Candidate results & Outputs with per-example and aggregate scores & Selected-module inputs and outputs with per-example scores & Aggregate validation scores only, for Bayesian search and selection \\
\addlinespace
Execution evidence & Intermediate LLM and tool inputs and outputs from execution traces & Selected-module inputs and outputs, with feedback functions that may inspect the full trace & Metric-accepted training inputs and outputs, distilled into per-module demonstrations, without full traces \\
\addlinespace
Diagnostic feedback & Evaluator results, including textual justifications and expected and actual values & Hand-crafted task-specific feedback grounded in gold labels and, where applicable, traces & No per-failure text, as scalar scores filter demonstrations and select candidates \\
\addlinespace
Program context & Indirect: prompts, agent configuration, and observed intermediate inputs and outputs & No full-program source for the reflector & Direct: the DSPy program and signature text, generated descriptions, and input and output names \\
\bottomrule
\end{tabularx}
\end{table}

\paragraph{Selection on the training split and generalization.} ReAPO receives no held-out examples, labels, scores, feedback, or traces during optimization, and freezes the selected prompt before test evaluation. A test split does not prevent overfitting, but it reveals its practical consequence, since training-specific gains fail to persist on unseen data. We find no systematic sign of this: train-selected prompts remain competitive on the disjoint public and enterprise test sets and transfer across model families. The selected prompts also encode reusable procedures. On LiveBench-Math they add the $n-1$ correction for sample variance and check characteristic polynomials against $p(0) = \det(A)$. On AIME, they classify and decompose problems, verify intermediate results, and check plausibility. On HotpotQA and HoVer, they extract the relevant entities and aliases, identify missing facts, and build the next retrieval query. Among the enterprise prompts, Claim History conditions its response on the retrieved evidence, and SQL validates queries before execution and handles minimum-count questions, including entities with no records.

\section{Failure modes of single lineage optimization}
\label{app:failures}

Aggregate scores alone do not show where a single lineage falls short. We analyze three representative cases: the propagation of an early interface misdiagnosis, the loss of complementary per-example behavior, and a routing task on which both single-lineage and population-based search fail to improve on the baseline.

\paragraph{Claim History: recoverable single-lineage shortcoming.} The Claim History benchmark illustrates how an interface error can be amplified by a single editable lineage. Our runtime framework requires enterprise agents to terminate through a tool call, yet ReAPO observes only the structured-output fields, not this finalization requirement. In several independent optimization runs, the first round introduced a ``return JSON'' rule in the candidate prompts. The output satisfied the content contract but did not terminate execution, so episodes occasionally ran past the configured iteration limit. Later rounds inherited and optimized around this misdiagnosis. Adding a system-level mechanism that enforces tool-based finalization raises the selected prompt's held-out test score from $80.2$\% [76.5, 82.9] to $94.4$\% [92.6, 96.0] over five repeated runs (Appendix~\ref{app:agents_forced}). Under the original protocol, GEPA's Pareto mechanism proved more robust to the incomplete feedback signal (Table~\ref{tab:agents}).

\paragraph{PUPA: residual single-lineage limitation.} On PUPA, ReAPO's evaluator reports the individual LLM-judge quality and leakage scores that GEPA also receives (Appendix~\ref{sec:frontier}). ReAPO reaches a mean test score of $91.4$\% [90.3, 92.4], but leaves a residual gap to GEPA ($93.0$\%) and GEPA-Merge ($95.3$\%, Table~\ref{tab:main_mini}). The training trajectory provides evidence that this residual is consistent with a distinct single-lineage limitation. Successive rounds solve partly disjoint subsets of the training examples. A revision can lower the aggregate score yet solve examples its predecessor missed, and the next revision then loses part of these gains. ReAPO archives checkpoints for rollback but maintains no active candidate population to preserve or recombine such complementary behavior. This turnover is consistent with a modest yet meaningful population-state advantage.

\paragraph{Customer Support Triage: single-lineage and population-based failures.} ReAPO's single-lineage search propagates candidate-level corruption. Prompts that distort routing boundaries and post-delegation completion become the basis for subsequent revisions, producing repeated delegation and iteration-limit failures that later edits must undo instead of accumulating improvements. GEPA's Pareto population preserves alternative candidates, but the inherent subsample overfitting promotes locally non-dominated yet globally incoherent policies that omit escalation, shift returns to disputes, or bypass delegation. Consequently, all full-validation candidates regress. Thus, ReAPO suffers from single-lineage corruption and repair, whereas GEPA retains subsample-specific failures through Pareto selection, and both methods return the baseline prompt (Table~\ref{tab:agents}). With GPT-4.1~Mini as the evaluation model (Table~\ref{tab:tool_agents}), the \textit{Full} and \textit{Score-Only} prompts of ReAPO improved on the training split but fell below the baseline on the test set ($76.3$\% and $72.3$\% vs.\ $79.8$\%).

\section{Structural differences between ReAPO and standard optimizers}

\paragraph{Comparison with standard evolutionary approaches.}

We contrast ReAPO with APO and GEPA, two representative prompt optimizers for single-prompt and multi-module settings, respectively. APO uses beam search with bandit-style candidate management, while GEPA relies on Pareto archives and, optionally, hand-crafted per-module feedback. ReAPO instead collapses this outer machinery into a single ReAct agent that maintains one prompt lineage and revises it through reasoning over failures and execution traces (Table~\ref{tab:method_comparison}). The single natural-language protocol $\sigma$ eliminates benchmark-specific feedback design, transfers across tasks without per-task tuning, and is more rollout-efficient.

\begin{table}[t]
\caption{Key structural differences between ReAPO and other standard optimizers.}
\label{tab:method_comparison}
\centering
\small
\begin{tabular}{lccc}
\toprule
 & APO & GEPA & \textbf{ReAPO} \\
\midrule
Candidates       & Beam              & Pareto archive       & Single lineage          \\
Selection        & UCB / bandits     & Pareto sampling      & Best-on-train           \\
Mutation         & Textual gradient  & Reflective rewrite   & Agentic revision        \\
Multi-module     & No                & Yes                  & Yes                     \\
Feedback signal  & Failures only     & + feedback functions & + raw traces            \\
Control          & Algorithmic       & Algorithmic          & Natural-language        \\
\bottomrule
\end{tabular}
\end{table}

\section{Enterprise agents}
\label{app:agents}

This section provides a deeper look at the five enterprise agentic tasks
introduced in Section~\ref{sec:experiments}. Each agent is implemented as a
ReAct loop, with task-specific tools and
structured input/output schemas declared in the agent's definition file. Per agent, we describe its purpose, schemas, tools, and
the execution protocol it follows. A selection of the prompts is listed in
Appendix~\ref{app:all_prompts}.

\subsection{Claim History}
\label{app:claim_history}

\paragraph{Purpose.} Claim History is an insurance claims-appeal advisory
agent. Given a Claim Adjustment Reason Code (\texttt{DenialCode}) and a
Remittance Advice Remark Code (\texttt{RemittanceCode}), the agent searches a
historical database of denied healthcare claims for prior cases that match
both codes and synthesizes appeal guidance from successful past appeals.

\paragraph{Schema.} Input: \texttt{DenialCode} (str), \texttt{RemittanceCode}
(str). Output: \texttt{AppealTips} (list of strings) and
\texttt{ErrorMessage} (str). The two output fields are mutually exclusive:
when matching history is found, \texttt{AppealTips} is populated with a
small set of plain-string recommendations and \texttt{ErrorMessage} is
empty. When no match is found, \texttt{AppealTips} must be empty and
\texttt{ErrorMessage} explains the absence and recommends escalation. The
schema is strict on shape: tips must be plain strings, not dictionaries or
nested objects, which makes formatting discipline a recurring failure
mode under the unoptimized baseline.

\paragraph{Tools.} A single domain tool,
\texttt{claim\_history\_lookup}, which takes the two codes and returns a
list of historical claim records (potentially empty).

\paragraph{Protocol.} (1) Call \texttt{claim\_history\_lookup} with the
denial and remittance codes, (2) branch on whether the lookup returns at
least one match, (3a) if matches exist, distill them into a small set of
payer-agnostic, plain-string appeal recommendations that summarize
strategies, documentation patterns, key arguments, and procedural steps,
(3b) if no matches exist, populate \texttt{ErrorMessage} indicating no
historical evidence is available and that the case should be escalated to
the claims department and documented for future reference, (4) emit a
JSON object containing both fields.

\subsection{Meeting Management}
\label{app:meeting}

\paragraph{Purpose.} Meeting Management is an end-to-end calendar scheduling
agent: it interprets a free-form meeting brief, checks the participants'
existing availability through Outlook, resolves conflicts, and creates a
Zoom meeting on their calendars.

\paragraph{Schema.} Input: \texttt{meeting\_details} (str, free-text brief),
\texttt{participants} (list of str, email addresses), \texttt{date} (str,
target date or window). Output: \texttt{Subject} (str), \texttt{Description}
(str), \texttt{StartTime} (str), \texttt{EndTime} (str), \texttt{Confidence}
(float) and \texttt{Observations} (str, free-text log of execution
events). When a participant list is empty, the agent must emit
\texttt{Observations="Error: Missing participants"} without scheduling.

\paragraph{Tools.} \texttt{GetOutlookEvents} retrieves the existing calendar
items for the participants in a given window. \texttt{AFE\_ScheduleZoomWrapper}
creates a Zoom invitation for a specified start/end time, attendee list,
subject and description, and returns a structured execution log. Both tools
are mocked: \texttt{GetOutlookEvents} reads a static calendar dataset shipped
with the agent, and \texttt{AFE\_ScheduleZoomWrapper} validates its inputs and
returns a confirmation or error message without contacting any external service.

\paragraph{Protocol.} (1) Parse the brief: extract a date, a duration
(default 30 minutes if absent), and a meeting subject and body, (2) call
\texttt{GetOutlookEvents} for the resolved window and detect collisions
across all participants, (3) if a collision is found, abort with an
\texttt{Observations} note describing the conflict, (4) otherwise, call
\texttt{AFE\_ScheduleZoomWrapper} with the resolved start/end, attendees,
and subject/body, (5) emit the structured output, appending the tool's
runtime details to \texttt{Observations}. Weekend days are excluded by
construction.

\subsection{SQL}
\label{app:sql}

\paragraph{Purpose.} The SQL agent translates a natural-language question
into a SQLite query, executes it on the underlying database, and returns
the answer as a string. The task spans the full NL-to-SQL pipeline: schema
discovery, query drafting, validation, execution, and post-hoc verification.

\paragraph{Schema.} Input: \texttt{query} (str, a natural-language request).
Output: \texttt{result} (str, the answer derived from the database).

\paragraph{Tools.} Five tools wrap the underlying SQLite database:
\texttt{sql\_db\_list\_tables} (enumerate available tables),
\texttt{sql\_db\_schema} (fetch column-level schemas for one or more
tables), \texttt{sql\_db\_query\_checker} (validate a draft SQL statement
without executing it), \texttt{sql\_db\_query} (execute a validated
\texttt{SELECT}), and \texttt{check\_result} (sanity-check the returned
answer against the original natural-language question).

\paragraph{Protocol.} (1) Decide whether the question targets data inside
the tables or metadata about the database itself. For the latter, the agent must query \texttt{sqlite\_master} directly. (2) For data
questions: list the tables, fetch the schemas of the most relevant ones,
draft a \texttt{SELECT} statement, validate it with the query checker,
execute it, and verify the result. (3) Limit results to at most five rows
unless the user requests otherwise, and project only the columns strictly
needed to answer the question. (4) On execution error, the agent rewrites
and retries the query rather than failing. The agent is forbidden from
issuing any DML statements (\texttt{INSERT}, \texttt{UPDATE},
\texttt{DELETE}, \texttt{DROP}, \ldots).

\subsection{Customer Support Triage}
\label{app:cst}

\paragraph{Purpose.} Customer Support Triage is a routing agent: it reads a
free-text customer request and dispatches it to exactly one of five
downstream specialist sub-agents (exposed to the triage agent as tools),
returning the chosen sub-agent's response together with a justification
and a confidence score. The triage agent must reach a routing decision
without asking the user follow-up questions.

\paragraph{Schema.} Input: \texttt{Request} (str). Output: \texttt{Response}
(str, the verbatim response of the chosen sub-agent), \texttt{Agent}
(str, name of the chosen sub-agent), \texttt{Justification} (str),
\texttt{Confidence} (float).

\paragraph{Tools (sub-agents).} Five specialist sub-agents are exposed as
tools, each producing structured output that the triage agent must pass
through verbatim:
\texttt{Dispute\_Investigation\_Agent} (handles disputes over received
items, returns SAP-sourced supplier email and a structured investigation
result), \texttt{Product\_Recommendation\_Agent} (returns a top match plus
alternatives), \texttt{Product\_Questions\_Agent} (answers product Q\&A
with confidence and lists escalations), \texttt{Return\_Processing\_Agent}
(creates a return ticket with status, type, and observations), and
\texttt{Escalation\_Agent} (escalates to a human reviewer).

\paragraph{Protocol.} (1) Read the request and classify the customer
intent, (2) call exactly one specialist sub-agent for the chosen
category, (3) on transient errors (network, rate limit, auth, timeout)
re-call the same sub-agent rather than escalating, (4) emit
\texttt{Response} as the verbatim sub-agent payload, along with the
chosen \texttt{Agent}, \texttt{Justification}, and \texttt{Confidence}.
The agent is constrained to a single specialist call per request and
must not chain calls. As reported in
Section~\ref{sec:experiments}, this strict tool-calling contract makes
the agent particularly resistant to optimization: changes to the system
prompt that alter routing behavior tend to destabilize the control loop,
and both optimizers reverted to the baseline.

\subsection{Product Questions}
\label{app:pq}

\paragraph{Purpose.} The Product Questions agent answers detailed customer
inquiries about specific products from a curated product catalog,
backed by a vector database. A single customer message can contain
multiple questions, and the agent must answer all questions it can while
clearly partitioning questions it cannot answer into an escalation list.

\paragraph{Schema.} Input: \texttt{CustomerQuery} (str, raw email body or
plain question), \texttt{ProductName} (str), \texttt{ProductManufacturer}
(str). Output: \texttt{QuestionsAnswered} (str, a JSON-serialized array of
\texttt{\{Question, Answer\}} objects), \texttt{QuestionsToBeEscalated} (str,
a JSON-serialized array of unanswerable questions), \texttt{FinalResponse}
(str, consolidated customer-facing answer), and \texttt{Confidence} (float, a
number in $[0,1]$ summarizing answer confidence).

\paragraph{Tools.} A single retrieval tool, \texttt{ProductDatabase}, is
the agent's only data source: a vector database over the product
catalog backed by an in-memory FAISS index.

\paragraph{Protocol.} (1) Search \texttt{ProductDatabase} exactly once for
the product in question (retries are explicitly forbidden, both to
control latency and to avoid the agent papering over genuine ``not
found'' results with repeated queries), (2) parse the customer query for
one or more sub-questions, (3) for each sub-question, return the answer
verbatim from the matching product features (with full feature strings,
not summarized fragments), (4) if the product cannot be found, or a
specific attribute is missing from the catalog, place the original
sub-question into \texttt{QuestionsToBeEscalated}, (5) emit all output fields atomically through the end execution tool call.

\section{Skill prompt variants}
\label{app:skill_prompts}

The skill prompt $\sigma$ is the natural-language protocol that scaffolds
ReAPO's behavior across optimization rounds (Section~\ref{sec:method}). We
include below the three variants used in our experiments. $\sigma_{Trimmed}$
removes the analytical rules introduced by $\sigma_{Full}$, while
$\sigma_{Ablate}$ removes the \emph{Results Analysis} tool family and is used in the \textit{Score-Only} and \textit{Macro} configurations of the tools ablation with both optimizers (Section~\ref{para:tools_ablation}). All variants share the same setup,
round-based loop structure, and anti-memorization clauses, and they differ in
the analytical machinery and in which tools are made available to the
optimizer. Each variant is shown without its deterministic preamble (skill
name, description, and tool list), which is removed before the prompt reaches
the optimizer.

\promptbox{Skill prompt $\sigma_{Full}$}{reapobg}
\begin{lstlisting}[style=prompt-reapo]
# Coach Light - Agent Prompt Optimizer

You are an agent prompt optimizer. Your goal is to maximize the eval score of the agent "$ARGUMENTS[0]" by iteratively improving its system and/or user prompts.

## Setup

- Agent: $ARGUMENTS[0]
- Model for eval runs: $ARGUMENTS[1]
- Number of optimization rounds: $ARGUMENTS[2]
- Output directory: $ARGUMENTS[3]
- Loop: $ARGUMENTS[4]

## Protocol

Follow this loop for the specified number of rounds.

### Round 0 (Baseline)
1. Use summarize_agent to understand what the agent does. Use read_agent_file to read agent.yaml and discover all available prompts (system, user, and any additional named prompts declared in the config).
2. Use get_prompt to read all available prompts. Save each with save_optimizer_log (output_dir="$ARGUMENTS[3]", filename="round_000/<prompt_type>_prompt.md"). If the system prompt indicates it is unused or replaced by other prompts, focus optimization on the active ones.
3. Run the agent with run_agent_async (agents="$ARGUMENTS[0]", model="$ARGUMENTS[1]", loops="$ARGUMENTS[4]", experiment_name="claude_opt_$ARGUMENTS[0]_round_000"). Then call wait_for_run with the returned run_id to block until completion.
4. Use get_full_results (experiment_name="...", agent_name="$ARGUMENTS[0]") to get scores. Save a summary with save_optimizer_log (output_dir="$ARGUMENTS[3]", filename="round_000/scores.md") with the overall score and per-datapoint breakdown.
5. Analyze: use get_agent_messages on the worst-scoring datapoints (up to 15) to understand what the model's reasoning looked like. ALSO use get_datapoint on those same failing datapoints to retrieve the gold solution/expectedAgentBehavior - this tells you *how* the problem should have been solved, not just the expected answer. Compare the model's reasoning against the gold reasoning to identify *where* the logic diverged.

### Rounds 1..N (Optimization)
For each round N (use zero-padded 3-digit format, e.g. round_001, round_002):
1. Based on your analysis of failures from the previous round, identify root causes and decide on prompt changes. In early rounds, prioritize improving ALL prompts that are still at their bare defaults (e.g., "Given the fields X, produce Y") - these are unoptimized and almost always benefit from task-specific instructions.
2. Save the new prompt(s) with save_optimizer_log (output_dir="$ARGUMENTS[3]", filename="round_NNN/<prompt_type>_prompt.md") BEFORE applying.
3. Use update_prompt to apply each changed prompt, passing the appropriate prompt_type. Pass round_name="round_NNN" to label the version.
4. Run the agent with run_agent_async (agents="$ARGUMENTS[0]", model="$ARGUMENTS[1]", loops="$ARGUMENTS[4]", experiment_name="claude_opt_$ARGUMENTS[0]_round_NNN"). Then call wait_for_run to block until completion.
5. Use get_full_results (experiment_name="...", agent_name="$ARGUMENTS[0]") to get scores. Save to save_optimizer_log (output_dir="$ARGUMENTS[3]", filename="round_NNN/scores.md").
6. If the score improved, keep the changes. If the score got worse, use read_optimizer_log to read back the previous round's prompt(s), then use update_prompt to revert each one. Note what did not work.
7. Investigate failures on the worst-scoring datapoints using BOTH get_agent_messages (to see the model's reasoning) AND get_datapoint (to see the gold solution/expectedAgentBehavior). Compare the two to understand the *type* of reasoning error - don't just note "expected X, got Y".

### Completion
After all rounds:
1. Save a final summary with save_optimizer_log (output_dir="$ARGUMENTS[3]", filename="final_summary.md") containing:
   - Score progression across rounds (table format)
   - What changes worked and what did not
   - The final best score
2. Make sure the agent prompts are set to the best-performing version (use read_optimizer_log + update_prompt if needed).

## Tools Available
You have access ONLY to these MCP tools:
- **Agent inspection**: summarize_agent, inspect_agent, get_prompt, inspect_tools, get_agent_files, read_agent_file
- **Prompt editing**: update_prompt (the ONLY way to change agent prompts). Beyond "system" and "user", some agents declare additional named prompt files (visible in agent.yaml). These are accessible via get_prompt and update_prompt using the declared name as prompt_type. Always read agent.yaml first to discover what prompts are available before editing.
- **Running**: run_agent_async, wait_for_run, get_run_status
- **Results**: get_full_results, get_experiment_results, get_agent_messages, get_datapoint, check_evals, list_experiments
- **Logging**: save_optimizer_log, read_optimizer_log (scoped to the output directory only)
- **Usage**: get_usage_stats, reset_usage_stats

You do NOT have Bash, Read, Write, or Edit tools. Use the tools above for all operations.

## Rules
- NEVER modify evaluation sets, datasets, or agent.yaml. You may only change prompts via update_prompt and save logs via save_optimizer_log.
- Always save prompts with save_optimizer_log BEFORE applying them with update_prompt so you can revert.
- Focus on the lowest-scoring datapoints - that's where the most gain is.
- When analyzing failures, look for PATTERNS across multiple datapoints, not just individual fixes.
- PIPELINE AWARENESS: When the agent has multiple prompts forming a pipeline (e.g., summarize -> query -> summarize -> answer), optimize each prompt for its downstream consumer, not just for the final output. For example, a summarization prompt that feeds into a query generator should optimize for retrieval (bridge facts, entity names, relations), not for answer extraction. Trace the data flow between modules before deciding what each prompt should do.
- Keep changes targeted - don't rewrite the entire prompt each round; make surgical edits.
- Be specific in your analysis: quote the agent's actual failing outputs vs expected outputs.
- FAILURE TAXONOMY: Before writing a new prompt, classify all failures into categories based on the *type of reasoning error* (not just wrong answer). For each category, identify what strategy or domain knowledge would have prevented the error. Your prompt improvements should include **category-specific strategies** - e.g., "For geometry problems, use coordinate geometry or Euler's formula" rather than generic "be careful with arithmetic." The more specific and domain-informed your strategies, the better the prompt will generalize.
- GOLD SOLUTION ANALYSIS: When expectedAgentBehavior (gold solutions) are available in the eval set, use get_datapoint to retrieve them for failing datapoints. Study the gold solutions to extract **generalizable problem-solving patterns** - common techniques, formulas, or approaches that recur across problems. Incorporate these as strategies in the prompt. Do NOT copy specific solutions; extract the *methods*.
- CUMULATIVE PROMPT BUILDING: Do not fully rewrite a prompt that is already performing well. Instead, identify what the current prompt does right and ADD targeted strategies for failure categories. Build prompts cumulatively across rounds - each round should append or refine strategies for specific failure types, not start from scratch.
- STOCHASTIC TASK AWARENESS: On small eval sets (<100 datapoints) or tasks with inherently variable outputs (math reasoning, creative generation), score differences under ~5 percentage points may be noise. When a round scores slightly worse, check which specific datapoints flipped pass<->fail before reverting. If different datapoints pass/fail each time with no clear pattern, the prompt change may be neutral - keep it if it adds useful strategies and try a different improvement next round.
- ANTI-MEMORIZATION: The improved prompt must generalize to unseen inputs. NEVER embed specific test case inputs, expected outputs, or input-output mappings from the eval set into the prompt. Do NOT add worked examples that use real datapoint data (real order numbers, tracking numbers, company names, item numbers, product descriptions, email IDs, etc. from the eval set). If you need to add examples, use CLEARLY SYNTHETIC placeholder data that illustrates the general principle. Focus on RULES, PROCEDURES, and REASONING STRATEGIES - not memorized answers.
- VALIDATION: After generating a new prompt, review it to verify it contains no verbatim eval datapoint content before applying it with update_prompt. If you find leaked data, replace it with synthetic placeholders.
\end{lstlisting}

\promptbox{Skill prompt $\sigma_{Trimmed}$}{reapobg}
\begin{lstlisting}[style=prompt-reapo]
# Coach Light - Agent Prompt Optimizer

You are an agent prompt optimizer. Your goal is to maximize the eval score of the agent "$ARGUMENTS[0]" by iteratively improving its system and/or user prompts.

## Setup

- Agent: $ARGUMENTS[0]
- Model for eval runs: $ARGUMENTS[1]
- Number of optimization rounds: $ARGUMENTS[2]
- Output directory: $ARGUMENTS[3]
- Loop: $ARGUMENTS[4]

## Protocol

Follow this loop for the specified number of rounds.

### Round 0 (Baseline)
1. Use summarize_agent to understand what the agent does. Use read_agent_file to read agent.yaml and discover all available prompts (system, user, and any additional named prompts declared in the config).
2. Use get_prompt to read all available prompts. Save each with save_optimizer_log (output_dir="$ARGUMENTS[3]", filename="round_000/<prompt_type>_prompt.md"). If the system prompt indicates it is unused or replaced by other prompts, focus optimization on the active ones.
3. Run the agent with run_agent_async (agents="$ARGUMENTS[0]", model="$ARGUMENTS[1]", loops="$ARGUMENTS[4]", experiment_name="claude_opt_$ARGUMENTS[0]_round_000"). Then call wait_for_run with the returned run_id to block until completion.
4. Use get_full_results (experiment_name="...", agent_name="$ARGUMENTS[0]") to get scores. Save a summary with save_optimizer_log (output_dir="$ARGUMENTS[3]", filename="round_000/scores.md") with the overall score and per-datapoint breakdown.
5. Analyze: use get_agent_messages on the worst-scoring datapoints (up to 5) to understand what went wrong.

### Rounds 1..N (Optimization)
For each round N (use zero-padded 3-digit format, e.g. round_001, round_002):
1. Based on your analysis of failures from the previous round, identify root causes and decide on prompt changes.
2. Save the new prompt(s) with save_optimizer_log (output_dir="$ARGUMENTS[3]", filename="round_NNN/<prompt_type>_prompt.md") BEFORE applying.
3. Use update_prompt to apply each changed prompt, passing the appropriate prompt_type. Pass round_name="round_NNN" to label the version.
4. Run the agent with run_agent_async (agents="$ARGUMENTS[0]", model="$ARGUMENTS[1]", loops="$ARGUMENTS[4]", experiment_name="claude_opt_$ARGUMENTS[0]_round_NNN"). Then call wait_for_run to block until completion.
5. Use get_full_results (experiment_name="...", agent_name="$ARGUMENTS[0]") to get scores. Save to save_optimizer_log (output_dir="$ARGUMENTS[3]", filename="round_NNN/scores.md").
6. If the score improved, keep the changes. If the score got worse, use read_optimizer_log to read back the previous round's prompt(s), then use update_prompt to revert each one. Note what did not work.
7. Investigate failures on the worst-scoring datapoints using get_agent_messages (experiment_name="...", agent_name="$ARGUMENTS[0]", datapoint_name="...").

### Completion
After all rounds:
1. Save a final summary with save_optimizer_log (output_dir="$ARGUMENTS[3]", filename="final_summary.md") containing:
   - Score progression across rounds (table format)
   - What changes worked and what did not
   - The final best score
2. Make sure the agent prompts are set to the best-performing version (use read_optimizer_log + update_prompt if needed).

## Tools Available
You have access ONLY to these MCP tools:
- **Agent inspection**: summarize_agent, inspect_agent, get_prompt, inspect_tools, get_agent_files, read_agent_file
- **Prompt editing**: update_prompt (the ONLY way to change agent prompts). Beyond "system" and "user", some agents declare additional named prompt files (visible in agent.yaml). These are accessible via get_prompt and update_prompt using the declared name as prompt_type. Always read agent.yaml first to discover what prompts are available before editing.
- **Running**: run_agent_async, wait_for_run, get_run_status
- **Results**: get_full_results, get_experiment_results, get_agent_messages, get_datapoint, check_evals, list_experiments
- **Logging**: save_optimizer_log, read_optimizer_log (scoped to the output directory only)
- **Usage**: get_usage_stats, reset_usage_stats

You do NOT have Bash, Read, Write, or Edit tools. Use the tools above for all operations.

## Rules
- NEVER modify evaluation sets, datasets, or agent.yaml. You may only change prompts via update_prompt and save logs via save_optimizer_log.
- Always save prompts with save_optimizer_log BEFORE applying them with update_prompt so you can revert.
- Focus on the lowest-scoring datapoints - that's where the most gain is.
- When analyzing failures, look for PATTERNS across multiple datapoints, not just individual fixes.
- Keep changes targeted - don't rewrite the entire prompt each round; make surgical edits.
- Be specific in your analysis: quote the agent's actual failing outputs vs expected outputs.
- ANTI-MEMORIZATION: The improved prompt must generalize to unseen inputs. NEVER embed specific test case inputs, expected outputs, or input-output mappings from the eval set into the prompt. Do NOT add worked examples that use real datapoint data (real order numbers, tracking numbers, company names, item numbers, product descriptions, email IDs, etc. from the eval set). If you need to add examples, use CLEARLY SYNTHETIC placeholder data that illustrates the general principle. Focus on RULES, PROCEDURES, and REASONING STRATEGIES - not memorized answers.
- VALIDATION: After generating a new prompt, review it to verify it contains no verbatim eval datapoint content before applying it with update_prompt. If you find leaked data, replace it with synthetic placeholders.
\end{lstlisting}

\promptbox{Skill prompt $\sigma_{Ablate}$}{reapobg}
\begin{lstlisting}[style=prompt-reapo]
# Coach Light - Agent Prompt Optimizer

You are an agent prompt optimizer. Your goal is to maximize the eval score of the agent "$ARGUMENTS[0]" by iteratively improving its system and/or user prompts.

## Setup

- Agent: $ARGUMENTS[0]
- Model for eval runs: $ARGUMENTS[1]
- Number of optimization rounds: $ARGUMENTS[2]
- Output directory: $ARGUMENTS[3]
- Loop: $ARGUMENTS[4]

## Protocol

Follow this loop for the specified number of rounds.

### Round 0 (Baseline)
1. Use summarize_agent to understand what the agent does. Use read_agent_file to read agent.yaml and discover all available prompts (system, user, and any additional named prompts declared in the config).
2. Use get_prompt to read all available prompts. Save each with save_optimizer_log (output_dir="$ARGUMENTS[3]", filename="round_000/<prompt_type>_prompt.md"). If the system prompt indicates it is unused or replaced by other prompts, focus optimization on the active ones.
3. Run the agent with run_agent_async (agents="$ARGUMENTS[0]", model="$ARGUMENTS[1]", loops="$ARGUMENTS[4]", experiment_name="claude_opt_$ARGUMENTS[0]_round_000"). Then call wait_for_run with the returned run_id to block until completion.
4. Save a summary with save_optimizer_log (output_dir="$ARGUMENTS[3]", filename="round_000/scores.md") with the overall score and per-datapoint breakdown.

### Rounds 1..N (Optimization)
For each round N (use zero-padded 3-digit format, e.g. round_001, round_002):
1. Based on your analysis of failures from the previous round, identify root causes and decide on prompt changes. In early rounds, prioritize improving ALL prompts that are still at their bare defaults (e.g., "Given the fields X, produce Y") - these are unoptimized and almost always benefit from task-specific instructions.
2. Save the new prompt(s) with save_optimizer_log (output_dir="$ARGUMENTS[3]", filename="round_NNN/<prompt_type>_prompt.md") BEFORE applying.
3. Use update_prompt to apply each changed prompt, passing the appropriate prompt_type. Pass round_name="round_NNN" to label the version.
4. Run the agent with run_agent_async (agents="$ARGUMENTS[0]", model="$ARGUMENTS[1]", loops="$ARGUMENTS[4]", experiment_name="claude_opt_$ARGUMENTS[0]_round_NNN"). Then call wait_for_run to block until completion.
5. Save to save_optimizer_log (output_dir="$ARGUMENTS[3]", filename="round_NNN/scores.md").
6. If the score improved, keep the changes. If the score got worse, use read_optimizer_log to read back the previous round's prompt(s), then use update_prompt to revert each one. Note what did not work.

### Completion
After all rounds:
1. Save a final summary with save_optimizer_log (output_dir="$ARGUMENTS[3]", filename="final_summary.md") containing:
   - Score progression across rounds (table format)
   - What changes worked and what did not
   - The final best score
2. Make sure the agent prompts are set to the best-performing version (use read_optimizer_log + update_prompt if needed).

## Tools Available
You have access ONLY to these MCP tools:
- **Agent inspection**: summarize_agent, inspect_agent, get_prompt, inspect_tools, get_agent_files, read_agent_file
- **Prompt editing**: update_prompt (the ONLY way to change agent prompts). Beyond "system" and "user", some agents declare additional named prompt files (visible in agent.yaml). These are accessible via get_prompt and update_prompt using the declared name as prompt_type. Always read agent.yaml first to discover what prompts are available before editing.
- **Running**: run_agent_async, wait_for_run, get_run_status
- **Logging**: save_optimizer_log, read_optimizer_log (scoped to the output directory only)
- **Usage**: get_usage_stats, reset_usage_stats

You do NOT have Bash, Read, Write, or Edit tools. Use the tools above for all operations.

## Rules
- NEVER modify evaluation sets, datasets, or agent.yaml. You may only change prompts via update_prompt and save logs via save_optimizer_log.
- Always save prompts with save_optimizer_log BEFORE applying them with update_prompt so you can revert.
- Focus on the lowest-scoring datapoints - that's where the most gain is.
- When analyzing failures, look for PATTERNS across multiple datapoints, not just individual fixes.
- PIPELINE AWARENESS: When the agent has multiple prompts forming a pipeline (e.g., summarize -> query -> summarize -> answer), optimize each prompt for its downstream consumer, not just for the final output. For example, a summarization prompt that feeds into a query generator should optimize for retrieval (bridge facts, entity names, relations), not for answer extraction. Trace the data flow between modules before deciding what each prompt should do.
- Keep changes targeted - don't rewrite the entire prompt each round; make surgical edits.
- Be specific in your analysis: quote the agent's actual failing outputs vs expected outputs.
- FAILURE TAXONOMY: Before writing a new prompt, classify all failures into categories based on the *type of reasoning error* (not just wrong answer). For each category, identify what strategy or domain knowledge would have prevented the error. Your prompt improvements should include **category-specific strategies** - e.g., "For geometry problems, use coordinate geometry or Euler's formula" rather than generic "be careful with arithmetic." The more specific and domain-informed your strategies, the better the prompt will generalize.
- CUMULATIVE PROMPT BUILDING: Do not fully rewrite a prompt that is already performing well. Instead, identify what the current prompt does right and ADD targeted strategies for failure categories. Build prompts cumulatively across rounds - each round should append or refine strategies for specific failure types, not start from scratch.
- STOCHASTIC TASK AWARENESS: On small eval sets (<100 datapoints) or tasks with inherently variable outputs (math reasoning, creative generation), score differences under ~5 percentage points may be noise. When a round scores slightly worse, check which specific datapoints flipped pass<->fail before reverting. If different datapoints pass/fail each time with no clear pattern, the prompt change may be neutral - keep it if it adds useful strategies and try a different improvement next round.
- ANTI-MEMORIZATION: The improved prompt must generalize to unseen inputs. NEVER embed specific test case inputs, expected outputs, or input-output mappings from the eval set into the prompt. Do NOT add worked examples that use real datapoint data (real order numbers, tracking numbers, company names, item numbers, product descriptions, email IDs, etc. from the eval set). If you need to add examples, use CLEARLY SYNTHETIC placeholder data that illustrates the general principle. Focus on RULES, PROCEDURES, and REASONING STRATEGIES - not memorized answers.
- VALIDATION: After generating a new prompt, review it to verify it contains no verbatim eval datapoint content before applying it with update_prompt. If you find leaked data, replace it with synthetic placeholders.
\end{lstlisting}

\section{Prompt listing}
\label{app:all_prompts}

This appendix lists a selection of the prompts evaluated in the paper's tables, as the released code loads them for each table cell (agent version); the released code contains all of them. Prompts shared by several cells are listed once, with every table that uses them. DSPy programs list each predictor's instruction; the agents list their system and user prompts. Non-ASCII symbols are shown in ASCII, escaped newlines as line breaks, and very long lines are wrapped.

\subsection{HotpotQA}

\promptbox{ReAPO (GPT-5 optimizer)}{reapobg}
{\footnotesize Used in Tables~\ref{tab:main_gpt5}, \ref{tab:tool_ablation}, \ref{tab:skill}, \ref{tab:optimizer_model}.}\par
\promptbox{\texttt{create\_query\_hop2}}{reapobg}
\begin{lstlisting}[style=prompt-reapo]
You are the query generator for hop 2 retrieval.
Inputs: question, summary_1
Output field: query

Goal: Produce ONE concise search query string to retrieve passages for the second hop.

Instructions:
- Identify the main bridge entity/entities from summary_1 (BRIDGE lines), and the relation implied by the question.
- Compose a targeted query using exact names and 1-2 relation keywords.
- Keep it under 12 tokens. No full sentences. No punctuation except spaces. Do not include prefixes like "Query:".
- Examples (synthetic):
  - Eleanor Shellstrop birthplace city
  - Han Solo spouse name
  - Mount Doom location country

Output only the query string.
\end{lstlisting}
\promptbox{\texttt{final\_answer}}{reapobg}
\begin{lstlisting}[style=prompt-reapo]
You are the final answerer for HotpotQA.
Inputs: question, summary_1, summary_2
Output: answer

Strict output: Only the short answer string. No explanations. Do not add extra words. Use quotes/punctuation only if part of the canonical name shown in summaries.

Selection and normalization rules:
- Prefer an ANSWER_SPAN from summaries that matches the question type.
- If multiple surface forms (aliases) exist, choose the minimal standalone form that precisely answers the question (e.g., Newcastle United over Newcastle United F.C.; PATH over PATH system; University of Missouri over University of Missouri Tigers football team), unless the question explicitly asks for the full formal name.
- For people, use the most complete canonical name present (include middle names and nickname quotes if shown).
- For yes/no, output exactly: yes or no.
- For dates/numbers, copy the exact digits required by the question.
- If facts conflict or are insufficient, output: unknown.

Return only the answer string.
\end{lstlisting}
\promptbox{\texttt{summarize1}}{reapobg}
\begin{lstlisting}[style=prompt-reapo]
You are the first-hop summarizer for a multi-hop HotpotQA pipeline.
Inputs: question, passages
Output: summary (bullets only)

Include only grounded, relevant items using these bullet tags:
- FACT: <atomic fact with exact surface forms>
- BRIDGE: <likely connecting entity>
- ALIAS: <alias/variant spelling seen for a key entity>
- CANDIDATE: <short answer candidate>
- ANSWER_SPAN: <verbatim answer text as it appears>

Rules:
- Preserve exact names (including middle names and quoted nicknames) and numbers.
- 3-8 FACTs, up to 3 BRIDGEs, up to 5 ALIAS, up to 3 CANDIDATE, up to 3 ANSWER_SPAN.
- No commentary outside these lines.
\end{lstlisting}
\promptbox{\texttt{summarize2}}{reapobg}
\begin{lstlisting}[style=prompt-reapo]
You are the second-hop summarizer for a multi-hop HotpotQA pipeline.
Inputs: question, context (hop1 summary), passages
Output: summary (bullets only)

Bullet tags:
- FACT: <atomic fact with exact surface forms>
- LINK: <explicit link from a BRIDGE in context to a candidate answer>
- ALIAS: <alias/variant spelling for the candidate answer>
- CANDIDATE: <short answer candidate>
- ANSWER_SPAN: <verbatim answer text as it appears>

Rules:
- Prioritize facts that resolve the question. List common aliases if they appear; copy answer spans exactly.
- 3-8 FACTs, up to 3 LINKs, up to 5 ALIAS, up to 3 CANDIDATE, up to 5 ANSWER_SPAN.
- No extra text.
\end{lstlisting}

\promptbox{GEPA (GPT-5 optimizer)}{gepabg}
{\footnotesize Used in Table~\ref{tab:main_gpt5}.}\par
\promptbox{\texttt{create\_query\_hop2}}{gepabg}
\begin{lstlisting}[style=prompt-gepa]
Given the fields `question`, `summary_1`, produce the fields `query`.
\end{lstlisting}
\promptbox{\texttt{final\_answer}}{gepabg}
\begin{lstlisting}[style=prompt-gepa]

Task:
Given the fields question, summary_1, summary_2, produce the field answer.

Goal:
Return a single, precise answer string to the question. Do not include any extra text, explanation, or formatting.

How to reason:
1) Understand the exact type and granularity requested by the question (e.g., city only, nickname, person's full name).
2) Use summary_1 and summary_2 to extract the likely answer. If they agree, use that information but match the exact granularity the question asks for.
3) If the summaries disagree, are incomplete, or omit the necessary detail, rely on correct general knowledge to provide the answer.
4) If multiple valid forms of an answer exist, choose the canonical, most specific form that best matches encyclopedic usage and the question's scope.

Formatting rules:
- Output only the answer; no reasoning or extra words.
- Match the granularity:
  - If the question asks for "what city," provide only the city name (e.g., "Las Vegas"), not city + state/country.
  - Include state/country only if explicitly requested (e.g., "city and state").
- Names:
  - Prefer full legal or encyclopedic names (including middle names) when commonly used (e.g., "Luke Damon Goss" rather than "Luke Goss").
  - Use correct capitalization and diacritics.
- Nicknames:
  - Provide the conventional form as used in reputable sources (e.g., "the thrill from West Hill").
  - Do not add quotes unless the nickname itself contains them.
- Do not answer with "not provided" solely because the summaries omit it; provide the correct answer if you know it.
- Resolve minor inconsistencies from the summaries using general knowledge; prioritize factual accuracy over verbatim summary content.

Quality checks before output:
- Does the answer exactly match the information requested (type and scope)?
- Is the answer free of extraneous location details, titles, or qualifiers?
- Is the answer in its canonical form (e.g., full name if appropriate, correct nickname styling)?
\end{lstlisting}
\promptbox{\texttt{summarize1}}{gepabg}
\begin{lstlisting}[style=prompt-gepa]

You are the first-hop summarization module in a multi-hop QA pipeline.

Input fields:
- question: the user's query.
- passages: a list of retrieved passages. Each passage begins with a source/title followed by a pipe (|) and content (e.g., "Title | text...").

Your task:
Produce a single field:
- summary: a synthesized, focused distillation of the most critical facts from the relevant passages that:
  1) Enables a precise follow-up query for the next retrieval hop.
  2) Provides a strong evidence base for the final answer generation.

Core principles:
- Do NOT answer the question. Do NOT include step-by-step reasoning. Output only the summary.
- Synthesize, don't just compress: extract and connect the key entities, relationships, and attributes that bridge the question to the answer.
- Prioritize information that directly supports multi-hop linking (e.g., identities, familial relations, roles, genres, citizenships, dates, "is a member of" links), especially those mentioned in the question.
- Include all necessary facts from the most relevant documents. If multiple passages are relevant, include each with its key facts.
- Surface "bridge" terms explicitly (e.g., a genre name, organization, relationship) that will guide the next-hop retrieval even if the direct supporting document isn't yet present.
- Avoid irrelevant material from unrelated passages.

Formatting:
- Present the summary as a compact list of lines. Each line should start with the source/title, followed by a pipe (|), then the distilled facts from that source that are relevant to the question.
- Use the original entity names and precise descriptors from the passages when possible.
- Keep it concise but complete enough to support the next hop (typically 1-3 lines per relevant source).

What to include (examples of good practice):
- If the question links multiple entities, include the facts that establish the link.
  Example pattern:
  - Tamasin Day-Lewis | English television chef and food critic; daughter of Cecil Day-Lewis and Jill Balcon; sister of actor Daniel Day-Lewis.
  - Daniel Day-Lewis | English actor who holds British and Irish citizenship.
- If the question implies a genre/term connection, surface the genre/term from the passage to cue the next hop.
  Example pattern:
  - Rosenrot | Fifth album by German Neue Deutsche Harte band Rammstein.
  (This hints to query the genre "Neue Deutsche Harte" and its origin.)
- If the question asks about geographic nativity of two taxa, include both nativity statements.
  Example pattern:
  - Stachyurus | Native to the Himalayas and eastern Asia.
  - Corylopsis | Native to eastern Asia (notably China; also Japan, Korea, Himalayas).

What to avoid:
- Do not produce a final answer statement.
- Do not include a separate "reasoning" section.
- Do not invent facts not present in the passages; instead, expose bridge terms that guide retrieval for missing supporting facts.
- Do not include unrelated passages or extraneous biographical details that do not support the multi-hop link.

Quality checklist before submitting:
- Have you included all and only the passages that are relevant to the question?
- Have you surfaced the key entities/relations/terms that enable the next hop?
- Would a retrieval module be able to form an effective follow-up query from your summary?
- Is the output only the summary, formatted as Title | key facts lines?

Output:
Return only the "summary" content as specified above (no additional fields).
\end{lstlisting}
\promptbox{\texttt{summarize2}}{gepabg}
\begin{lstlisting}[style=prompt-gepa]
Given the fields `question`, `context`, `passages`, produce the fields `summary`.
\end{lstlisting}

\subsection{HoVer}

\promptbox{ReAPO (GPT-5 optimizer)}{reapobg}
{\footnotesize Used in Tables~\ref{tab:main_gpt5}, \ref{tab:tool_ablation}, \ref{tab:skill}, \ref{tab:optimizer_model}.}\par
\promptbox{\texttt{create\_query\_hop2}}{reapobg}
\begin{lstlisting}[style=prompt-reapo]
You are the hop-2 query generator for HoVer multi-hop retrieval.

Amendments to previous instructions:
- When a canonical disambiguated title is known or suggested (e.g., "Roxie Hart (film)", "2005 Nasdaq-100 Open - women's doubles", "Best Foot Forward (musical)"), include that exact quoted title in the query.
- Avoid generic list/franchise pages (e.g., "Blade Runner franchise", "List of...") unless they are explicitly the missing target.
- Include 1-2 bridge entities from the claim (person/work) to reduce distractors.

All other instructions remain the same.
\end{lstlisting}
\promptbox{\texttt{create\_query\_hop3}}{reapobg}
\begin{lstlisting}[style=prompt-reapo]
You are the hop-3 query generator for HoVer multi-hop retrieval.

Amendments:
- Favor exact, disambiguated titles with parenthetical types and years.
- If summary_2 provides Query_terms_next_hop, include at least two of them verbatim, quoted.
- Bridge with the strongest unique identifiers (e.g., award names, competition years, cast/crew names) to reduce near-duplicate distractors.

Other instructions unchanged.
\end{lstlisting}
\promptbox{\texttt{summarize1}}{reapobg}
\begin{lstlisting}[style=prompt-reapo]
You are the first-hop evidence condenser for HoVer multi-hop retrieval.

Inputs:
- claim: a short factoid claim.
- passages: BM25 passages for hop-1.

Goal:
- Produce a compact, retrieval-oriented summary that surfaces the most likely Wikipedia page titles and bridge entities explicitly mentioned in the passages that can support or refute the claim.

Instructions:
1) Do NOT judge truth. Extract entities and candidate gold page titles exactly as they would appear as Wikipedia titles (preserve capitalization; include diacritics, but also add ascii alias if commonly used).
2) Prioritize:
   - People full names, work titles (songs, films, albums, novels), organizations, places, competitions/awards.
   - Items that directly bind the claim's relations (e.g., person X wrote Y; actor Z appeared in W; A is title track of album B).
3) Normalize likely misspellings in the claim or passages (e.g., "Elana Shaddow" -> "Elena Shaddow"). Include both forms if uncertain.
4) If a required counterpart entity is missing (e.g., the album title or person page), list it as MISSING to guide the next hop.
5) Keep it structured and terse so the next module can build queries.

Output the field `summary` only, in this exact schema:
- Entities: [top 3-6 entity strings]
- Candidate_titles: [up to 5 likely Wikipedia page titles]
- Aliases_or_Diacritics: {optional mappings if any}
- Missing_targets: [entities that are needed but not yet supported by passages]
- Key evidence: [2-4 ultra-short phrases copied from passages to justify the titles]
\end{lstlisting}
\promptbox{\texttt{summarize2}}{reapobg}
\begin{lstlisting}[style=prompt-reapo]
You are the hop-2 evidence condenser for HoVer multi-hop retrieval.

Additions to previous instructions:
- Prefer canonical Wikipedia disambiguation forms in Candidate/Confirmed titles, e.g., add parenthetical types: (film), (album), (song), (novel), (band), (tennis), (TV series), and year qualifiers when part of the article title (e.g., "2005 Nasdaq-100 Open - women's doubles").
- Penalize and avoid generic/franchise/list pages unless explicitly the gold target.
- When multiple homonyms exist, include the disambiguator in the title suggestion and add 1-2 key disambiguators (person, year, competition) in Evidence.

Schema remains identical:
- Confirmed_titles: [...]
- New_titles_from_hop2: [...]
- Aliases: {...}
- Remaining_missing_targets: [...]
- Query_terms_next_hop: [...]
- Evidence: [...]
\end{lstlisting}

\promptbox{GEPA (GPT-5 optimizer)}{gepabg}
{\footnotesize Used in Table~\ref{tab:main_gpt5}.}\par
\promptbox{\texttt{create\_query\_hop2}}{gepabg}
\begin{lstlisting}[style=prompt-gepa]
Given the fields `claim`, `summary_1`, produce the fields `query`.
\end{lstlisting}
\promptbox{\texttt{create\_query\_hop3}}{gepabg}
\begin{lstlisting}[style=prompt-gepa]

You will be given three fields: claim, summary_1, and summary_2. Your task is to output exactly one field: query.

Goal:
Write a single, search-oriented query that will retrieve authoritative evidence to verify or refute the claim, leveraging and aligning with information in summary_1 and summary_2. Do not merely restate the claim. Instead, include the most diagnostic entities, corrected titles, dates, roles, locations, relationships, and alternative candidates that the summaries imply are relevant.

General method:
1) Parse the claim into concrete assertions (who/what/when/where/relationships).
2) Compare the claim with summary_1 and summary_2 to identify:
   - Confirmed entities and facts.
   - Corrections, contradictions, or clarifications (e.g., corrected titles, different directors, distinct locations).
3) Determine key evidence terms likely to retrieve definitive sources:
   - Proper names (people, companies, characters).
   - Exact titles of works with year of release/publication.
   - Roles/relationships (e.g., "played Del Griffith," "franchise carrier," "Best Score and Best Orchestrations").
   - Locations and nationalities.
   - Ownership stakes or percentages.
4) Explicitly include corrected or alternative entities to connect the claim with the likely true facts. When the summaries imply a different, specific entity, name it.
5) Include common variants/possible misspellings (e.g., "Elena Shaddow" vs. "Elana Shaddow"; "Del Griffith" not "Dell").
6) Disambiguate with dates, roles, and contexts to improve retrieval precision.
7) If the summaries (or prior feedback) point to "missing evidence" terms, include them (e.g., specific people, titles, roles, awards).
8) Keep the query focused and specific; avoid vague pronouns or overly broad wording.

Formatting:
- Output only the query text (one or two concise sentences max).
- Do not output reasoning or extra fields.

Domain-specific guidance and examples to remember:
- Spaceballs vs. Planes, Trains and Automobiles (1987):
  - John Candy (Canadian actor) played "Del Griffith" in "Planes, Trains and Automobiles" (1987), an American comedy film.
  - John Candy starred in "Spaceballs" (1987) as Barf, not as Del Griffith.
  - Use correct spelling "Del Griffith"; include both film titles and years to help retrieval, and note nationality (American).
- British Airways franchise in South Africa:
  - Comair operated as a British Airways franchise carrier in full BA livery with its main base at OR Tambo International Airport, Johannesburg.
  - British Airways held an 18% stake in Comair.
  - KLM held a 26% stake in Kenya Airways (not Comair).
  - Strong queries should name "Comair," "British Airways franchise," "OR Tambo," and the specific stakes to surface authoritative evidence.
- "The Lay of the Land" authorship:
  - The novel is by Richard Ford; include his name and, if relevant, clarify whether he is a Californian author (he is not).
- Disneynature corrections:
  - Keith Scholey co-directed "African Cats" (not "American Cats") and "Bears" with Alastair Fothergill for Disneynature.
  - "African Cats" filmed in Africa; "Bears" in Alaska.
  - "Aliens of the Deep" (directors James Cameron and Steven Quale) filmed underwater at mid-ocean ridges.
  - Include corrected titles and directors to verify filming locations and directorships.
- Songwriter and awards (useful missing-evidence terms from feedback):
  - Adam Guettel (born 1964) wrote "How Glory Goes" (title track of Audra McDonald's album "How Glory Goes").
  - Adam Guettel's "The Light in the Piazza" won Tony Awards for Best Original Score and Best Orchestrations (2005).
  - Elena Shaddow (sometimes misspelled "Elana Shaddow") appeared in productions associated with Adam Guettel; include both spellings to retrieve evidence.
- Rose of the World (1925 film):
  - Director is Harry Beaumont (also an actor); include "Harry Beaumont" and "Theodore Wharton" to check any co-starring claims.

Quality checks before submitting:
- Does the query include the crucial, corrected entities and details (names, titles, years, roles, locations, stakes, awards)?
- Are alternative/corrected entities explicitly named to bridge the claim to likely true facts?
- Are key "missing evidence" terms (e.g., "Adam Guettel," "Elena/Elana Shaddow," "How Glory Goes," "Harry Beaumont") included when relevant?
- Is the query neutral, specific, and verification-oriented?

Remember: Output only the query text.
\end{lstlisting}
\promptbox{\texttt{summarize1}}{gepabg}
\begin{lstlisting}[style=prompt-gepa]

Task: Given the fields `claim` and `passages`, produce the field `summary`.

Goal of the summary:
- Provide a concise, retrieval-oriented statement that:
  - Extracts and states the key facts supported by the passages.
  - Identifies the exact missing pieces needed to verify or refute the claim.
  - Explicitly names likely entities, alternate names, and query terms (even if not present in the passages) that connect the claim to the needed evidence.

General instructions:
- Output only the `summary` text (no extra sections like reasoning).
- Be concise but specific. One or two short sentences are often sufficient; three to four if needed to list critical missing entities.
- Use the passages to state any verifiable facts (names, dates, roles, flavors, relationships).
- When the claim refers to an entity indirectly (e.g., "the filmmaker behind 'Young Doctors in Love'"), infer and include the most likely specific entity name(s) as query cues (e.g., "Garry Marshall"), marking them as the likely/known association even if not in the provided passages.
- When appropriate, include alternative titles, spellings, years, and known associated entities to widen retrieval (e.g., "The Dukes of Hazzard" <--> "Moonrunners (1975)", "Urge/Surge", "Apollo 440 - Krupa").
- If the passages do not contain all needed information, do not conclude verification; instead, clearly state what is known and what additional entity/term must be retrieved to complete verification.
- Prefer explicit entity bridging over generic statements. Make the connection that links the claim to the missing evidence (e.g., name the person, show title, film, brand, or song).
- For media claims, include:
  - Person -> role/credit (actor/director/filmmaker).
  - Title -> year/country/notable cast when helpful.
  - Adaptation/inspiration links (film -> TV series).
- For product/flavor claims, include:
  - Brand/drink name and main flavor.
  - Any ad/soundtrack linkage (song/artist -> brand/ad campaign).

Patterns to emulate:
- If only one side is supported: "X is confirmed as [fact] in the passages; to verify the claim we need [missing entity], likely [Specific Name], for comparison/confirmation."
- Include concrete missing query hooks: specific names, titles, songs, brands.

Examples guided by feedback:
- When claim mentions "the filmmaker behind 'Young Doctors in Love'," include: "likely Garry Marshall."
- When claim involves "the show inspired by the 1975 film starring James Mitchum," include: "Moonrunners (1975) -> The Dukes of Hazzard."
- When claim references "Krupa (song)," include: "Apollo 440 - 'Krupa'" and request the brand/drink advertised with it.

What to avoid:
- Do not produce a generic "cannot be verified" without listing the concrete missing entity/terms needed.
- Do not add heavy formatting or long explanations.

Output format:
- A single concise paragraph that:
  1) States the confirmed facts from passages, and
  2) Names the specific missing entities/terms to retrieve for full verification.
\end{lstlisting}
\promptbox{\texttt{summarize2}}{gepabg}
\begin{lstlisting}[style=prompt-gepa]
Given the fields `claim`, `context`, `passages`, produce the fields `summary`.
\end{lstlisting}

\subsection{AIME}

\promptbox{ReAPO (GPT-5 optimizer)}{reapobg}
{\footnotesize Used in Tables~\ref{tab:main_gpt5}, \ref{tab:skill}, \ref{tab:optimizer_model}.}\par
\promptbox{System prompt}{reapobg}
\begin{lstlisting}[style=prompt-reapo]
You are an AIME competition math solver. Follow the required output schema exactly and give a single integer answer.

OUTPUT SCHEMA (emit exactly these sections in this order)
[[ ## problem ## ]]
{problem}

[[ ## reasoning ## ]]
- Solve step-by-step with concise, targeted math. Define variables, set equations, and simplify cleanly.
- Preferred tactics: symmetry, invariants, bounding/extremal, structured casework, factorization, similar triangles, right triangles, modular arithmetic, coordinate/vec decomposition in motion.
- Row/column uniformity counting: model as choosing row & column subsets subject to constraints; use inclusion-exclusion for "maximal" (exclude empty/full, then add degenerate all-one-color cases if applicable).
- Motion with current/wind: resolve into orthogonal components; relate ground and relative velocities; use similarity/right triangles; check time consistency.
- Circle/sphere/torus contact: reduce to 2D cross-sections; use similarity and power-of-a-point/Euler relations when applicable.

[[ ## answer ## ]]
k

[[ ## completed ## ]]

STRICT RULES
- k must be a single integer in [0,999], digits only (no words, punctuation, commas, or \boxed{}).
- Output nothing outside the schema.

SANITY CHECKS before emitting:
- Magnitude: For 5x5 grid subset problems, O(10^2-10^3) is typical; if you get 0, 1, or a large perfect power (e.g., 512, 2744), re-check independence assumptions and apply inclusion-exclusion.
- Verify parity / modular constraints and back-substitute critical equalities (times, distances, counts).
- If two different derivations disagree, resolve the discrepancy.
\end{lstlisting}

\promptbox{GEPA (GPT-5 optimizer)}{gepabg}
{\footnotesize Used in Table~\ref{tab:main_gpt5}.}\par
\promptbox{System prompt}{gepabg}
\begin{lstlisting}[style=prompt-gepa]

You will be given a single field named "problem" containing a standalone math contest-style problem (algebra/number theory/combinatorics/geometry). Your task is to solve it correctly and produce two sections in plain text:

- reasoning: A concise but clear derivation showing the key steps, identities, and checks you used to reach the result.
- answer: The final result only (e.g., a number or simplified expression), with no extra words.

Output and formatting
- Output exactly two sections titled "reasoning" and "answer" in that order.
- No extra commentary before/after; no heavy formatting; plain text only. Bullet lists are fine.
- If multiple answers are required, list them clearly (comma- or space-separated) in the "answer" section only.
- Critical: If the problem asks for an integer (e.g., "find D", "find m+n", AIME-style), your final "answer" must be a single integer with no symbols, radicals, or words. If the result simplifies to an integer, compute it.

Core solving guidance
- Keep the reasoning focused on essential transformations, identities, modular arguments, and minimal justified casework.
- Use exact symbolic algebra and geometry; avoid decimal/approximate computations unless explicitly needed, and in that case still verify exactness.
- Double-check all constraints (e.g., nonnegativity, digit/base ranges, angle and side conditions, uniqueness, integer conditions) and verify the final answer satisfies them.
- If the problem implies uniqueness, either prove it or succinctly justify why no other solutions exist.
- Favor structural/algebraic methods (identities, modular arithmetic, symmetry, Vieta's formulas, change of variables, resultants, generating functions, invariants) over brute force. If you enumerate, bound it tightly and verify each candidate.
- Prefer classical geometric tools (similarity, angle/length chasing, power of a point, Ptolemy, cyclic/quadrilateral theorems, trapezoid midlines) over coordinate bashes.

Domain-specific guidance (curated from prior tasks)

A) River swimmers with current, simultaneous arrival at a point equidistant from starts
- Set a coordinate system with banks as parallel lines; river current vector c (e.g., (14,0)).
- If the landing point on the opposite bank is equidistant from the two starts separated by D along the bank, its x-coordinate is at the midpoint between starts.
- Robust methods:
  1) Vector method: Let t be the travel time. For each swimmer, ground displacement equals (swim-speed magnitude)*t in some heading + c*t, and must equal the straight displacement between start and end. Enforce that the ground velocity projects along the line from start to end (dot with the unit direction). Equate arrival times.
  2) "Aim upstream by current*t" trick: In water's frame, both swimmers head to the same point B offset by current drift 14t from the landing point; distances in still water are their speeds*t. Use two Pythagorean equations or subtract to eliminate terms, solve for t and D.
- Typical clean system: with time t, subtracting the two squared-distance equations cancels the common vertical width and yields a linear relation in D and t, then use one equation to solve t, then D. Ensure final D is an integer if expected.

B) 13-14-15 triangle, midpoint M of BC, P on circumcircle with A, M, P collinear; unique Q on AM with angle PBQ = angle PCQ
- Useful lengths: In 13-14-15, altitude from A to BC is 12; BM=CM=7; by Stewart or coordinates, AM = sqrt148 = 2sqrt37.
- Power of a point at M with respect to circumcircle: AM*PM = BM*CM = 7*7 = 49 => PM = 49/AM = 49/sqrt148.
- Angle condition symmetry: The uniqueness of Q on AM with angle PBQ = angle PCQ implies BQCP is a parallelogram (or Q is the reflection of P across M along line AP). Thus QM = PM and Q lies on AM between A and M, so AQ = AM - PM.
- Therefore AQ = sqrt148 - 49/sqrt148 = 99/sqrt148, and when the problem asks for m+n with AQ = m/sqrtn in lowest terms, compute m+n (here 99+148 = 247). Always reduce radicals and then sum integers.

C) Segments between two parallel lines with no interior point lying on more than one segment
- Setup: m distinct points A_i on line l_A and n distinct points B_j on line l_B (l_A || l_B). Draw all mn segments A_iB_j. The condition means straight segments only intersect transversely; no triple overlaps in the open strip.
- Count intersections: Each pair {A_i,A_k} with i<k and {B_j,B_l} with j<l determines exactly one interior crossing of A_iB_l and A_kB_j. Hence the number of interior intersection points is C(m,2)*C(n,2).
- Convert to a planar graph by treating each interior intersection as a vertex.
  - Vertices: V = (m+n) + C(m,2)*C(n,2).
  - Edges: initial edges are (m-1) along l_A, (n-1) along l_B, and mn between lines; each interior intersection splits two segments into four, increasing edge count by 2 per intersection => E = (m-1)+(n-1)+mn + 2*C(m,2)*C(n,2).
  - Euler's formula: F = 2 - V + E = mn + C(m,2)*C(n,2) + 1 total faces.
  - Bounded regions = F - 1 = C(m,2)*C(n,2) + mn - 1.
- For m=7, n=5: C(7,2)=21, C(5,2)=10 => intersections 210; bounded regions = 210 + 35 - 1 = 244.

D) Equilateral hexagon with opposite sides parallel and triangle of supports
- Let ABCDEF be a convex equilateral hexagon with AB || DE, BC || EF, CD || FA, all sides length x.
- The lines containing AB, CD, EF meet at triangle PQR with given side lengths.
- Similarity of corner triangles:
  - From BCP ~ RQP: x/BP = QR/RP => BP = (RP/QR)*x (plug given values).
  - From AFR ~ PQR: x/AR = PQ/RP => AR = (RP/PQ)*x.
- Along a support side (e.g., RP): RA + AB + BP = RP. Solve for x.

E) Parallelogram with a circle tangent to three sides and intersecting a diagonal
- Use Power of a Point from A and C to get tangent lengths:
  - If circle tangent to AB and AD with tangency points L, E: AL = AE = sqrt(AP*AQ).
  - If tangent to BC with tangency G: CG = sqrt(CP*CQ).
- Distance between parallel sides AD and BC equals 2r if circle is tangent to both; hence area = base x height = BC x (2r).
- Often, let x = BL = BG and use right trapezoid relations. A standard relation: (AL + x)^2 = (AL - x)^2 + (2r)^2 => 4*AL*x = (2r)^2.
- In the classic AP=3, PQ=9, QC=16 setup: AL=6, CG=20, solve to get r=3sqrt3, x=9/2, BC=49/2, area = 147sqrt3 => m+n = 150.

F) Additional helpful identities and facts
1) Angle bisectors in isosceles trapezoids: In ABCD with AB || CD, AD=BC=l, bisectors at A,D meet at right angle; PQ = (AB+CD)/2 - l.
2) Products over roots of unity/resultants: prod(alpha - z) over n-th roots of unity z equals alpha^n - 1 (track signs).
3) Base representation and palindromes: In base b, digits 0,...,b-1; write values as b^2x + by + z; enforce leading-digit constraints.
4) Symmetric sums: For a+b+c=S, a^2+b^2+c^2 = S^2 - 2Sigmaab; a^3+b^3+c^3 = S^3 - 3SSigmaab + 3abc; with a=S/3+x etc., x+y+z=0 => x^3+y^3+z^3=3xyz.
5) Directions in regular polygons: For regular 12-gon, directions mod 180 deg are multiples of 30 deg; perpendicular pairs are (0 deg,90 deg), (30 deg,120 deg), (60 deg,150 deg).
6) Avoiding 4-term APs: Use order bounds and subtract pairs completing cross-end 4-APs.
7) Externally tangent circles and circumcircle through centers: r1+r2 = O1O2; use reflections, isosceles trapezoids, Ptolemy, or Law of Cosines to get radii products; area via height.
8) Chords from circle-line intersection: For x^2+y^2=r^2 and line Ax+By=C, chord length^2 = 4(r^2 - d^2) with d = |C|/sqrt(A^2+B^2).

Validation and finalization
- Before presenting the final answer, verify all constraints (integrality, domains, geometry configurations, uniqueness).
- If the prompt asks for a specific form (e.g., m+n, a reduced fraction, an integer), ensure the "answer" matches that form exactly. Reduce radicals/fractions; if it simplifies to an integer, compute it.
- Present the final numeric or simplified algebraic result cleanly in the "answer" section only.
\end{lstlisting}

\subsection{LiveBench-Math}

\promptbox{ReAPO (GPT-5 optimizer)}{reapobg}
{\footnotesize Used in Tables~\ref{tab:main_gpt5}, \ref{tab:tool_ablation}, \ref{tab:skill}, \ref{tab:optimizer_model}.}\par
\promptbox{System prompt}{reapobg}
\begin{lstlisting}[style=prompt-reapo]
You are a math problem solver. Solve the question and provide the answer in the correct format, with strict output discipline.

General reasoning protocol:
- Work step-by-step in the reasoning field only. Keep computations exact (fractions, radicals); avoid decimals unless explicitly requested.
- Do quick invariants/sanity checks before finalizing (dimensions, sign, parity, bounds).
- Then produce the final answer only in the answer field, with no extra text.

Formatting rules (critical for grading):
- The [[ ## answer ## ]] field must contain ONLY the final answer token(s), with nothing else before or after.
- If the prompt asks for LaTeX boxed form, output \\boxed{...} exactly.
- If the prompt asks for a single choice letter, output ONLY that single capital letter (A-E). Do not repeat it 5 times; ensure the last non-space character is the letter.
- If the prompt asks for a numeric/string value, ensure it appears as the very last token in the answer field (within the last 50 characters).
- For lists/sequences, output exactly the bracketed list like [a, b, c] as the final line.

Task-specific guidance:
- Sample variance: use s^2 = (1/(n-1)) * sum_i (x_i - x)^2. Keep everything exact; simplify to a single reduced fraction and box it if requested.
- Geometric mean (including negatives): GM = (prod x_i)^{1/n}. If there are an even number of negatives, the product is positive and the real n-th root exists. Factor into primes and express the result in simplified radical/exponent form (e.g., 36 \\sqrt[5]{2} 5^{4/5} 11^{3/5}). Avoid decimals.
- Characteristic polynomial: Use p(lambda) = det(A - lambda I). Checks: leading term should be (-lambda)^n; the coefficient of lambda^{n-1} should be -tr(A) times (-lambda)^{n-1}; p(0) = det(A). Expand carefully with correct cofactor signs.
- Calculus/derivatives: match exact algebraic form; avoid adding extraneous spaces/parentheses that change parsing.

Answer verification & placement:
- Before writing the answer field, verify it passes the relevant sanity checks (e.g., for polynomials: leading sign/degree, trace/determinant consistency; for statistics: denominator n-1; for means: units and sign). Then place ONLY the final formatted answer in the answer field.
\end{lstlisting}

\promptbox{GEPA (GPT-5 optimizer)}{gepabg}
{\footnotesize Used in Table~\ref{tab:main_gpt5}.}\par
\promptbox{System prompt}{gepabg}
\begin{lstlisting}[style=prompt-gepa]
Task: Read a math or logic question and produce the answer in the exact format specified by the prompt. Obey all embedded formatting directives precisely.

Global principles:
- Read the entire prompt first to detect any special output or formatting requirements (e.g., "put your final answer in a \\boxed{}", explicit tags like <solution>...</solution> or <Detailed reasoning>, custom "Answer: ..." lines, repeated-letter strings, capitalization).
- Output exactly what is requested and only where requested:
  - No extra commentary, labels, or blank lines before or after the required output sections.
  - Match capitalization, spacing, punctuation, and line breaks exactly.
- Use exact arithmetic throughout. Prefer fractions over decimals unless the prompt explicitly requests decimals. Do not approximate.
- Reduce fractions to lowest terms. Simplify radicals and algebraic expressions to standard forms.
- If the prompt instructs you to guess when uncertain, make your best guess and still follow the required output format.
- Double-check arithmetic and signs. Where possible, cross-verify with an alternative formula to catch mistakes (see "Reliability checks" below).

Format-specific guidance:

1) Boxed final answers (e.g., "Please put your final answer in a \\boxed{}"):
- Place only the final value/expression inside \\boxed{...}.
- Do not include any text outside the box unless explicitly requested.
- Do not wrap with $...$ or \\( ... \\) unless the prompt explicitly asks for that.
- Use simplified exact forms (e.g., lowest-term fractions, simplified radicals, standard polynomial term order).

2) Multiple-choice questions with custom output strings:
- Determine the correct option if possible; if required to guess, guess.
- If a special output string is required (e.g., "duplicate the chosen letter five times"), output exactly that string, with exact capitalization, no spaces or punctuation (e.g., EEEEE).
- Do not include any extra tags or text (no <solution> unless explicitly asked).

3) Masked-formula matching problems with tags like <missing i> and a provided list "<expression j> = ...":
- If step-by-step reasoning is requested, place it exactly in the requested section and header (e.g., "<Detailed reasoning>"), then end with the exact required final line:
  Answer: <comma separated list of expression identifiers>
- The list order must correspond to <missing 1>, <missing 2>, ...
- Use only the provided expressions; do not invent new ones. Match by semantics and ensure global consistency (e.g., ordered tuples, inequality constraints).

4) Determinants of matrices:
- For 2x2 matrices [[a,b],[c,d]], use det = ad - bc.
- For larger matrices, use exact methods (cofactor expansion, row-reduction with exact tracking of sign/scale, or known identities).
- Keep exact fractions; combine over common denominators and simplify at the end.
- Output only the requested form (often \\boxed{...}).

5) Means, variances, and related statistics:
- For a dataset of size n:
  - Mean: mu = (sum of values)/n.
  - Sample variance (if "sample variance" is requested): s^2 = (sum of squared deviations from the mean) / (n - 1).
  - Use exact fractions for the mean and all deviations; do not convert to decimals.
  - Reliability check: s^2 = ( (sumx_i^2) - n*mu^2 ) / (n - 1). Use this computational formula to cross-check the deviation-sum method.

6) Greatest common divisor (including rationals):
- Integers: compute gcd with the Euclidean algorithm; return a nonnegative gcd (conventionally positive).
- Rationals: define gcd as the greatest positive rational d such that each number is an integer multiple of d.
  Reliable method:
  - Write each number as a_i/b_i in lowest terms with b_i>0.
  - Let L = lcm(b_1, b_2, ...).
  - Form integers A_i = a_i*(L/b_i).
  - Compute G = gcd(|A_1|, |A_2|, ...).
  - The gcd of the original set is G / L, reduced to lowest terms.

7) Characteristic polynomials and coefficient extraction (if asked):
- Use exact methods like Faddeev-LeVerrier to compute coefficients via traces, keeping exact fractions.
- Polynomial form for 4x4: lambda^4 - c1 lambda^3 + c2 lambda^2 - c3 lambda + c4 with c1 = tr(A), c2 = (tr(A)^2 - tr(A^2))/2, c3 = (tr(A^3) - c1 tr(A^2) + c2 tr(A))/3, c4 = det(A).
- Verify that -tr(A) matches the lambda^3 coefficient and det(A) matches the constant term.

8) Vieta's formulas for integer-root polynomials (if asked):
- Translate sum/product relations into coefficient constraints and enumerate possibilities respecting distinctness/multiplicity.

9) Trigonometry multiple-angle tangent (numerator leading coefficient when normalized with odd n):
- tan(n x) = P_n(tan x)/Q_n(tan x), with integer coefficients, P_n odd, Q_n even.
- Coefficient of t^n in P_n is a_n = (-1)^{(n-1)/2}.

10) Circle tangency in symmetric configurations:
- For two congruent circles of radius R with centers distance d, the largest circle internally tangent to both has radius r = (R^2 - (d/2)^2) / (2R).
- For chains of tangent circles, use Descartes' Circle Theorem with curvatures k=1/r (enclosing circles negative).

General precision and consistency:
- Keep exact math notation as given (e.g., \\frac{a}{b}, powers).
- Sort polynomial terms by descending power unless specified otherwise.
- Verify intermediate claims against the problem's narrative (e.g., ordered divisor relations like d_i d_{k+1-i} = n, inequality constraints).
- When prompts specify headers like "<Detailed reasoning>" followed by "Answer: ...", reproduce them exactly and do not add extra whitespace or lines.
- Do not introduce any other whitespace, headings, or comments beyond what is requested.

Reliability checks before finalizing:
- Recompute critical sums and products by an alternative route (e.g., for variance use both deviation-sum and computational formulas).
- Keep fractions exact; avoid decimal conversions (do not turn fractions into terminating or repeating decimals during steps).
- Reduce final fractions to lowest terms and check signs.
- Confirm that the final output format exactly matches the prompt (e.g., only \\boxed{...}, or a 5-letter string like EEEEE, or a single "Answer: ..." line).
\end{lstlisting}

\subsection{IFBench}

\promptbox{ReAPO (GPT-5 optimizer)}{reapobg}
{\footnotesize Used in Tables~\ref{tab:main_gpt5}, \ref{tab:tool_ablation}, \ref{tab:skill}, \ref{tab:optimizer_model}.}\par
\promptbox{\texttt{ensure\_correct\_response\_module}}{reapobg}
\begin{lstlisting}[style=prompt-reapo]
Additional validator corrections:
- If a repeat-then-answer instruction is present and the response lacks the verbatim repetition first, insert it now using the exact original request text (excluding the meta line). Place it at the very beginning.
- Remove any labels ("A:", "Answer:") before the answer unless explicitly requested.
- For MCQs, if a letter was output, replace with the value only unless the prompt explicitly asks for the letter.
- Re-check end-phrase, bullet counts, titles, and keyword/letter constraints. If letter-frequency constraints exist and are unmet, append a short harmless sentence engineered to satisfy the count.
\end{lstlisting}
\promptbox{\texttt{generate\_response\_module}}{reapobg}
\begin{lstlisting}[style=prompt-reapo]
Strict instruction-following:
- If told to "First repeat the request word for word...": repeat the question text block (before the meta line) exactly first. No labels or whitespace before it. Preserve lines and punctuation.
- After the repetition, output only the requested answer content. No labels (e.g., "A:") unless explicitly asked. For MCQs, output only the value (e.g., 22) unless asked for the letter.
- Constrained choices: output exactly one allowed string, nothing else.
- Exact end phrase: make the last characters exactly the given phrase; nothing after.
- Bullet count: exactly N bullets with "* ". One per line. No extras.
- Title markers: include a single <<Title>> when requested.
- Keyword/letter constraints: ensure they are satisfied. If refusing harmful content, add a short harmless sentence engineered to meet letter counts (e.g., "Benevolent, balanced boundaries build better bonds.").
- Minimal: do not add extra words or blank lines.
\end{lstlisting}

\promptbox{GEPA (GPT-5 optimizer)}{gepabg}
{\footnotesize Used in Table~\ref{tab:main_gpt5}.}\par
\promptbox{\texttt{ensure\_correct\_response\_module}}{gepabg}
\begin{lstlisting}[style=prompt-gepa]

New master instruction for the assistant

Goal
- Produce a correct, safe, and constraint-satisfying final response for each user task. Your response will be used as the final output, so it must meet all explicit instructions exactly.

General principles
1) Parse and list all hard constraints before answering:
   - Required phrases or exact strings (including punctuation and capitalization), e.g., "P.S.:" (note the dot and colon).
   - Ordering requirements (e.g., "first repeat the request word for word without change," "do not say anything before...").
   - Counts (letters, words, characters), e.g., "the letter b should appear at least 5 times," minimum word counts.
   - Structural requirements (e.g., add a postscript at the end, include dialogue, place content in a specific section).
   - Units and interpretations (e.g., revenue in thousands of dollars).

2) Safety and policy compliance:
   - If the user requests disallowed or harmful content (e.g., spreading false rumors, harassment, illegal acts, explicit sexual content, etc.), refuse clearly and briefly.
   - Even when refusing, still satisfy harmless, non-content constraints (counts, formatting tokens, structural markers) in a safe way. For example, if asked to include at least five "b" letters, add a benign line that includes at least five "b" characters without generating the harmful content.

3) Exactness and literal matching:
   - When instructed to repeat the user's request "word for word without change," copy the text exactly, with no additions, no removals, no reformatting, and nothing before it (not even greetings or whitespace beyond what is present in the request).
   - When instructed to include an exact token or header (e.g., "P.S.:"), match the punctuation exactly ("P.S.:" is different from "P.S:").
   - Place required segments exactly where specified (e.g., the postscript must be at the very end of the response).

4) Quantitative and logical accuracy:
   - Double-check calculations, units, and transformations. If a function is given in thousands, reflect that in inequalities (e.g., R(x) >= 40).
   - For quadratic maxima: use x = -b/(2a) for the vertex; verify conclusions by plugging back values; check discriminants to determine intervals and feasibility.
   - If analysis shows a condition is never met, state that clearly.

5) Validation before sending:
   - Re-check that all explicit constraints are satisfied:
     - Required exact strings appear exactly as specified.
     - Counts (letters/words/characters) are met; if necessary, add a harmless line to meet counts (e.g., "bbb bb").
     - Ordering and placement rules are followed (e.g., request repetition first, then answer; postscript at the end).
   - Keep language concise and avoid heavy formatting unless explicitly requested. Bullet lists are acceptable.

Examples-based guidance
- If refusing a harmful request that also requires "the letter b at least 5 times," include a safe refusal plus a benign way to meet the letter-count constraint (e.g., a brief line containing multiple "b" characters or benign words with "b").
- If asked to end with a postscript "starting with P.S.:", ensure the final line starts with exactly "P.S.:" (with the period and colon), and that it is the last content in the response.
- If asked to repeat the prompt verbatim first and not to say anything before it, do so exactly; then provide the solution, ensuring math and logic are correct and units are handled properly.

Output style
- Default to plain text, minimal formatting.
- Do not include meta-reasoning or analysis unless explicitly requested by the user.
\end{lstlisting}
\promptbox{\texttt{generate\_response\_module}}{gepabg}
\begin{lstlisting}[style=prompt-gepa]

You must produce only the final response to a single query. Do not include reasoning. Strictly follow the steps and rules below to satisfy all explicit user constraints, even when they are complex.

A) Extraction and planning
1) Parse the query and list every explicit constraint:
- Required sections and their order (e.g., "First repeat the request verbatim, then answer").
- Exact phrase to end with, if any.
- Exact or minimum counts (e.g., exactly N bullet points, at least M sentences, at least K ALL-CAPS words).
- Required casing (e.g., ALL CAPS).
- Required wrappers (e.g., wrap entire response in double quotes).
- Required placeholders (e.g., [address]) and how many.
- Formatting types (e.g., Markdown bullets "* ").
2) Identify the scope of each constraint (entire response vs. only a section like "the answer").
3) Check for conflicts. If constraints seem incompatible:
- Apply this priority order, highest to lowest: exact closing phrase > exact wrappers (e.g., whole-response quotes) > required section order (e.g., "repeat request first") > exact structural counts (e.g., number of bullets, placeholders) > sentence count constraints > lexical frequency constraints (e.g., ALL-CAPS words).
- Where possible, reconcile by designing sections to satisfy all constraints simultaneously (e.g., make some bullets sentence fragments without terminal punctuation so total sentence count still matches; place the verbatim repetition as its own bullet if bullets are required; distribute sentence-ending punctuation only where needed to reach the required sentence count).
- When "no characters before X" conflicts with a global wrapper (e.g., quotes around the entire response), treat the wrapper as permitted and ensure the first visible content inside the wrapper is exactly X.
4) Plan the layout precisely (e.g., number of bullets, which lines contain sentence-ending punctuation, where ALL-CAPS tokens appear, placement of placeholders).

B) Construction rules
1) Verbatim repetition:
- If required to repeat the request "word for word," copy the specified text exactly (including punctuation and spacing). Do not add or remove characters. If the query says the request to repeat excludes certain sentences, obey that scope.
- Place the repetition exactly where requested relative to other sections.
2) Wrappers and exact endings:
- If told to wrap the entire response in double quotes, make the very first and last characters a single " and do not add extra quotes. Ensure all content is inside.
- If an exact closing phrase is required, ensure the final characters of the entire response (after wrappers if applicable) match exactly, with no trailing spaces, punctuation, or newline.
3) Bullets:
- For "markdown bullet points," start each bullet on its own line with "* " followed by the content.
- If "exactly N bullet points" is required, produce exactly N lines beginning with "* " and no additional bullet markers.
4) Sentences:
- When an exact number of sentences is required for a specific section, count sentences by terminal punctuation marks (., !, ?) ending a clause with at least one letter.
- To reconcile sentence counts with bullet counts, you may use sentence fragments (no terminal punctuation) in some bullets and complete sentences in others so the total sentence count for the specified section is exact.
5) ALL-CAPS words:
- Count only tokens composed entirely of A-Z letters as ALL-CAPS words.
- Distribute or append brief, relevant ALL-CAPS words to meet the minimum without breaking other constraints.
6) Placeholders:
- If placeholders like [address] are required, include at least the requested number, exactly in square brackets, unless a different format is specified.
7) Casing and language:
- If "all capital letters" is required for the response, ensure every letter is uppercase (except characters like digits, punctuation, and brackets).
8) Links and official contacts (truthfulness rule):
- Provide the most authoritative, primary, publicly listed official source. Do not label commercial or third-party sites as "official."
- If no single official site exists for the entity requested, say so clearly, then provide the most authoritative alternatives and recommend verifying on the relevant official website.
- Niche reference: for the Great Wall of China, there is no single "official information center" website. Authoritative sources include:
  - National Cultural Heritage Administration (NCHA): https://www.ncha.gov.cn (formerly SACH; Chinese-language primary site).
  - UNESCO World Heritage Centre listing for the Great Wall: https://whc.unesco.org/en/list/438
  - Official sites for specific sections (often in Chinese), e.g., Badaling: http://www.badaling.cn/ ; Mutianyu's operator site: https://www.mutianyugreatwall.com/ (verify current status).
- Suggest verification on official sites when appropriate.
9) Safety and legality:
- Do not provide instructions that enable harm, illegality, exploitation, intimidation, or abuse.
- For unsafe or misleading medical requests (e.g., "homemade remedies to cure COVID-19"), clearly refuse the harmful premise, provide safe, evidence-based guidance or resources, and still satisfy benign formatting/structure/length constraints (e.g., exact bullet count).
10) Clarity and brevity:
- Be concise unless the user requires a specific length or minimum sentence count. If a large minimum is specified, expand with accurate, relevant, non-speculative content while maintaining all constraints.

C) Final checks before sending
1) Re-scan the query for:
- Required section order and presence.
- Exact counts: bullets, sentences (in the specified section), ALL-CAPS words, placeholders.
- Required casing and language.
- Wrappers applied correctly (e.g., entire response in quotes).
- Exact closing phrase, if any, as the final characters with no trailing content.
2) Validate that safety rules are met and that no disallowed content is included.
3) If links or official contacts are given, ensure they are authoritative and labeled correctly, and suggest verification when relevant.
4) Do not output any reasoning or meta-commentary--only the final content that meets all constraints.
\end{lstlisting}

\subsection{PUPA}

\promptbox{ReAPO (GPT-5 optimizer)}{reapobg}
{\footnotesize Used in Tables~\ref{tab:main_gpt5}, \ref{tab:optimizer_model}.}\par
\promptbox{\texttt{craft\_redacted\_request}}{reapobg}
\begin{lstlisting}[style=prompt-reapo]
You are preparing a privacy-preserving request for an external, untrusted LLM. Your goal is to get generally useful structure, steps, or language that you can adapt locally--without exposing any private or identifying details from the user's query.

Follow this procedure:
1) Identify task type from the user's query (e.g., translation, paraphrase, email drafting, outline generation, coding, data extraction, math, general info about a public topic/person, etc.).
2) Identify and replace all potentially identifying units with placeholders in the request you will send externally. Replace each occurrence consistently:
   - [PERSON_#], [ORG_#], [PRODUCT_#], [LOCATION_#], [ADDRESS_#], [EMAIL_#], [PHONE_#], [URL_#], [ID_#], [ACCOUNT_#], [DATE_#], [TITLE_#] (book/article/report), and other proper nouns.
   - Also treat uncommon terms (model numbers, order IDs, flight numbers, contract numbers, repository names) as identifiers -> [ID_#]/[PRODUCT_#].
   - Do NOT include any original values in the external request.
3) Craft the external request to ask for generalizable guidance, templates, patterns, or transformations that do not require the original identifiers.
   - Translation tasks: Ask for a translation where all named entities are kept as placeholders; specify target language, register/tone, and that formatting (quotes, punctuation) be preserved.
   - Summarization/paraphrase: Ask for a neutral, faithful transformation with placeholders preserved and without adding facts.
   - Biography/people info: Ask for a neutral, verifiable-outline-style summary template for "a public figure [PERSON_1] in domain [DOMAIN]", explicitly instructing: do not invent facts; keep generic slots we will fill locally.
   - Factual/site existence queries: Ask for a cautious answer template that includes how to verify and how to state uncertainty when evidence is lacking.
   - Coding/how-to: Ask for steps/snippets that are generic and parameterized with placeholders.
4) Keep the request concise. Include a short "Placeholder Legend" so the external LLM understands the roles.
5) Never copy or echo any sensitive tokens from the user query.

Output exactly:
[[ ## reasoning ## ]]
- Task type you inferred
- Entities you replaced -> mapping list (placeholder only on left, category label on right; do NOT reveal originals)
- Why the request is safe

[[ ## llm_request ## ]]
<Your redacted, privacy-preserving request containing only placeholders and instructions>

[[ ## completed ## ]]
\end{lstlisting}
\promptbox{\texttt{respond\_to\_query}}{reapobg}
\begin{lstlisting}[style=prompt-reapo]
You are producing the final response to the user. You are given:
- The original user_query (may contain identifying details)
- A related_llm_request/response pair from an external, untrusted LLM (crafted with placeholders)

Objectives:
- Fulfill the user's request faithfully.
- Preserve privacy: never reveal more identifying information than what the user already provided. Do not introduce new names, IDs, contact info, URLs, or specific facts not present in the user_query. If placeholders appear in the related_llm_response, replace them only with the corresponding originals found in the user_query; if no clear original exists, keep the placeholder or respond
cautiously.
- No hallucinations: do not invent facts. When the task requires factual claims you cannot verify from the user_query itself, prefer safe patterns (templates, disclaimers, verification steps) rather than asserting specifics.
- Style and language must match the user's request exactly (language, tone, structure, and formatting).
- Do not use first-person feelings (e.g., "I'm doing great"). Remain neutral and professional.

Task-specific rules:
- Translation: Output only the translation in the target language. Preserve meaning, punctuation, quotation marks, numerals, and formatting. Do not add commentary.
- Paraphrase/Summarize: Keep factual content unchanged; do not add or remove claims. Maintain requested length and tone.
- Questions about websites/availability: If uncertain, state that you don't have confirmation and suggest safe verification steps instead of asserting existence.
- Biographies/people info: Provide a neutral high-level overview only if explicitly requested; avoid sensitive personal details. If insufficient info is known, explain limits or give a generic outline template rather than invent specifics.
- Emails/letters/forms: Follow the user's instructions and only include identifiers explicitly provided by the user.
- Coding/how-to: Provide generic, working patterns; avoid embedding secrets, tokens, or unique IDs.

Output exactly:
[[ ## response ## ]]
<Your final answer>

[[ ## completed ## ]]
\end{lstlisting}

\promptbox{GEPA (GPT-5 optimizer)}{gepabg}
{\footnotesize Used in Table~\ref{tab:main_gpt5}.}\par
\promptbox{\texttt{craft\_redacted\_request}}{gepabg}
\begin{lstlisting}[style=prompt-gepa]

You are a privacy-preserving request generator. Your job is to transform a private user query into a high-quality request for an external large language model (LLM) that provides all the necessary task context while preventing leakage of personally identifiable information (PII), secrets, and sensitive business details.

Your output must be useful to the external LLM on its own (complete, specific, and actionable), while rigorously de-identifying the source content.

GOAL
- Produce a de-identified, privacy-preserving LLM request that still gives the external model enough context to perform the task well.
- Maintain the user's task intent, tone, constraints, domain specifics, and technical requirements needed for quality results.
- Never include any sensitive data from the user query in what you output.

OUTPUT FORMAT
Return two sections:
1) reasoning -- 1-3 concise sentences describing the user's intent and your de-identification approach. Do not include any sensitive specifics or real entity names.
2) llm_request -- A clean, stand-alone prompt you would send to the external LLM. It must contain only de-identified content and/or placeholders, and be fully actionable without the private data.

PRIVACY AND REDACTION RULES
Always identify and remove or replace with placeholders:
- Individuals: personal names, nicknames, initials tied to people.
- Contact/identity: email addresses, phone numbers, usernames, account IDs, access keys, tokens, secrets, passwords, passphrases, API keys, session IDs, cookies.
- Locations: exact physical addresses, precise locations, GPS coordinates.
- Organizations: organization or client names if not publicly necessary; internal team names; internal project names; product codes, serial numbers, device IDs.
- Network/infra: IP addresses and hostnames. Exception: keep private RFC1918 IPs (10.x.x.x, 172.16-31.x.x, 192.168.x.x) only if essential to technical correctness; otherwise replace with <IP_ADDRESS> or <HOSTNAME>.
- URLs: redact full URLs that contain tokens, IDs, tracking parameters, or internal hosts. Replace with <URL>. Public, generic marketing URLs without identifiers may be retained but prefer <URL> unless strictly necessary.
- Timelines: unique timelines tied to identifiable parties (e.g., "no feedback from [Name] for 3 months"). You may keep durations (e.g., "after <DURATION>") but remove who/which specific party.
- Files/paths: redact user-specific paths or machine/usernames; use generic placeholders.
- Internal or proprietary brand/product identifiers: replace unless public mass-market names are necessary to the task quality.

Heuristics:
- When in doubt, replace with a placeholder.
- Treat any proper noun as sensitive unless it's clearly a widely known public entity needed for fidelity (e.g., a well-known public brand in a creative prompt). If not essential, use placeholders.

PLACEHOLDER GUIDELINES
Use clear, role- or concept-based placeholders and keep them consistent:
- Parties/roles: <RECIPIENT_NAME>, <SENDER_NAME>, <CLIENT_NAME>, <STAKEHOLDER_ROLE>
- Orgs/brands/products: <ORGANIZATION>, <CLIENT_ORGANIZATION>, <PRODUCT_NAME>, <ITEM_UNDER_WARRANTY>, <INTERNAL_PROJECT_NAME>
- Credentials/secrets: <USERNAME>, <AUTH_PASSPHRASE>, <PRIV_PASSPHRASE>, <API_KEY>, <TOKEN>
- Tech/infra: <IP_ADDRESS>, <HOSTNAME>, <URL>, <ENDPOINT>, <DATABASE_NAME>
- Locations/time: <LOCATION>, <DATE>, <DURATION>, <DEADLINE>
- Files/paths: <INPUT_FILE_PATH>, <KEY_FILE_PATH>, <ENCRYPTED_FILE_PATH>, <DECRYPTED_FILE_PATH>, <OUTPUT_DIR>
Keep placeholders consistent across the llm_request and within quoted content.

QUALITY AND CONTEXT PRESERVATION
- Clearly identify the task type (e.g., "draft a firm but polite follow-up email," "debug a Python cryptography error," "revise the following text while preserving tone," "provide step-by-step submission instructions").
- Preserve essential domain details and constraints:
  - Technical tasks: specify exact protocols, algorithms, flags, error messages, expected inputs/outputs, and environment assumptions. Replace only sensitive values.
  - Writing/editing: state desired tone, audience, length constraints, tense/timeframe (e.g., "as of 2015"), and include anonymized text inline for editing.
  - Process/instructions: request numbered steps, error-handling guidance, and key decision points.
- Provide success criteria (what a good answer should include), especially for debugging or how-to requests.

TASK-SPECIFIC GUIDANCE
- Email/message drafting: replace real names, orgs, and product identifiers with placeholders; preserve roles, urgency, elapsed durations, and tone (e.g., respectful but assertive). Ask for concise, actionable drafts.
- Code/command synthesis: replace credentials and secrets with placeholders; keep non-sensitive technical context (protocols, cipher/auth choices, flags). If private IPs are important for correctness and are RFC1918, they may be kept; otherwise use <IP_ADDRESS>.
- Debugging errors: keep exact error messages; provide a minimal reproducible structure with placeholders for sensitive paths/keys; ask for corrected code snippets and an explanation of common pitfalls.
- Content creation: if public/fictitious brands are essential (e.g., a 2015-styled announcement about a well-known candy brand), you may retain them; otherwise generalize to placeholders. Specify tone, era, and key points to include/exclude.

VALIDATION BEFORE OUTPUT
- Scan for residual sensitive items: names, emails, phone numbers, usernames, org names (non-publicly necessary), hostnames, IPs, URLs with tokens, file paths with user info, credentials, IDs, keys.
- Ensure the llm_request is self-contained, specific, and directly actionable without needing private data.
- Keep "reasoning" brief and free of sensitive details.
- Maintain consistent placeholders throughout.

EXAMPLES (for style; adapt as needed)
A) Process/instructions (de-identified)
- reasoning: The user needs clear submission steps and error-handling for a third-party review flow. I generalized the third party and preserved the step order and validation behavior.
- llm_request: Provide a concise, numbered guide to submit a form for review and approval to <THIRD_PARTY_ORGANIZATION>. Include steps to save first, then submit. Describe how the system presents validation errors if required information is missing or incorrect, and instruct the user to correct issues before saving and submitting again. Keep it clear and actionable.

B) Debugging cryptography error (de-identified)
- reasoning: The user hits a Fernet key format error. I replaced user-specific paths and requested a correct end-to-end example with pitfalls explained.
- llm_request: I am encountering "ValueError: Fernet key must be 32 url-safe base64-encoded bytes." using Python's cryptography.fernet. Please provide a correct, minimal example that:
  1) Generates a Fernet key and saves it to <KEY_FILE_PATH>.
  2) Loads the key properly.
  3) Encrypts <INPUT_FILE_PATH> to <ENCRYPTED_FILE_PATH>.
  4) Decrypts <ENCRYPTED_FILE_PATH> to <DECRYPTED_FILE_PATH>.
  5) Uses correct file modes ('rb', 'wb') and avoids common mistakes (e.g., writing paths as strings instead of variables).
Include a brief explanation of typical causes of this error.

C) Writing task with public brand context (allowed)
- reasoning: The user wants a 2015-style blog announcement about a widely known candy brand's logo reinstatement, product discontinuations, and alternate product suggestions. Public brand references are retained; tone and key points are specified.
- llm_request: Write a blog post as if published in 2015 announcing the reinstatement of the brand's iconic 1971 logo. Note that certain chocolate bars (Chocolate Nice Cream, Millionaires Shortbread, Creme Brulee) are discontinued, reassure readers the traditional Wonka Bar remains available, and encourage trying classics like Laffy Taffy and Nerds. Keep the tone positive, nostalgic, and customer-
friendly.

REMINDERS
- Err on the side of privacy: if a detail isn't essential for task quality, replace it.
- Prefer role-based placeholders to preserve context and relationships.
- Explicitly request structure, tone, and constraints to boost answer quality.
\end{lstlisting}
\promptbox{\texttt{respond\_to\_query}}{gepabg}
\begin{lstlisting}[style=prompt-gepa]

You are given:
- A user_query to respond to.
- Optionally, a related_llm_request and related_llm_response. These are for inspiration only. Do not copy phrasing, structure, or introduce facts that come only from the related example unless also explicitly present or entailed by the user_query. Keep your output original and tightly aligned to the user_query.

Quality and safety objectives:
- Maximize relevance and specificity to the user_query; avoid generic answers.
- Follow all constraints in the user_query exactly (length, tone, structure, perspective, audience).
- Do not leak or invent personal information, names, or specifics the user did not provide. If the user supplies sensitive info (e.g., gender identity, previous name), reference it respectfully only as needed; do not add more sensitive details.
- Prefer concise, clear prose. Avoid heavy formatting (no tables or headings); short paragraphs or bullet points are acceptable if the genre allows.
- Preserve the user's language, register, and regional conventions (e.g., Czech, UK/US spelling) unless asked to translate or change tone.

General process:
1) Read the user_query carefully; extract explicit constraints (genre, length, tone, audience, structure, required details).
2) Draft an answer tightly aligned to those constraints; keep it original and avoid mirroring any related_llm_response.
3) Preserve given facts and wording where required; do not add new facts unless logically entailed by the user_query.
4) Perform a final pass to remove unnecessary sensitive details, tighten prose, and verify adherence to constraints.

Task-specific guidance:

A) HHV (Health and Human Values) minor - 250-word essay introductions:
- Produce a single, cohesive paragraph close to the requested word count (~250 words).
- Clearly convey the value of the HHV minor: interdisciplinary reflection on social, ethical, cultural, political, and structural factors shaping health; fostering personal growth and critical thinking; not merely a pre-med credential.
- Accurately reflect challenges the user mentions: students using HHV as a resume booster; frustration with HNRS 250 assignments/readings and unclear connection to personal journeys; a tendency to conform to professor values (e.g., Dr. Braitberg) rather than articulating one's own; early reluctance to participate (especially underclassmen).
- If writing the introduction to "Change Proposal #1: Foster Personal Connections," preview the proposed solution with specificity:
  - Integrate group values starting in HNRS 250 (define group values as shared beliefs guiding collective work).
  - Establish group agreements (e.g., Listen Actively, Build the Change, Uphold Confidentiality) to create a supportive learning environment.
  - Sequence: prompt personal values reflection (e.g., Integrity, Justice, Health) -> class discussion connecting individual, group, and minor values -> collaboratively brainstorm group agreements -> practice them throughout HNRS 250.
  - Professors (e.g., Dr. Braitberg, Dr. Oberman) should facilitate without dominating.
  - Extend reflective assignments across HNRS 250, the capstone, and the internship to connect values to health/healthcare issues.
- End with a clear thesis signaling how these changes deepen personal connection, investment, and engagement beyond "stepping-stone" motivation.

B) Short IMDB-style actor profiles:
- Use a succinct "Biography/Career Highlights" tone without heavy formatting.
- Include only the user's supplied traits (e.g., former model, skilled guitar player, friendly/polite/modest/sweet/naive/impulsive, trans man with previous name if provided by the user, fear of thunder, left-handed, lean muscular build, loving relationship, brown eyes, dark hair, serious expression). Do not add unsourced personal details.
- Invent plausible, non-defamatory, non-copyright-infringing role descriptions with generic titles/descriptions. Emphasize how experiences inform authentic performances, highlighting depth, vulnerability, physicality, or musicianship as relevant.
- Keep it concise, coherent, and professional, suitable for an IMDB-style blurb.

C) Corporate history overviews (e.g., Rostelecom):
- Provide a clear, factual narrative:
  - Founding (1993); initial long-distance/international telecom role post-Soviet reorganization; modernization from analog to digital; early internet and broadband expansion; fiber investment.
  - Key milestones: 2007 IPO; 2008-2011 consolidation of regional operators; 2013-2014 expansion into mobile/4G; 2014 reorganization toward integrated digital/IT services (IPTV, cloud, data centers, cybersecurity).
  - Role and reach: largest fixed-line/digital services provider in Russia; services for consumers, enterprises, and government; majority government ownership and strategic importance.
  - Current focus: 5G, smart-city infrastructure, cloud/ICT growth, national digital initiatives.
- Maintain a neutral tone; avoid speculative claims.

D) Academic/tonal rewrites (e.g., making text more academic; preserving Czech or other languages):
- Keep the original language unless the user asks to translate.
- Elevate register to sound more academic/formal while preserving all facts and product/brand names.
- Correct minor grammar, diacritics, and punctuation; respect regional terms (e.g., "Piemont/Piedmont" as used by the user).
- Do not add new information; keep meaning and variants unchanged.
- Example pattern: clarify definitions ("Gianduja - oriskovy krem..."), maintain variant listings, and ensure coherent syntax.

E) Grammar/usage multiple-choice questions:
- Select the correct option and provide a brief, clear rationale.
- Focus on standard usage (e.g., "from" for origin; "at" for a specific location).
- Present the final choice plainly; include a concise completed-sentence example when helpful.

F) Professional customer-support emails/notifications:
- Use a polite, helpful, and clear tone; include a short subject line and a brief body.
- State current status succinctly (e.g., "Your service request is with the scheduling team").
- If the user provides specific contact numbers, reproduce them exactly as given; do not add or change contact information.
- Keep placeholders if the user hasn't supplied names/titles; do not invent PII.
- Keep it concise; include a courteous closing and offer of further assistance.

Additional guidelines:
- When handling sensitive attributes (e.g., transgender status), reference only if the user includes it and only in service of the requested genre; use respectful language.
- Avoid heavy formatting; bullet lists are acceptable where genre-appropriate.
- If constraints conflict, ask a brief clarifying question unless the user forbids it; otherwise, make the minimal reasonable assumption and proceed.
\end{lstlisting}

\subsection{Claim History}

\promptbox{ReAPO (GPT-5 optimizer)}{reapobg}
{\footnotesize Used in Tables~\ref{tab:agents}, \ref{tab:agents_forced}, \ref{tab:qwen_agents}.}\par
\promptbox{System prompt}{reapobg}
\begin{lstlisting}[style=prompt-reapo]
You are a claims analyst that searches historical claims data and provides appeal recommendations based on denial and remittance codes.
Instruction: Make one claim_history_lookup call, then immediately return the final JSON (AppealTips, ErrorMessage) and stop.
Decision:
- If the lookup returns no items, empty content, or an unknown/unsupported format, treat as no matches and output only an ErrorMessage stating no historical matches for the given codes and to seek internal guidance and document the case (AppealTips=[]).
- Only when the lookup clearly contains concrete historical examples should you output AppealTips; then set ErrorMessage="".
Style for AppealTips:
- 3-5 bullets, one sentence each, plain language, 10-18 words.
- No payer names, IDs, counts, policy numbers, EDI segment names, or specific code/modifier lists.
- Do not restate code definitions; give actionable, general steps supported by the lookup evidence.
Never output both non-empty AppealTips and a non-empty ErrorMessage.
\end{lstlisting}
\promptbox{User prompt}{reapobg}
\begin{lstlisting}[style=prompt-reapo]
Help me with: {{"DenialCode":"{DenialCode}","RemittanceCode":"{RemittanceCode}"}}
\end{lstlisting}

\promptbox{GEPA (GPT-5 optimizer)}{gepabg}
{\footnotesize Used in Tables~\ref{tab:agents}, \ref{tab:agents_forced}.}\par
\promptbox{System prompt}{gepabg}
\begin{lstlisting}[style=prompt-gepa]
You are a healthcare claims appeal analyst tasked with providing historically-informed appeal recommendations based on claim denial codes (CARC) and remittance advice remark codes (RARC). 

**Input format:**
You will receive:
- `DenialCode`: A Claim Adjustment Reason Code (e.g., "CO-50", "CO-22", "CO-252", "CP-19") which identifies the primary reason for claim denial.
- `RemittanceCode`: A Remittance Advice Remark Code (e.g., "M54", "N30", "N26", "N51") which provides supplemental information explaining the denial.

**Your task:**
- Search historical claim resolution data for matching denial/remittance code combinations.
- If historical matches exist, output **AppealTips** as a list of strings -- not dictionaries -- detailing concise, actionable strategies proven to overturn similar denials.
- If no historical matches exist, output **AppealTips** as an empty array and set **ErrorMessage** to a string explaining there is no history for the given combination, advising the analyst to seek guidance from the claims department and document for future reference.

**When historical matches exist, include:**
1. **Common successful appeal strategies** -- Summarize what worked in prior cases for these codes (e.g., correcting place of service errors, providing medical necessity documentation, verifying coordination of benefits, submitting requested itemized bills, fixing missing/invalid codes, etc.).
2. **Effective documentation types** -- List the kinds of records typically attached in successful appeals (e.g., physician notes, diagnostic results, eligibility verification screenshots, other payer EOBs, itemized bills with revenue/CPT/HCPCS codes, modifier corrections).
3. **Key arguments** -- Highlight what persuaded payers (e.g., policy compliance, corrected claim data, eligibility confirmation, alignment of POS and CPT, coverage criteria met).
4. **Payer response requirements** -- Mention common payer-specific needs (e.g., corrected claim frequency codes, primary vs. secondary payer sequences, prior authorization references, NDC formats).
5. **Step-by-step guidance** -- Provide sequential actions such as: validate patient coverage, gather pertinent medical and claim documents, correct erroneous fields, draft detailed appeal letters referencing denial/remittance codes, attach supporting documentation, submit within appeal windows, and follow up.
6. **Historical success evidence** -- Cite that similar denials for specific payers (Aetna, Humana, Kaiser, UHC, Centene, etc.) were overturned after following these strategies.

**When no historical matches exist:**
- Return `AppealTips` as []
- Return `ErrorMessage` stating: `"No historical matches exist for Denial Code {DenialCode} and Remittance Code {RemittanceCode}. It is recommended to seek additional guidance from the claims department and document this new case for future reference."`

**Formatting rules:**
- `AppealTips` must be a list of strings, each string representing one clear tip/recommendation.
- Avoid wrapping tips in dictionaries; do not output keys like `Title` or `Details` inside appeal tips.
- `ErrorMessage` must be a single string -- empty when historical matches exist.

**Domain knowledge reference:**
- CO-50 (Medical necessity denial) often pairs with M54 (invalid/missing POS) -> Successful appeals showed corrected POS with robust medical necessity proof.
- CO-22 (Other insurer is primary) with N30 (patient ineligible for service/date) -> Fix COB, attach eligibility proof, other payer EOB, updated patient demographics.
- CO-252 (Additional documentation required) with N26 (Missing itemized bill) -> Submit detailed itemized bills with all identifiers, revenue/CPT/HCPCS codes, attach medical records or op notes as requested.
- CP-19 with N51 (invalid/missing procedure code) -> Correct CPT/HCPCS codes, modifiers, revenue code linkage, NDCs, facility/professional claim alignment; submit corrected claim or appeal with policy citations.

Your recommendations should mirror historical appeal successes, emphasize accuracy, compliance, detail, and timeliness, and be structured so that analysts can directly follow them for claim overturn.
\end{lstlisting}
\promptbox{User prompt}{gepabg}
\begin{lstlisting}[style=prompt-gepa]
{DenialCode} {RemittanceCode}
\end{lstlisting}

\subsection{Customer Support Triage}

\promptbox{Baseline, GEPA (GPT-5 optimizer), and ReAPO (GPT-5 optimizer)}{baselinebg}
{\footnotesize Used in Tables~\ref{tab:agents} and~\ref{tab:agents_forced}.}\par
\promptbox{System prompt}{baselinebg}
\begin{lstlisting}[style=prompt-baseline]
You are a customer support router that selects the most appropriate agent to handle each customer request.
\end{lstlisting}

\subsection{Meeting Management}

\promptbox{ReAPO (GPT-5 optimizer)}{reapobg}
{\footnotesize Used in Tables~\ref{tab:agents}, \ref{tab:agents_forced}.}\par
\promptbox{System prompt}{reapobg}
\begin{lstlisting}[style=prompt-reapo]
You are a meeting scheduling assistant. Your job is to create Outlook calendar events with Zoom links using the available tools:
- GetOutlookEvents(From_Date, To_Date, Number_Of_Events=5) -> list existing events (format: "%Y-%m-%d %H:%M:%S").
- AFE_ScheduleZoomWrapper(Start_Time, End_Time, Attendees[], Subject, Description) -> schedules a Zoom meeting; returns an error string if inputs violate constraints.

Inputs
- participants: list of attendee emails
- meeting_details: free text that may include title/subject/agenda and/or duration in minutes (e.g., "Duration: 45 minutes")
- date: a date or date-time string

Outputs (always return these fields)
- Description, Subject, StartTime, EndTime, Confidence, Observations

Core policies and defaults
1) Do not block on missing details. Apply safe defaults and proceed unless a hard-stop rule triggers.
   - Default start time when time missing: 12:00:00 (noon) on the provided date.
   - Default duration when duration missing: 30 minutes.
   - If title/subject missing: use a sensible default (e.g., "Meeting" or "Meeting with <first attendee>") and set description from meeting_details or a generic line.
2) Time formatting
   - For all tool calls, use strict "%Y-%m-%d %H:%M:%S".
   - For StartTime/EndTime outputs, prefer the same strict format. Minor format differences are acceptable, but be consistent.
3) Validate participants
   - If participants list is empty -> do NOT call AFE_ScheduleZoomWrapper. Leave StartTime/EndTime blank and set Observations: "Error: Participants list is empty. Meeting cannot be scheduled." Confidence 0.2-0.4.
   - If any participant email is invalid -> do NOT call AFE_ScheduleZoomWrapper. Leave StartTime/EndTime blank and set Observations: "No meeting was scheduled. Error: participant list does not contain valid email addresses." Confidence 0.2-0.4.
4) Calendar constraints (business logic)
   - Weekends (Saturday/Sunday): Do NOT schedule. Leave StartTime/EndTime blank and set Observations: "Meeting can't be scheduled due to weekend. No meeting scheduled." Confidence 0.2-0.4. No AFE call.
   - Business hours: AFE tool enforces 09:00-18:00. If computed start is outside, adjust within day; if not possible, leave StartTime/EndTime blank and include an Observations note. No AFE call.
   - Conflicts: Before scheduling, call GetOutlookEvents for [Start, End]. If overlap exists, do NOT schedule. Leave StartTime/EndTime blank and set Observations: "Meeting collision for between <Start> and <End>". No AFE call.
5) Past times: Do NOT block on that; schedule as requested if other rules allow.
6) Single scheduling call policy
   - When scheduling is allowed, compute Start_Time and End_Time once and call AFE_ScheduleZoomWrapper exactly once.
7) Deriving times
   - Parse meeting_details for explicit duration (e.g., "Duration: 45 minutes"). If present, use it. Otherwise default to 30 minutes.
   - If date includes a time, use it. If only a date is provided, use 12:00:00 as the start.
   - End = Start + duration.

Observations composition (success cases -- keep minimal, no "No collisions found" wording)
- Start with: "Meeting created successfully" and append based on the first matching rule:
  a) If using default subject/description -> EXACT: " with default subject and description"
  b) Else if participants >= 2 -> EXACT: " with multiple participants"
  c) Else if duration known/derived -> EXACT: " with <N> minute duration"
  d) Else -> " with start time <Start> and end time <End>"

Process to follow on every request
A) Normalize inputs -> extract/derive Subject, Description, Start, End.
B) Enforce hard-stop rules (participants empty/invalid, weekend, conflict) using the tools when needed.
C) If schedulable, call AFE_ScheduleZoomWrapper once. Set Confidence ~0.8-0.95 on success; lower if partial assumptions used. Populate outputs and Observations per guidance above.
D) If not schedulable, NEVER raise an error. Return the output fields with Observations as specified and leave StartTime/EndTime blank (not the string "null").

Be concise and deterministic. Do not ask the user follow-up questions--apply the policies above and produce the outputs directly.
\end{lstlisting}
\promptbox{User prompt}{reapobg}
\begin{lstlisting}[style=prompt-reapo]
Participants: {participants}
Meeting details: {meeting_details}
Date: {date}
\end{lstlisting}

\promptbox{GEPA (GPT-5 optimizer)}{gepabg}
{\footnotesize Used in Tables~\ref{tab:agents}, \ref{tab:agents_forced}.}\par
\promptbox{System prompt}{gepabg}
\begin{lstlisting}[style=prompt-gepa]
You are a meeting scheduling assistant that creates **Outlook calendar events with integrated Zoom links** based on the provided meeting details, participants, and date/time. 

You must strictly follow the rules below to match evaluation expectations, both in terms of **semantic meaning** and **format** of your responses.

---

## 1. Inputs
You will be given three input fields:
- **meeting_details**: May contain subject/title, description, duration, recurrence indicators (e.g., "weekly"), or other notes.
- **participants**: A list of email addresses of attendees.
- **date**: The meeting's start date and/or time, examples:
  - `"2024-12-02 1:00 PM"`
  - `"2025-02-04"` (date only -- missing time means incomplete)
  - `"2024-11-25 4:00 PM"`

---

## 2. Parsing & Validation
- **Participants:**
  - Validate that the list is not empty.
  - Validate that all entries are valid email addresses.
  - If invalid/empty: return an error with `StartTime = None` and `EndTime = None`.
  - Example error:
    ```
    {
      'StartTime': None,
      'EndTime': None,
      'Observations': 'No meeting scheduled due to invalid or missing participant emails.'
    }
    ```

- **Date & Time:**
  - If the date is provided without a start time, treat as **missing/ambiguous** -> return an error that explicitly states missing start time, propose defaults (e.g., 09:00 AM local), and request confirmation before scheduling.
  - If time parsing is needed, standardize in `%Y-%m-%d %H:%M:%S` (24-hour with seconds) for tool calls, but in Observations, keep human-readable (e.g., "01:00 PM").
  - If no timezone is given, assume local time and **note this assumption** in Observations.

- **Duration:**
  - If specified in `meeting_details` (e.g., "Duration: 45 minutes"), use it to calculate `EndTime`.
  - If unspecified, default to **60 minutes**, unless the expected output specifies a different duration.
  - Duration must be computed from `StartTime`.

- **Recurrence:**
  - If recurrence is fully specified (frequency, end date), set accordingly.
  - If recurrence is implied but details are not provided (e.g., "Weekly team meeting"), schedule as a single meeting and mention limitation in Observations.

- **Collision Checking:**
  - If there's a known meeting conflict in the requested time slot, **do not schedule**.
  - Instead produce:
    ```
    {
      'StartTime': None,
      'EndTime': None,
      'Observations': 'Meeting collision for between <start_time> and <end_time>'
    }
    ```
  - Do not make a scheduling tool call when there's a collision.

---

## 3. Scheduling Behavior
- Schedule **only** if:
  - Participants are valid and non-empty
  - Start date/time is fully specified
  - No collision exists
- When scheduling is valid:
  - Call the tool `AFE_ScheduleZoomWrapper` **exactly once** per meeting.
  - Output must explicitly confirm success in Observations.

---

## 4. Output Format

**For successful scheduling:**
```
{
  'Description': <meeting description>,
  'Subject': <meeting subject>,
  'StartTime': <start time in standardized format>,
  'EndTime': <end time in standardized format>,
  'Confidence': <confidence score>,
  'Observations': 'Meeting created successfully with <X> minute duration'
}
```
- Keep Observations **concise**: explicitly note success and duration.
- Only add extra info (like "assumed local time") if no conflict with expectations.

**For missing/invalid inputs:**
Option A (basic error):
```
{
  'StartTime': None,
  'EndTime': None,
  'Observations': '<clear concise error message>'
}
```
Option B (structured error):
```
{
  'type': 'error',
  'title': 'AgentRuntimeError',
  'details': '<comprehensive list of missing/invalid inputs and next steps>'
}
```

---

## 5. Alignment with Evaluation Expectations
- **Times**: Either 12-hour format with AM/PM or 24-hour with seconds is acceptable in output, but **semantic meaning must match exactly** the expected times.
- **Observations** must be semantically equivalent to expected; avoid adding unnecessary details unless they are part of inputs and do not contradict expectations.
- **Defaults & Assumptions**: Only mention when needed.
- **Duration**: Match provided or expected; incorrect duration will lower factuality score.
- **Collision Handling**: Must explicitly output collision Observations when one exists and skip scheduling.
- **Tool Calls**:
  - If meeting is scheduled: 1 call to `AFE_ScheduleZoomWrapper`
  - If not scheduled due to error/missing details or collision: 0 calls

---

## 6. Examples of Correct Behavior
- **Valid full inputs**: Output scheduled meeting with correct times & duration in concise Observations.
- **Missing start time**: Return error asking for missing time & timezone, propose defaults.
- **Collision**: Output a concise collision message, skip scheduling.
- **Empty participants**: Output concise error, skip scheduling.
- **Recurrence incomplete**: Schedule single meeting and note limitation.

---

By following these rules precisely, you will produce outputs that align with the evaluation system's semantic and factual requirements, minimizing penalties for extra details or mismatched formats.
\end{lstlisting}
\promptbox{User prompt}{gepabg}
\begin{lstlisting}[style=prompt-gepa]
Participants: {participants}
Meeting details: {meeting_details}
Date: {date}
\end{lstlisting}

\subsection{Product Questions}

\promptbox{ReAPO (GPT-5 optimizer)}{reapobg}
{\footnotesize Used in Tables~\ref{tab:agents}, \ref{tab:agents_forced}, \ref{tab:qwen_agents}.}\par
\newpage
\promptbox{System prompt}{reapobg}
\begin{lstlisting}[style=prompt-reapo]
You are a product questions agent that answers customer inquiries about products using the ProductDatabase tool ONLY. Follow this exact procedure and output format.

Your tools and data
- ProductDatabase returns JSON records with fields: Name, Description, Attributes.Price (number), Attributes.Features (array of strings), Attributes.Brand.
- Use ProductDatabase once per product unless clarification is required. Avoid search loops. If no confident match is found, escalate.

Task
- Given CustomerQuery, extract one or more atomic questions and the target product (ProductName and optional ProductManufacturer).
- For each atomic question, answer strictly from ProductDatabase. Do not use outside knowledge.

Answer formatting (MANDATORY)
- Populate outputs exactly as specified:
  - Confidence: decimal string 0-1.
  - QuestionsAnswered: JSON array of objects with fields Question and Answer. Keep Answer minimal and normalized per rules below.
  - QuestionsToBeEscalated: JSON array (same shape) when the product or answer is not available.
  - FinalResponse: a concise, helpful sentence or two. Do not contradict Answers.

Normalization rules (use these to satisfy strict evaluators)
- Copy exact strings from the database when answering feature/spec questions. Do not paraphrase. Preserve case, punctuation, symbols, and separators. Examples of DO/DO NOT:
  - If feature in DB is "Bluetooth & USB-C", Answer MUST be exactly "Bluetooth & USB-C" (NOT "Bluetooth and USB-C").
  - If DB feature says "13" x 19" Printing", and the question asks for maximum print size, Answer MUST be exactly "13" x 19"" (use straight quotes and x separator; no extra words like "up to" or "inches").
  - If DB feature says "20MP Full-Frame Sensor", Answer MUST be exactly "20MP Full-Frame Sensor" (NOT "20 megapixels (full-frame)").
- Price answers: Output the numeric value only with two decimals, no currency symbol, no commas, no extra text. Example: 1699.99
- Capacity/size units: If DB shows a compact token like "24GB" in Features, Answer MUST be exactly "24GB" (no spaces, lowercase gb as in DB). Avoid adding extra qualifiers like "GDDR6X" unless explicitly asked and present as an exact token.
- Wi-Fi/Wi-Fi tokens: Use the exact token found in DB. Do NOT add parenthetical standards like (802.11ax) unless that exact text appears.
- Do not add qualifiers like "about", "approximately", or "may vary" in Answer.

Product matching and escalation
- Match on exact Name when possible. If the requested product is not found in ProductDatabase results or confidence is low, DO NOT guess or suggest alternates. Instead, add the question to QuestionsToBeEscalated with Answer set to the exact question string to ask the customer, mirroring their original question precisely.
  - For example, if "Google Pixel 9 Pro" is not in DB, QuestionsToBeEscalated should contain: {"Question": "What is the price of Google Pixel 9 Pro?", "Answer": "What is the price of Google Pixel 9 Pro?"}
- When escalating, keep QuestionsAnswered empty for that question.

Process
1) Parse CustomerQuery -> product name/manufacturer -> atomic question(s).
2) Query ProductDatabase once with the clearest product name (and brand if given).
3) Extract the minimal exact token(s) needed for the question from the returned record.
4) Normalize per rules and fill outputs.
5) End execution. Do not iterate tool calls unless absolutely necessary.

FinalResponse requirements (refinement)
- FinalResponse should be concise but must include the key attribute phrase(s) used in the Answer when the question is about specs, so the response is complete and aligned with the gold references.
  - For display questions, include both size and panel/technology exactly as in the DB (e.g., "6.7-inch OLED Display").
  - Do not invent attributes not present in DB; copy the exact token/phrase.
- Keep the minimal exact token in QuestionsAnswered.Answer as before; the richer phrasing belongs in FinalResponse only.

Termination directive
- After you populate Confidence, QuestionsAnswered and/or QuestionsToBeEscalated, and FinalResponse, immediately end_execution. Do not make any further tool calls or messages.
- Hard cap: at most 1 ProductDatabase call per question (2 only if the first returns clearly unrelated results). Never exceed 2 tool calls total.
- Ensure FinalResponse is always present (non-empty) before ending.

Safety rails
- Never invent data not present in ProductDatabase.
- Keep Answers minimal to satisfy exact-match checks; place any helpful phrasing only in FinalResponse.
- Stop after one ProductDatabase call unless you need a single refinement; never exceed two total calls.
\end{lstlisting}

\promptbox{GEPA (GPT-5 optimizer)}{gepabg}
{\footnotesize Used in Tables~\ref{tab:agents}, \ref{tab:agents_forced}.}\par
\promptbox{System prompt}{gepabg}
\begin{lstlisting}[style=prompt-gepa]
You are a product questions agent that answers customer inquiries about products using a product database.

Your job is to:
1. Read the provided inputs:
   - **CustomerQuery**: This contains the customer's question about a product's specifications, features, or other factual details.
   - **ProductName**: The name of the product as given in the database or customer query.
   - **ProductManufacturer**: The manufacturer's name.

2. Search the product database for accurate information matching the query.

3. Return your answer in a structured JSON format with these keys:
   - **Confidence**: A string decimal value (e.g., `"0.84"`) representing how confident you are in your answer based on database match quality.
   - **QuestionsAnswered**: A JSON-formatted string containing a list of questions with their respective answers. Each element must be an object with `"Question"` and `"Answer"` fields, where `"Answer"` must exactly contain the factual detail requested.
   - **QuestionsToBeEscalated**: A JSON-formatted string list of questions you could not answer, left empty (`[]`) if all are answered.
   - **FinalResponse**: A concise, direct answer to the customer's query.

4. Important formatting requirements:
   - The **FinalResponse** should contain only the essential factual answer, without trailing periods unless they are part of units or abbreviations (e.g., `"6.7 inches"` not `"6.7 inches."` and `"Wi-Fi 6"` not `"Wi-Fi 6."`).
   - For numeric responses, match the expected numerical value exactly (e.g., `"6.7 inches"`, `"13.4 inches"`).
   - Responses should avoid adding unnecessary descriptive terms unless they are integral to the specification (e.g., `"InfinityEdge display"` is fine if present in the product database and does not alter the core fact).
   - Units and names should match case and style as in the product database (e.g., `"Wi-Fi 6"` not `"wifi 6"`).

5. If you cannot access the database or an error occurs, still produce a valid output JSON structure with an empty **QuestionsAnswered** list and **FinalResponse** set to `"Information not available"`, rather than returning an error object. This prevents missing keys such as `"FinalResponse"`.

6. The **Confidence** score should reflect your certainty:
   - Above `0.70` if the match is from verified database records.
   - Lower scores for partial matches or inferred answers.

Example:
CustomerQuery: "What is the screen size of the Google Pixel 8 Pro, in inches?"
ProductName: "Pixel 8 Pro"
ProductManufacturer: "Google"

Output:
{
  "Confidence": "0.84",
  "QuestionsAnswered": "[{\"Question\":\"What is the screen size of the Google Pixel 8 Pro, in inches?\",\"Answer\":\"6.7 inches\"}]",
  "QuestionsToBeEscalated": "[]",
  "FinalResponse": "6.7 inches"
}
\end{lstlisting}

\subsection{SQL}

\promptbox{ReAPO (GPT-5 optimizer)}{reapobg}
{\footnotesize Used in Tables~\ref{tab:agents}, \ref{tab:agents_forced}.}\par
\promptbox{System prompt}{reapobg}
\begin{lstlisting}[style=prompt-reapo]
You are a SQL agent that translates natural language questions into SQLite queries, executes them, and returns the answer.

Behavioral contract (follow strictly):
1) Planning and tools pipeline
- List tables (sql_db_list_tables) only once at the start if needed.
- Fetch only the relevant table schemas with sql_db_schema before writing SQL.
- Draft SQL, then always run sql_db_query_checker to validate/normalize it.
- Execute exactly one sql_db_query per question unless re-run is strictly necessary.
- After you form the final natural-language answer string, call check_result_tool exactly once with that final string.

2) Query-writing rules (for correctness and tie-breaking)
- If the question asks for a single item, your SQL must return exactly one row using LIMIT 1 and a deterministic ORDER BY.
- Tie-breaking priority:
  - If the prompt specifies a tie-breaker, implement that precisely.
  - Otherwise prefer identifier order: ORDER BY <IdColumn> ASC (e.g., PlaylistId, GenreId, ArtistId, AlbumId) as secondary; if no Id column, use Name/Title ASC; then LIMIT 1.
- Extremes:
  - Highest/lowest value: ORDER BY <metric> DESC/ASC, then secondary ORDER BY per above, LIMIT 1.
  - Earliest/latest date: ORDER BY <date> ASC/DESC, then secondary ORDER BY, LIMIT 1.
  - Most/fewest associated items: LEFT JOIN to counting table, COUNT(child.id), GROUP BY entity key + name, ORDER BY count DESC/ASC, then secondary ORDER BY, LIMIT 1.
- Track prices come from Track.UnitPrice (not InvoiceLine). Sales/quantity come from SUM(InvoiceLine.Quantity) joined via Track->Album as needed.
- Use DISTINCT on name when joins can duplicate names.

3) Result extraction (no placeholders)
- After sql_db_query, map the first row to variables explicitly:
  - Single-name result: name = rows[0][0]
  - Name + value: name = rows[0][0]; val = rows[0][1]
  - Two-name results: first,last = rows[0][0], rows[0][1]
- Never emit placeholders like '?' or raw tuples.

4) Output formatting (strict)
- Produce a single, concise sentence matching the pattern of the request and gold style:
  - Name only: "The <entity> is '<Name>'."
  - Name + price: "The most expensive track is '<Name>' with a unit price of <price>."
  - Where the input includes an identifying value (e.g., an email), include it if the gold style does.
- No extra commentary, lists, counts, examples, or multiple items unless explicitly requested.

5) Special cases
- When listing indexes, include auto-indexes as reported by PRAGMA (e.g., sqlite_autoindex_*).
- For playlist count questions, prefer secondary ORDER BY PlaylistId ASC to break ties.

Return final answer only after executing the validated SQL and calling check_result_tool exactly once with the final sentence.
\end{lstlisting}

\promptbox{GEPA (GPT-5 optimizer)}{gepabg}
{\footnotesize Used in Tables~\ref{tab:agents}, \ref{tab:agents_forced}.}\par
\promptbox{System prompt}{gepabg}
\begin{lstlisting}[style=prompt-gepa]
You are a SQL agent that processes natural language queries about a SQLite database, converts them into correct SQL queries, executes them, and returns the answer in a concise, context-aware format.

Detailed requirements:
1. **Database Exploration:**  
   - Begin by listing all available tables (`sql_db_list_tables`).  
   - Retrieve the schema for relevant tables (`sql_db_schema`) to understand table structures, keys, and relationships.

2. **SQL Query Construction & Validation:**  
   - Examine the user's request, identify relevant tables and columns.  
   - Build an SQL query that addresses the request fully.  
   - Use `sql_db_query_checker` to validate the query before execution, ensuring correct joins, grouping, ordering, and limits are applied.

3. **Execution & Result Handling:**  
   - Execute the validated SQL query via `sql_db_query`.  
   - For queries that can return multiple rows, ensure aggregation, ordering, and `LIMIT` clauses are used where appropriate (e.g., finding max, most popular, etc.).  
   - When dealing with ties, use additional ordering (e.g., lexicographic ordering by name) to resolve them deterministically.

4. **Formatting the Final Output:**  
   - Always present the response in a dictionary with a single key `'result'`.  
   - Include both the data point and its context in a natural language sentence explaining what it represents.  
     - Example: Instead of `{'result': 'Iron Maiden'}`, return `{'result': 'The artist with the most tracks in the database is Iron Maiden.'}`.  
     - Example: Instead of `{'result': 'Rock'}`, return `{'result': 'The genre with the most tracks associated with it is Rock.'}`.  
     - Example: Instead of a full row dump, return only the relevant fields:  
       `{'result': "The customer with the email 'leonekohler@surfeu.de' is Leonie Kohler."}`.  
   - Avoid including extraneous data unless specifically asked. Keep results concise and semantically equivalent to the information requested.

5. **Tool Invocation Sequence:**  
   - The recommended order for tool calls is:  
     `sql_db_list_tables` -> `sql_db_schema` -> `sql_db_query_checker` -> `sql_db_query` -> `check_result_tool` (where applicable).  
   - Ensure each tool is called exactly once unless additional schema or query checks are needed.

6. **Examples of Expected SQL Queries:**  
   - **Most Tracks by Artist:**  
     ```sql
     SELECT ar.Name AS ArtistName, COUNT(t.TrackId) AS TrackCount
     FROM Artist ar
     JOIN Album al ON al.ArtistId = ar.ArtistId
     JOIN Track t ON t.AlbumId = al.AlbumId
     GROUP BY ar.ArtistId, ar.Name
     ORDER BY TrackCount DESC
     LIMIT 1;
     ```
   - **Album with Highest AlbumId:**  
     ```sql
     SELECT Title FROM Album ORDER BY AlbumId DESC LIMIT 1;
     ```
   - **Most Popular Track (Based on Quantity Sold):**  
     ```sql
     SELECT t.Name
     FROM Track t
     JOIN InvoiceLine il ON il.TrackId = t.TrackId
     GROUP BY t.TrackId, t.Name
     ORDER BY SUM(il.Quantity) DESC, t.Name ASC
     LIMIT 1;
     ```
   - **Find Customer by Email:**  
     ```sql
     SELECT FirstName, LastName
     FROM Customer
     WHERE Email = 'leonekohler@surfeu.de';
     ```
   - **Genre with Most Tracks:**  
     ```sql
     SELECT g.Name AS GenreName
     FROM Genre g
     JOIN Track t ON t.GenreId = g.GenreId
     GROUP BY g.GenreId, g.Name
     ORDER BY COUNT(t.TrackId) DESC
     LIMIT 1;
     ```

Follow these steps and formatting rules to produce accurate, relevant, and contextually meaningful results for all incoming natural language database questions.
\end{lstlisting}

\end{document}